\documentclass[10pt,twocolumn,letterpaper]{article}

\usepackage{iccv}
\usepackage{times}
\usepackage{epsfig}
\usepackage{graphicx}
\usepackage{amsmath}
\usepackage{amssymb}

\usepackage{xcolor,colortbl}
\usepackage{adjustbox}
\usepackage{multirow}

\usepackage[font=small,labelfont=bf]{caption}
\usepackage{subcaption}
\usepackage{dirtytalk}
\usepackage[pagebackref=true,breaklinks=true,letterpaper=true,colorlinks,bookmarks=false]{hyperref}
\usepackage[capitalize]{cleveref}
\usepackage{xspace}
\usepackage{mwe}
\usepackage[accsupp]{axessibility}  % Improves PDF readability for those with disabilities.

\iccvfinalcopy % *** Uncomment this line for the final submission

\def\iccvPaperID{12703} % *** Enter the ICCV Paper ID here

\ificcvfinal\fi
\newcommand{\camready}[1]{\textcolor{black}{#1}}

\newcommand{\rulesep}{\unskip\ \vrule\ }

\def\@fnsymbol#1{\ensuremath{\ifcase#1\or *\or \dagger\or \ddagger\or
   \mathsection\or \mathparagraph\or \|\or **\or \dagger\dagger
   \or \ddagger\ddagger \else\@ctrerr\fi}}
\newcommand{\ssymbol}[1]{^{\@fnsymbol{#1}}}

\begin{document}

%%%%%%%%% TITLE
% \title{Leveraging Text-To-Image Models for Text-Based Image Segmentation}
\title{LD-ZNet: A Latent Diffusion Approach for Text-Based Image Segmentation}
\author{Koutilya PNVR$\ssymbol{2}$\thanks{This work was done when Koutilya and Bharat were at Amazon.}
\and
Bharat Singh$\ssymbol{3}$$^\ast$
\and
Pallabi Ghosh$\ssymbol{4}$
\and
Behjat Siddiquie$\ssymbol{4}$
\and
David Jacobs$\ssymbol{2}$
\and
University of Maryland College Park$\ssymbol{2}\qquad$\hspace{0.3cm}Vchar.ai$\ssymbol{3}\qquad\qquad$\hspace{1.6cm}Amazon$\ssymbol{4}\qquad\qquad$\hspace{0.2cm}\\
{\tt\small $\{$koutilya, djacobs$\}$@umiacs.umd.edu$\qquad$bharat@vchar.ai$\qquad\{$gpallabi, behjats$\}$@amazon.com}
% \and
% \url{https://koutilya-pnvr.github.io/LD-ZNet/}
}

% \and

\maketitle
\ificcvfinal
\thispagestyle{empty}
\fi

\begin{abstract}
% Lala \url{www.example.com}.

% \urlstyle{sf}
% Lala \url{www.example.com}.

% \renewcommand\UrlFont{\color{red}\rmfamily\itshape}
% Lala \url{www.example.com}.
\camready{Large-scale pre-training tasks like image classification, captioning, or self-supervised techniques do not incentivize learning the semantic boundaries of objects. However, recent generative foundation models built using text-based latent diffusion techniques may learn semantic boundaries. This is because they have to synthesize intricate details about all objects in an image based on a text description. Therefore, }we present a technique for segmenting real and AI-generated images using latent diffusion models (LDMs) trained on internet-scale datasets. First, we show that the latent space of LDMs (z-space) is a better input representation compared to other feature representations like RGB images or CLIP encodings for text-based image segmentation. By training the segmentation models on the latent z-space, which creates a compressed representation across several domains like different forms of art, cartoons, illustrations, and photographs, we are also able to bridge the domain gap between real and AI-generated images. We show that the internal features of LDMs contain rich semantic information and present a technique in the form of LD-ZNet to further boost the performance of text-based segmentation. Overall, we show up to 6\% improvement over standard baselines for text-to-image segmentation on natural images. For AI-generated imagery, we show close to 20\% improvement compared to state-of-the-art techniques. \camready{The project is available at \href{https://koutilya-pnvr.github.io/LD-ZNet/}{https://koutilya-pnvr.github.io/LD-ZNet/}.}
\end{abstract}

\section{Introduction}
\label{sec:intro}
%first talk about how is language based image segmentation changing the world

%The use of language-based image segmentation has exploded in the past year due to its ability to provide fine-grained control to a user for image synthesis and editing tasks \cite{}. For example, given an image, the algorithm extracts a fine-grained mask associated to a user-provided text prompt. This mask is then used as input in several downstream editing applications like inpainting \cite{}, image to image translation \cite{}, background replacement \cite{} etc. Unlike the semantic and instance segmentation tasks that limit predictions to a predefined set of categories, language-based segmentation can segment regions in the image in an open-world setting and can generalize well to semantically similar concepts not encountered during training. %

Teaching neural networks to accurately find the boundaries of objects is hard and annotation of boundaries at internet scale is impractical. Also, most self-supervised or weakly supervised problems do not incentivize learning boundaries. For example, training on classification or captioning allows models to learn the most discriminative parts of the image without focusing on boundaries \cite{selvaraju2017grad, zeiler2014visualizing}. Our insight is that Latent Diffusion Models (LDMs) \cite{rombach2022high}, which can be trained without object level supervision at internet scale, must attend to object boundaries, and so we hypothesize that they can learn features which would be useful for open world image segmentation. We support this hypothesis by showing that LDMs can improve performance on this task by up to 6\%, compared to standard baselines and these gains are further amplified when LDM based segmentation models are applied on AI generated images.

\begin{figure}[!t]
    \centering
        \centering
        \includegraphics[width=\linewidth]{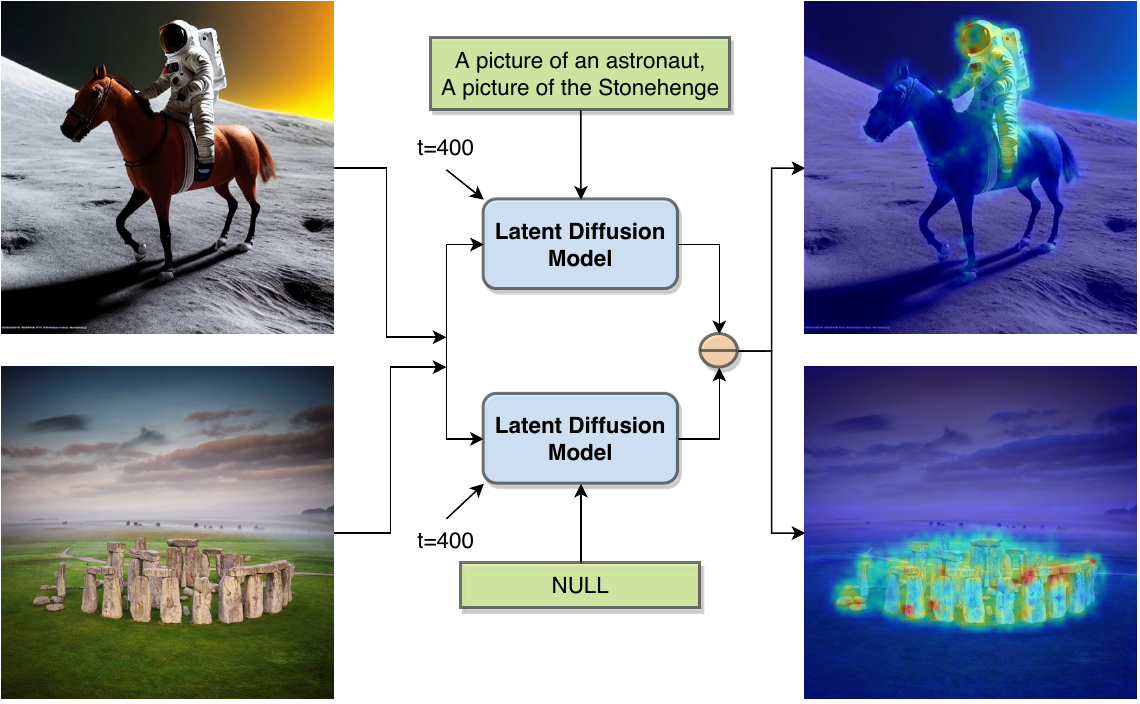}
        % \caption{Coarse segmentation from a pretrained LDM for two disctinct images, suggesting its internal features encode fine-grained object-level semantic information.}
        \caption{Coarse segmentation results from an LDM for two distinct images, demonstrating the encoding of fine-grained object-level semantic information within the model's internal features.}
        % {Given an image on the left, we compute the difference between the unconditional noise estimate from a pretrained LDM vs one conditioned on the text prompt. This difference represents a coarse segmentation of the astronaut being queried. These noise estimates are from an LDM which was not trained for the task of segmentation. The quality of the segmentation suggests that internal features of an LDM encodes fine-grained semantic information, that can be used for language-based segmentation.}
        \vspace{-1.5em}
        \label{fig:motivation1}
\end{figure}

\begin{figure*}[!t]
    \centering
        \centering
        \includegraphics[width=\linewidth]{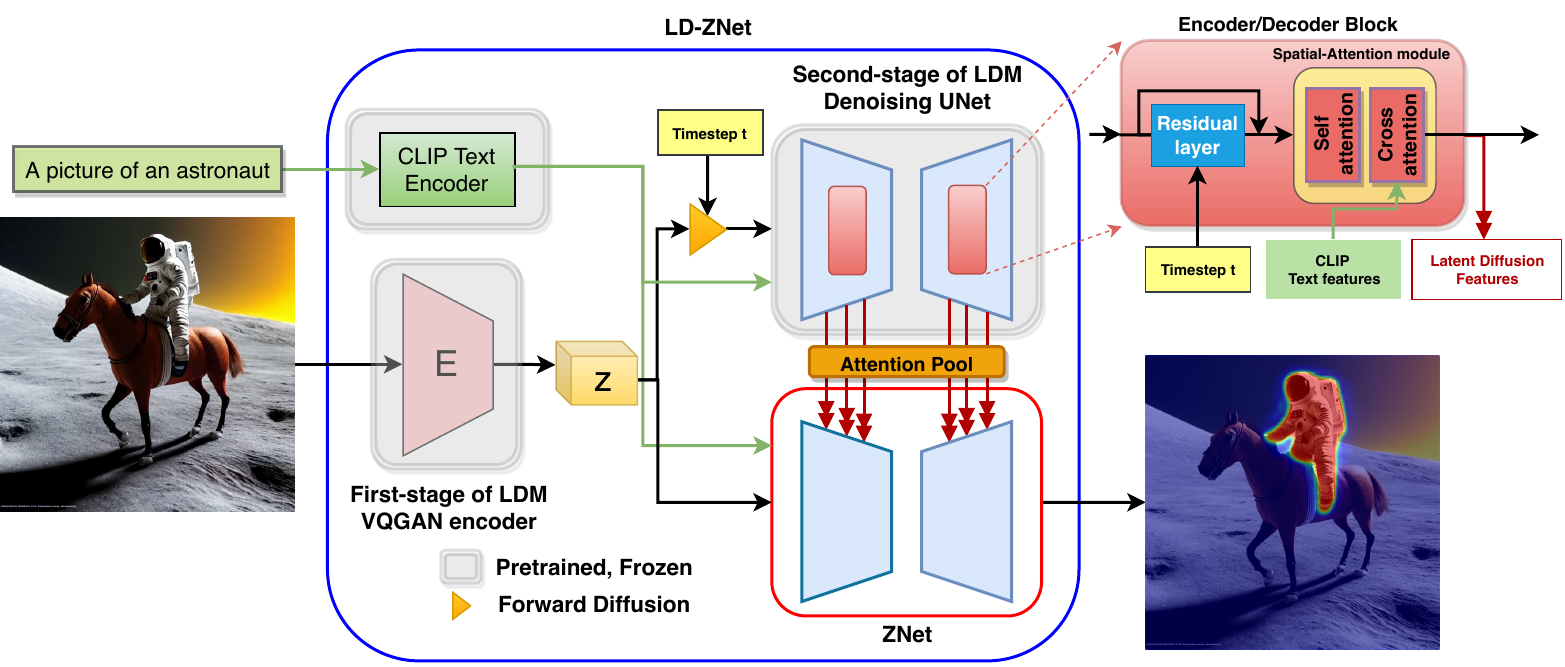}
        \caption{Overview of the proposed ZNet and LD-ZNet architectures. We propose to use the compressed latent representation $z$ as input for our segmentation network ZNet. Next, we propose LD-ZNet, which incorporates the latent diffusion features at various intermediate blocks from the LDM's denoising UNet, into ZNet.}
        \vspace{-1.5em}
        \label{fig:ldznet_summary}
\end{figure*}

To test the aforementioned hypothesis about the presence of object-level semantic information inside a pretrained LDM, we conduct a simple experiment. We compute the pixel-wise norm between the unconditional and text-conditional noise estimates from a pretrained LDM as part of the reverse diffusion process. This computation identifies the spatial locations that need to be modified for the noised input to align better with the corresponding text condition. Hence, the magnitude of the pixel-wise norm depicts regions that identify the text prompt. As shown in the Figure~\ref{fig:motivation1}, the pixel-wise norm represents a coarse segmentation of the subject although the LDM is not trained on this task. This clearly demonstrates that these large scale LDMs can not only generate visually pleasing images, but their internal representations encode fine-grained semantic information, that can be useful for tasks like segmentation.

\camready{Recently, text-based image segmentation has gained traction for creating and editing AI generated content (like AI art, illustrations, cartoons etc.) in image inpainting workflows \footnote{\href{https://github.com/brycedrennan/imaginAIry}{https://github.com/brycedrennan/imaginAIry}, \href{https://github.com/AUTOMATIC1111/stable-diffusion-webui}{https://github.com/AUTOMATIC1111/stable-diffusion-webui}} as it provides a conversational interface. Since the latent space $z$ \cite{esser2021taming}, extracted by a VQGAN is trained on several domains like art, cartoons, illustrations and real photographs, we posit that it is a more robust input representation for text-based segmentation on AI-generated images.} Furthermore, the internal layers of the LDM are responsible for generating the structure of the image and hence contain rich semantic information about objects. Soft masks from these layers have also been used as a latent input in recent work on image editing \cite{hertz2022prompt, brooks2022instructpix2pix}. Since this information is already present while generating the image, we propose an architecture in the form of LD-ZNet (shown in Figure~\ref{fig:ldznet_summary}) to decode it for obtaining the semantic boundaries of objects generated in the scene. Not only does our architecture benefit segmentation of objects in AI generated images, but it also improves performance over natural images. Overall our contributions are as follows: %
%1) We comprehensively evaluate the effectiveness of LDMs pretrained on large scale internet data for image segmentation based on text prompts. %
% \begin{itemize}
%     \item We propose a text-based segmentation architecture, ZNet that operates on the compressed latent space of the LDM ($z$).
%     \item Next, we study the internal representations at different stages of pretrained LDMs and show that they are useful for text-based image segmentation.
%     \item Finally, we propose a novel approach named LD-ZNet to incorporate the visual-linguistic latent diffusion features from a pretrained LDM and show improvements across several metrics and domains for text-based image segmentation.
% \end{itemize}
    \begin{itemize}
    \item We propose a text-based segmentation architecture, ZNet that operates on the compressed latent space of the LDM ($z$).
    \item Next, we study the internal representations at different stages of pretrained LDMs and show that they are useful for text-based image segmentation.
    \item Finally, we propose a novel approach named LD-ZNet to incorporate the visual-linguistic latent diffusion features from a pretrained LDM and show improvements across several metrics and domains for text-based image segmentation.
    \end{itemize}

\section{Related work}
\label{relatedworks}
\subsection{Text-based image segmentation}
Text-based image segmentation
is the general task of segmenting specific regions in an image, based on a text prompt. This is different from the referring expression segmentation (RES) task, which aims to extract instance-level segmentation of different objects through distinctive referring expressions. While RES helps applications in robotics that require localization of a {\em single} object in an image, text-based segmentation benefits image editing applications by being able to also segment 1) ``stuff" categories (clouds/ocean/beach \etc) and 2) multiple instances of an object category applicable to the text prompt. However, both these tasks have some shared literature in terms of approaches. Preliminary works \cite{hu2016segmentation, liu2017recurrent, shi2018key, li2018referring, ye2019cross} focused on the multi-modal feature fusion between the language and visual representations obtained from recurrent networks (such as LSTM) and CNNs respectively. The subsequent set of works \cite{margffoy2018dynamic, yu2018mattnet, wang2022cris, yang2022lavt} included variations of multi-modal training, attention and cross-attention networks etc. Recently, \cite{wang2022cris, luddecke2022image} used CLIP \cite{radford2021learning} to extract visual linguistic features of the image and the reference text separately. These features were then combined using a transformer based decoder to predict a binary mask. Alternately, \cite{kamath2021mdetr, zhang2022glipv2}, proposed vision-language pretraining on other text-based visual recognition tasks (object detection and phrase grounding) and later finetuned for the segmentation task. \camready{The concurrent works segment-anything (SAM) \cite{kirillov2023segment} and segment-everything-everywhere-all-at-once (SEEM) \cite{zou2023segment} allow interactive segmentation via point clicks, bounding boxes and text inputs \etc. demonstrating good zero-shot performance.} Different from all these works, we show the significance of using the latent space and the internal features from a pretrained latent diffusion model \cite{rombach2022high} for improving the more generic text-based image segmentation task.

%This task is different from the referring expression segmentation task, which is typically useful for robot localization, where the goal is to spatially localize a {\em unique object} with a distinctive referring expression (that can moreover contain complex positional references requiring dedicated training of the language encoder). This is necessary because for image editing applications, the algorithm should be able to localize ``stuff" categories like clouds/ocean/beach etc. and also be able to generate masks for multiple objects, if they are applicable to the text prompt.
\subsection{Text-to-Image synthesis}
Text-to-Image synthesis has initially been explored using GANs \cite{Xu_2018_CVPR, Zhu_2019_CVPR, tao2022df, zhang2021cross, ye2021improving, zhou2022towards} on publicly available image captioning datasets. Another line of work is by using autoregressive models \cite{ramesh2021zero, NEURIPS2021_a4d92e2c, gafni2022make} via a two stage approach. The first stage is a vector quantized autoencoder such as a VQVAE \cite{van2017neural, razavi2019generating} or a VQGAN \cite{esser2021taming} with an image reconstruction objective to convert an image into a shorter sequence of discrete tokens. This low dimensional latent space enables the training of compute intensive autoregressive models even for high resolution text-to-image synthesis. With the recent advancements in Diffusion Models (DM) \cite{nichol2021improved,NEURIPS2021_49ad23d1}, both in unconditional and class conditional settings, they have started gaining more traction compared to GANs. Their success in the text-to-image tasks \cite{saharia2022photorealistic,ramesh2022hierarchical} made them even more popular. However, the prior diffusion models worked in the high-dimensional image space that made training and inference computationally intensive. Subsequently, latent space representations \cite{nichol2021glide, gu2022vector, tang2022improved, rombach2022high} were proposed for high resolution text-to-image synthesis to reduce the heavy compute demands. More specifically, the latent diffusion model (LDM) \cite{rombach2022high} mitigates this problem by relying on a perceptually compressed latent space produced by a powerful autoencoder from the first stage. Moreover, they employ a convolutional backed UNet \cite{UNet} as the denoising architecture, allowing for different sized latent spaces as input. %
Recently this architecture is trained on large scale text-image data \cite{schuhmann2022laionb} from the internet and released as Stable-diffusion\footnote{\href{https://github.com/CompVis/stable-diffusion}{https://github.com/CompVis/stable-diffusion}}, which exhibited photo-realistic image generations. Subsequently, several language guided image editing applications such as inpainting \cite{couairon2022diffedit, Lugmayr_2022_CVPR, xie2022smartbrush}, text-guided image editing \cite{chen2018language, brooks2022instructpix2pix} became more popular and the usage for text-based image segmentation has surged, especially for AI generated images. We propose a solution for text-based image segmentation by leveraging the features which are already present as part of the synthesis process.

\begin{figure}[t]
    \centering
        \centering
        \includegraphics[width=\linewidth]{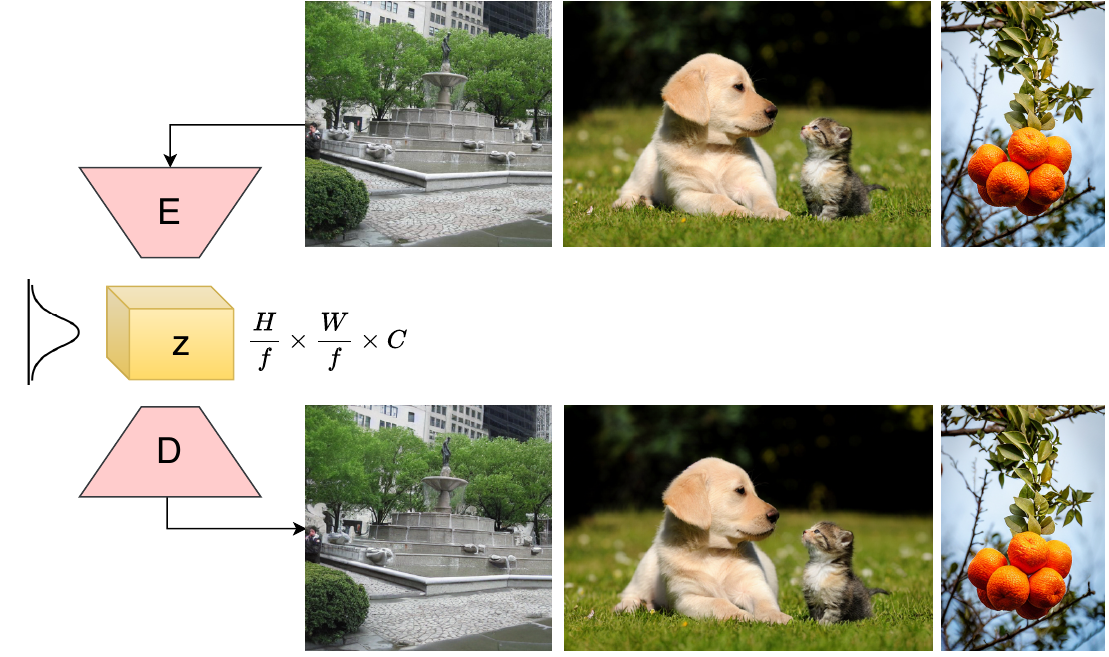}
        \caption{Reconstructions from the first stage of the LDM. Given an input image, the latent representation $z$ generated by the encoder, can be used to reconstruct images that are perceptually indistinguishable from the inputs. The high quality of these reconstructions suggests that the latent representation $z$, preserves most of the semantic information present in the input images.}
        \vspace{-1em}
        \label{fig:vqgan}
\end{figure}

\subsection{Semantics in generative models}
Semantics in generative models
such as GANs have been studied for binary segmentation~\cite{voynov2021object,melas2021finding} as well as multi-class segmentation~\cite{zhang2021datasetgan, tritrong2021repurposing, pakhomov2021segmentation} where the intermediate features have been shown to contain semantic information for these tasks. Moreover, \cite{semanticGAN} highlighted the practical advantages of these representations, such as out-of-distribution robustness. However, prior generative models (GANs \etc) as representation learners have received less attention compared to alternative unsupervised methods \cite{pmlr-v119-chen20j}, because of the training difficulties on complex, diverse and large scale datasets. Diffusion models \cite{nichol2021improved}, on the other hand are another class of powerful generative models that recently outperformed GANs on image synthesis \cite{NEURIPS2021_49ad23d1} and are able to train on large datasets such as Imagenet \cite{deng2009imagenet} or LAION \cite{schuhmann2022laionb}. In \cite{baranchuk2021label}, the authors demonstrated that the internal features of a pre-trained diffusion model were effective at the semantic segmentation task. However, this type of analysis \cite{zhang2021datasetgan, baranchuk2021label} has mostly been done in limited settings like few shot learning \cite{fei2006one} or limited domains like faces \cite{karras2019style}, horses \cite{yu15lsun} or cars \cite{yu15lsun}. Different from these works, we analyze the visual-linguistic semantic information present in the internal features of a text-to-image LDM \cite{rombach2022high} for text based image segmentation, which is an open world visual recognition task. %
Furthermore, we leverage these LDM features and show performance improvements when training with full datasets instead of few-shot settings.

%------------------------------------------------------------------------
% Method
\section{LDMs for Text-Based Segmentation}
\label{ldm analysis}
%reminder about how LDMs are awesome, where are they trained, brief overview of their architecture. Which part is GAN and which part is diffusion model.

%The latent diffusion architecture introduced in \cite{rombach2022high} is a diffusion model applied in the latent space of images generated using a powerful auto-encoder trained on a large dataset. These models are capable of generating photo-realistic images when conditioned on a variety of modalities such as text, class labels, semantic maps and for tasks such as inpainting and super-resolution. The latent diffusion architecture consists of two stages: \
The text-to-image latent diffusion architecture introduced in \cite{rombach2022high} consists of two stages:
%\vspace{-0.5em}
%\begin{enumerate} 
%\item 
1) An auto-encoder based VQGAN \cite{esser2021taming} that extracts a compressed latent representation ($z$) for a given image 
%\vspace{-0.5em}
%\item 
2) A diffusion UNet that is trained to denoise the noisy $z$ created in the forward diffusion process, conditioned on the text features. These text features are obtained from a pretrained frozen CLIP text encoder \cite{radford2021learning} and is conditioned at multiple layers of the UNet via cross-attention. 

In this paper, we show performance improvements on the text-based segmentation task in two steps. Firstly, we analyze the compressed latent space ($z$) from the first-stage and propose an approach named ZNet that uses $z$ as the visual input to estimate segmentation mask when conditioned on a text prompt. Secondly, we study the internal representations from the second stage of the stable-diffusion LDM for visual-linguistic semantic information and propose a way to utilize them inside ZNet for further improvements in the segmentation task. We name this approach as LD-ZNet.

% This stable diffusion model was trained for text-to-image synthesis on large scale datasets (laion-2B-en and laion-aesthetics v2.5+ \cite{schuhmann2022laionb}) using the text features obtained through a pre-trained CLIP network as an input at multiple layers of the UNet via cross-attention.

\subsection{ZNet: Leveraging Latent Space Features}
\label{sec:znet}

%Mention how you extract z features from transformer based VQ-GAN. Describe briefly the architecture of the GAN, what is the size of these features etc. which get extracted. Where was it trained? How do reverse decodings look like? Show that it is able to capture visual information in a compressed format by showing reconstructions of a few images. Hence this is a good feature
%The first stage of the LDM extracts a latent representation ($z$) of the image using an auto-encoder. 

We observe that the latent space ($z$) from the first-stage of the LDM is a compressed representation of the image that preserves semantic information, as depicted in Figure \ref{fig:vqgan}. The VQGAN in the first-stage achieves such semantic-preserving compression with the help of large scale training data as well as a combination of losses - perceptual loss \cite{zhang2018unreasonable}, a patch-based \cite{isola2017image} adversarial objective \cite{dosovitskiy2016generating, esser2021taming, yu2021vector}, and a KL-regularization loss.

% {\color{blue} For our ZNet experiments we use the auto-encoder layers of the LDM without the diffusion layers.}
% This auto-encoder consists of a sequence of ResNet blocks and strided-convolutions, generating a compressed feature representation $z$, which is used as an input for the next stage. This feature representation $z$, is then transformed back into the image domain using a decoder that consists of another sequence of ResNet blocks and deconvolutions. This latent space is obtained through perceptual loss based compression \cite{zhang2018unreasonable}, a patch-based \cite{isola2017image} adversarial objective \cite{dosovitskiy2016generating, esser2021taming, yu2021vector}, and a KL-regularization loss. %Using a combination of these losses, LDMs are able to generate high quality images without compromising on the finer-details, while also having relatively low compute requirements. 
% When trained on a large-scale dataset in conjunction with the aforementioned losses, the auto-encoder, despite having a high-compression rate, retains most of the information present in the input image as shown in Figure \ref{fig:vqgan} {\color{blue} while having relatively low compute}. %We first evaluate the effectiveness of the z-space instead of RGB images as an input for visual recognition tasks. %We first propose a baseline with the z-space as input for the referring image segmentation task and show improvements on incorporating the LDM features from the second stage. 

In our experiments, we observe that this compressed latent representation $z$ is more robust compared to the original image in terms of their association with the text prompts. We believe this is because $z$ is a $\frac{H}{8} \times  \frac{W}{8} \times 4$ dimensional feature with 48 $\times$ fewer elements compared to the original image, while preserving the semantic information. Several prior works \cite{turk1991eigenfaces,ke2004pca,de2001robust}, show that compression techniques like PCA, which create information preserving lower dimensional representations generalize better. Therefore, we propose using the $z$ representation along with the frozen CLIP text features \cite{radford2021learning} as an input to our segmentation network. Furthermore, because the VQGAN is trained across several domains like art, cartoons, illustrations, portraits, etc., it learns a robust and compact representation which generalizes better across domains, as can be seen in our experiments on AI generated images. We call this approach ZNet. The architecture of ZNet is shown in the bottom box of Figure \ref{fig:ldznet_summary}, and is the same as the denoising UNet module of the LDM. We therefore initialize it with pretrained weights of the second-stage of the LDM.

\subsection{LD-ZNet: Leveraging Diffusion Features}
\label{sec:caznet}
%Describe briefly the architecture of LDMs, where and how was it trained? How does language play a role in generating these features? What is the size of the features which get extracted. What are the timesteps which were used to extract the features? 

Given a text prompt and a timestep $t$, the second-stage of the LDM is trained to denoise $z_t$ - a noisy version of the latent representation $z$ obtained via forward diffusion process for $t$ timesteps. A UNet architecture is used whose encoder/decoder elements are shown in Figure \ref{fig:ldznet_summary} (top right). A typical encoder/decoder block contains a residual layer followed by a spatial-attention module that internally has self-attention and then cross-attention with the text features. We analyze the semantic information in the internal visual-linguistic representations developed at different blocks of encoder and decoder right after these spatial-attention modules. We also propose a way to utilize these latent diffusion features using cross-attention into the ZNet segmentation network and we call the final model as LD-ZNet.

% For LD-ZNet we use second stage of the LDM which is a diffusion model applied in the latent space of the first stage auto-encoder described in section~\ref{sec:znet}. Diffusion models \cite{pmlr-v37-sohl-dickstein15} learn a data distribution by gradually denoising a normally distributed variable. This variable is typically obtained from the forward diffusion process (fixed markov chain) over specific number of timesteps, where based on the timestep parameter, a certain amount of noise is added to the latent representation $z$. This noisy $z$ is input into the reverse diffusion UNet along with the timestep and the conditional text information. Using these inputs, the UNet predicts a noise estimate that mimics the ground truth noise added to $z$ in the forward diffusion process. The typical encoder and decoder elements of the UNet contain ResNet followed by spatial-attention layers which internally employ self-attention and cross-attention mechanisms with the text features as shown in Figure \ref{fig:unet}. We analyze the semantic information present in the internal visual-linguistic representations developed at different blocks right after these spatial-attention layers. We also propose a way to utilize these features using cross-attention for downstream visual recognition and we call the final model as LD-$Z$Net.

\subsubsection{Visual-Linguistic Information in LDM Features}
\label{ldm analysis}

\begin{figure}[t]
    \centering
    \includegraphics[width=\linewidth]{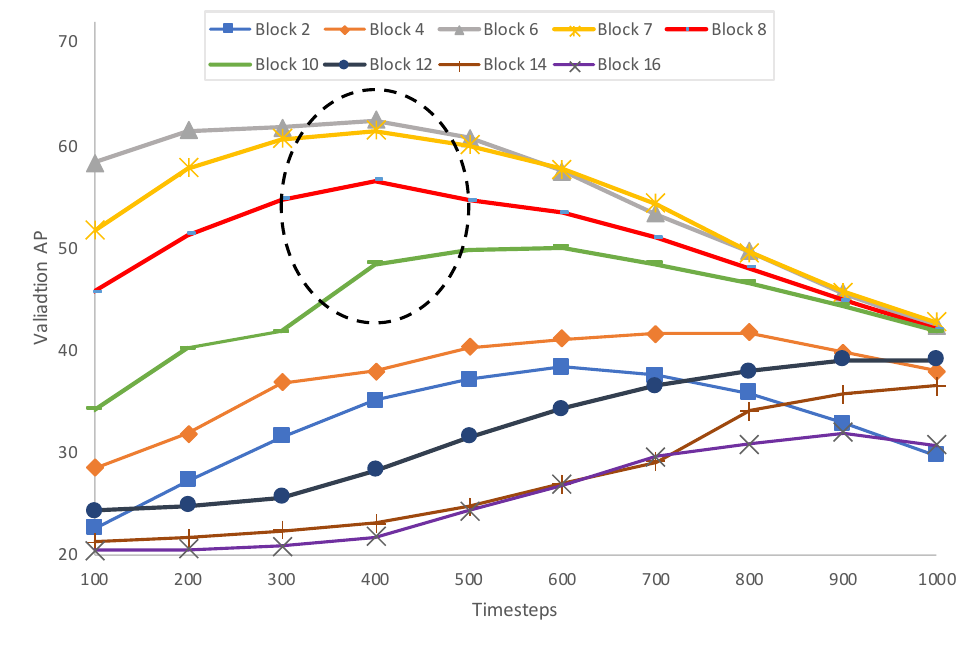}
    \caption{Semantic information present in the LDM features at various blocks and timesteps for the referring image segmentation task. AP is measured on a small validation subset of the PhraseCut dataset.}
    \vspace{-1em}
    \label{fig:ldm_analysis}
\end{figure}

% In order to decide which features from the UNet to use, we first 
We evaluate the semantic information present in the pretrained LDM at various blocks and timesteps for the text-based image segmentation task. In this experiment, we consider the latent diffusion features right after the spatial-attention layers 1-16 spanning across all the encoder and decoder blocks present in the UNet. At each block, we analyze the features for every $100^{th}$ timestep in the range $[100, 1000]$. We use a small subset of the training and validation sets from the Phrasecut dataset and train a simple decoder on top of these features to predict the associated binary mask. Specifically, given an image $I$ and timestep $t$, we first extract its latent representation $z$ from the first stage of LDM and add noise from the forward diffusion to obtain $z_t$ for a timestep $t$. Next we extract the frozen CLIP text features for the text prompt and input both of them into the denoising UNet of the LDM to extract the internal visual-linguistic features at all the blocks for that timestep. We use these representations to train the corresponding decoders until convergence. Finally, we evaluate the AP metric on a small subset of the validation dataset. The performance of features from different blocks and timesteps is shown in Figure \ref{fig:ldm_analysis}. 

Similar to \cite{baranchuk2021label}, we observe that the middle blocks \{6,7,8,9,10\} of the UNet contain more semantic information compared to either the early blocks of the encoder or the later blocks of the decoder. We also observe that the timesteps 300-500 contain the maximum visual-linguistic semantic information compared to other timesteps, for these middle blocks. This is in contrast to the findings of \cite{baranchuk2021label} that report the timesteps \{50, 150, 250\} to contain the most useful information when evaluated on an unconditional DDPM model for the few shot semantic segmentation task for horses \cite{yu15lsun} and faces\cite{karras2019style}. We believe that the reason for this difference is because, in our case, the image synthesis is guided by text, leading to the emergence of semantic information earlier in the reverse diffusion process (t=1000$\rightarrow$0), in contrast to unconditional image synthesis.

%\section{Leveraging LDMs for text-based Visual Recognition}
%\label{sec:method}

% \begin{figure}[!t]
%     \centering
%     \includegraphics[width=\linewidth]{Images/ZSEG}
%     \caption{Overview of the LD-$Z$Net to utilize the visual-linguistic features from LDM (top box) into ZNet (bottom box), for referring image segmentation.}
%     \vspace{-1em}
%     \label{fig:architectural_details}
% \end{figure}

%\vspace{-2em}
\subsubsection{LD-ZNet Architecture}
%describe how we combine large scale LDMs with these existing architectures. Concat vs cross attention? Where to combine and how?
% \begin{figure*}
%     \centering
%     \begin{subfigure}{0.6\linewidth}
%         \centering
%         \includegraphics[width=\linewidth]{Images/ZSEG}
%         \caption{}
%         \label{fig:architectural_details}
%     \end{subfigure}%
%     \begin{subfigure}{0.4\linewidth}
%         \centering
%         \includegraphics[width=\linewidth]{Images/AttentionPool.pdf}
%         \caption{}
%         \label{fig:spatial-attention}
%     \end{subfigure}
%     \caption{(a) Overview of the proposed method to utilize the visual-linguistic features from LDM (top box) into ZNet (bottom box), for referring image segmentation. (b) We propose to incorporate the visual-linguistic representations from LDM obtained at the spatial-attention layers via a cross-attention mechanism into the corresponding spatial-attention layers of the CA-ZNet decoder through an Attention Pool layer.}
%     \label{fig:CA-ZNET}
% \end{figure*}

\begin{figure}[!t]
    \centering
    \includegraphics[width=\linewidth]{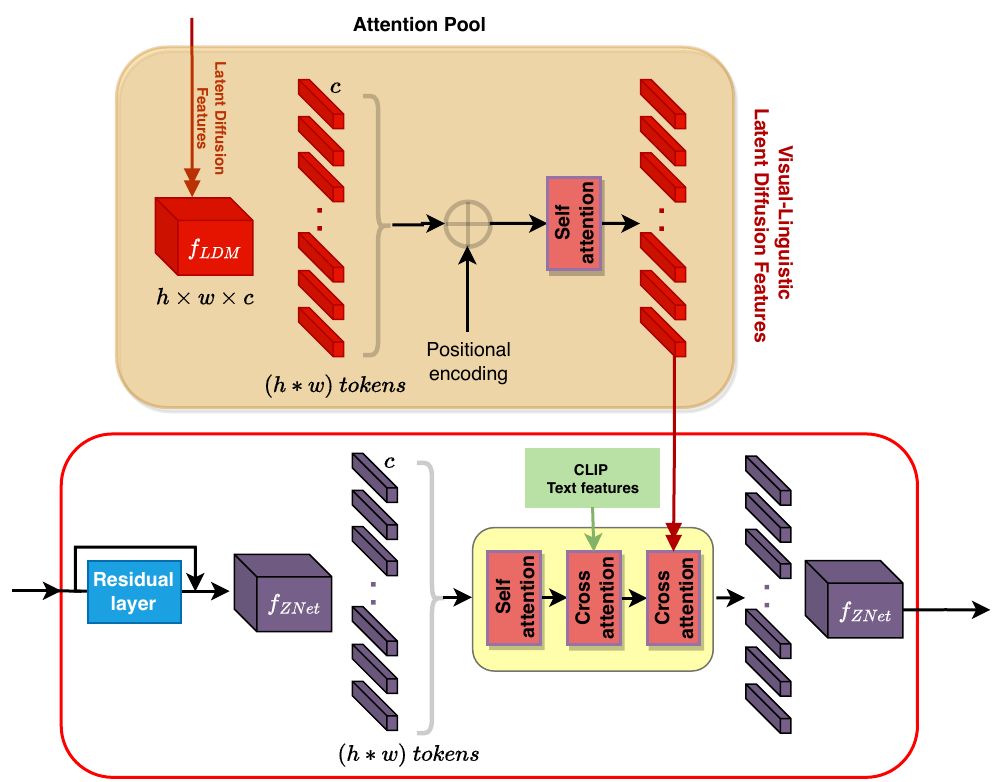}
    \caption{We propose to incorporate the visual-linguistic representations from LDM obtained at the spatial-attention modules via a cross-attention mechanism into the corresponding spatial-attention modules of the ZNet through an \emph{attention pool} layer.}
    \vspace{-1em}
    \label{fig:spatial-attention}
\end{figure}

We propose using the aforementioned visual-linguistic representations at multiple spatial-attention modules of the pretrained LDM into the ZNet as shown in Figure \ref{fig:ldznet_summary}. These latent diffusion features are injected into the ZNet via a cross-attention mechanism at the corresponding spatial-attention modules as shown in Figure \ref{fig:spatial-attention}. This allows for an interaction between the visual-linguistic representations from the ZNet and the LDM. Specifically, we pass the latent diffusion features through an \emph{attention pool} layer that not only acts as a learnable layer to match the range of the features participating in the cross-attention, but also adds a positional encoding to the pixels in the LDM representations. The outputs from the attention pool are now positional-encoded visual-linguistic representations that enable the proposed cross-attention mechanism to attend to the corresponding pixels from the ZNet features. ZNet when augmented with these latent diffusion features from the LDM (through cross-attention) is referred to as LD-ZNet.

Following the semantic analysis of latent diffusion features (Sec. \ref{ldm analysis}), we incorporate the internal features from blocks \{6,7,8,9,10\} of the LDM into the corresponding blocks of ZNet, in order to make use of the maximum semantic and diverse visual-linguistic information from the LDM. For AI generated images, these blocks are anyways responsible to generate the final image and using LD-ZNet, we are able to tap into this information which can be used for segmenting objects in the scene. 

% Specifically, we extract the representations from blocks [6,7,8,9,10] of the LDM at a random timestep between 300-500 and feed them into the corresponding block of the decoder UNet.

%------------------------------------------------------------------------
% Experiments
\section{Experiments}
\label{experiments}

%\subsection{Implementation details}
\textbf{Implementation details:}
In this paper, we use the stable-diffusion v1.4 checkpoint as our LDM that internally uses the frozen ViT-L/14 CLIP text encoder \cite{radford2021learning}. We implement the above described ZNet and LD-ZNet in pytorch inside the stable-diffusion library.  We also initialize our networks with the weights from the LDM wherever possible, while initializing the remaining parameters from a normal distribution. We train ZNet and LD-ZNet on 8 NVIDIA A100 gpus with a batch size of 4 using the Adam optimizer and a base learning rate of 5$e^{-7}$ per mini-batch sample, per gpu. For all our experiments, we keep the text encoder frozen and use an image resolution of 384 for a fair comparison with the previous works.

%\subsection{Datasets}
\textbf{Datasets:}
We use Phrasecut \cite{wu2020phrasecut}, which is currently the largest dataset for the \textit{text-based image segmentation} task, with nearly 340K phrases along with corresponding segmentation masks that not only permit annotations for stuff classes but also accommodate multiple instances. Following \cite{radford2021learning}, we randomly augment the phrases from a fixed set of prefixes. For the images, we randomly crop a square around the object of interest with maximum area, ensuring that the object remains at least partially visible. We avoid negative samples to remove ambiguity in the LDM features for non-existent objects.

We create a dataset consisting of AI-generated images which we name \textbf{AIGI} dataset, to showcase the usefulness of our approach for text-based segmentation on a different domain. We use 100 AI-generated images from {\em lexica.art} and manually annotated multiple regions for 214 text-prompts relevant to these images. 

We also use the popular referring expression segmentation datasets namely RefCOCO \cite{kazemzadeh2014referitgame}, RefCOCO+ \cite{kazemzadeh2014referitgame} and G-Ref \cite{nagaraja16refexp} to demonstrate the generalization abilities of ZNet and LD-ZNet. In RefCOCO, each image contains two or more objects and each expression has an average length of 3.6 words. RefCOCO+ is derived from RefCOCO by excluding certain absolute-location words and focuses on purely appearance based descriptions. For example it uses ``the man in the yellow polka-dotted shirt” rather than ``the second man from the left" which makes it more challenging. Unlike RefCOCO and RefCOCO+, the average length of sentences in G-Ref is 8.4 words, which have more words about locations and appearances. While we adopt the UNC partition for RefCOCO and RefCOCO+ in this paper, we use the UMD partition for G-Ref.

%\subsection{Metrics} 
\textbf{Metrics:}
We follow the evaluation methodology of \cite{luddecke2022image} and report best foreground IoU ($IoU_{FG}$) for the foreground pixels, the best mean IoU of all pixels (mIoU), and the Average Precision (AP).

%------------------------------------------------------------------------
% Results
\section{Results}
\label{results}

\subsection{Image Segmentation Using Text Prompts}
\begin{table}[t]
    \centering
    \begin{adjustbox}{width=0.9\columnwidth}
    \begin{tabular}{|c||c|c|c|c|}
    \hline
    Method & mIoU & $IoU_{FG}$ & AP \\
    \hline
        \hline
        MDETR \cite{kamath2021mdetr} & 53.7 & - & - \\
        GLIPv2-T \cite{zhang2022glipv2} & 59.4 & - & - \\
        \hline\hline
        RMI \cite{wu2020phrasecut} & 21.1 & 42.5 & - \\
        Mask-RCNN Top \cite{wu2020phrasecut} & 39.4 & 47.4 & -\\
        HulaNet \cite{wu2020phrasecut} & 41.3 & 50.8 & - \\
        CLIPSeg (PC+)  \cite{luddecke2022image}& 43.4 & 54.7 & 76.7\\
        CLIPSeg (PC, D=128) \cite{luddecke2022image} & 48.2 & 56.5 & 78.2\\
        \hline
        RGBNet & 46.7 & 56.2 & 77.2 \\
        \rowcolor{lightgray} ZNet (Ours) & 51.3 & 59.0 & 78.7 \\
        \rowcolor{lightgray} LD-ZNet (Ours) & \textbf{52.7} & \textbf{60.0} & \textbf{78.9} \\
        % \rowcolor{lightgray} CLIP Image features only &  & 49.04 & 57.76 & 77.51 \\
        % \rowcolor{lightgray} Z + CLIP &  & 49.8 & 58.8 & 79.8 \\
        % \rowcolor{lightgray} Z + CLIP Img + LDM features&  &  &  &  \\
    \hline
    \end{tabular}
    \end{adjustbox}
    \caption{Text-based image segmentation performance on the PhraseCut testset. The performance of ZNet and LD-ZNet is highlighted in gray. Both these models outperform the baseline RGBNet on all the metrics.}
    \vspace{-1em}
    \label{tab:ris_results}
\end{table}
\begin{figure}[t]
    \centering
    \begin{subfigure}[t]{0.19\columnwidth}
        \centering
        \includegraphics[width=\linewidth]{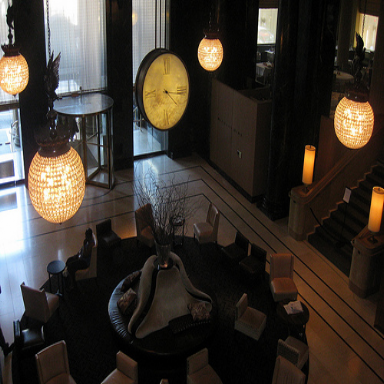}
        \includegraphics[width=\linewidth]{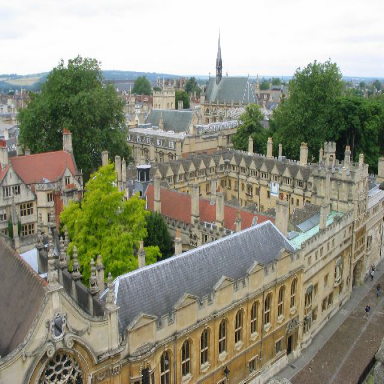}
        \caption{Input}
        % \label{fig:architecture}
    \end{subfigure}%
    \hspace{0.05cm}%
    \begin{subfigure}[t]{0.19\columnwidth}
        \centering
        \includegraphics[width=\linewidth]{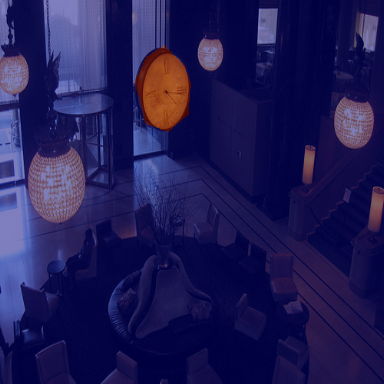}
        \includegraphics[width=\linewidth]{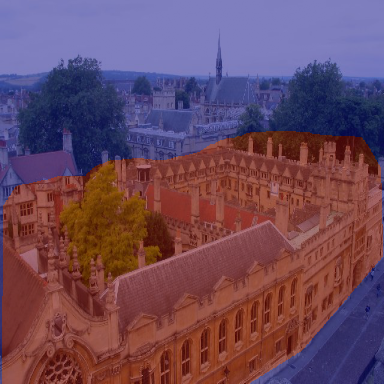}
        \caption{GT mask}
        % \label{fig:architecture}
    \end{subfigure}%
    \hspace{0.05cm}%
    \begin{subfigure}[t]{0.19\columnwidth}
        \centering
        \includegraphics[width=\linewidth]{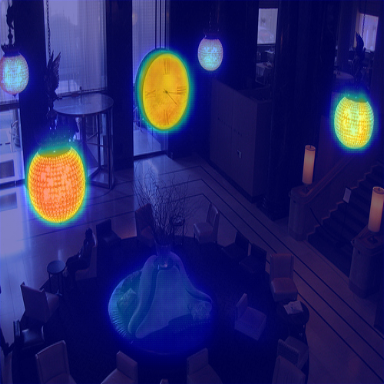}
        \includegraphics[width=\linewidth]{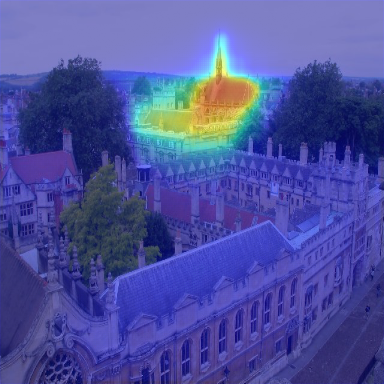}
        \caption{RGBNet}
        % \label{fig:architecture}
    \end{subfigure}%
    \hspace{0.05cm}%
    \begin{subfigure}[t]{0.19\columnwidth}
        \centering
        \includegraphics[width=\linewidth]{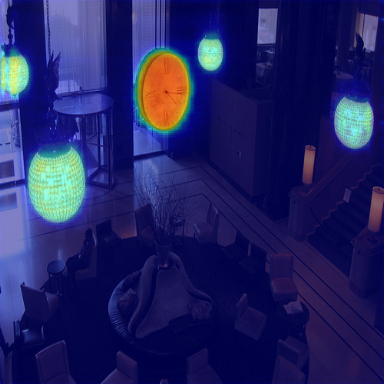}
        \includegraphics[width=\linewidth]{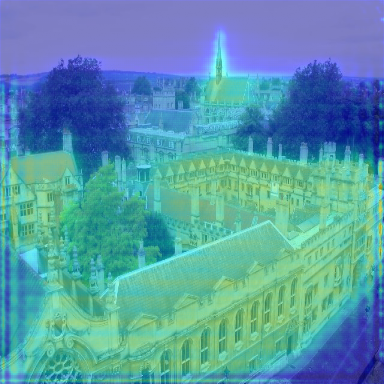}
        \caption{ZNet}
        % \label{fig:architecture}
    \end{subfigure}%
    \hspace{0.05cm}%
    \begin{subfigure}[t]{0.19\columnwidth}
        \centering
        \includegraphics[width=\linewidth]{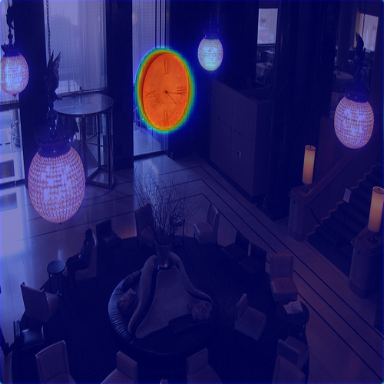}
        \includegraphics[width=\linewidth]{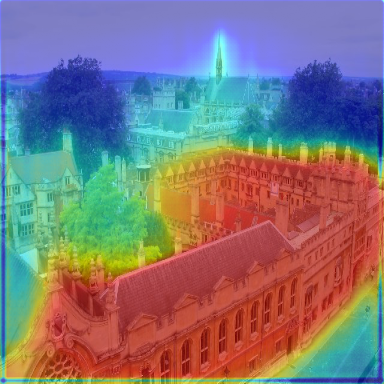}
        \caption{LD-ZNet}
        % \label{fig:architecture}
    \end{subfigure}
    
    \caption{Qualitative comparison on the PhraseCut test set. Each row contains an input image with a text prompt as an input, with the goal being to segment the image regions corresponding to the reference text. The text prompts are \emph{``hanging clock"} and \emph{``castle"} for the top and bottom rows. We show improvements using ZNet and LD-ZNet compared to the RGBNet.}
    \vspace{-1em}
    \label{fig:visual_results}
\end{figure}

% \textcolor{red}{define each approach z baseline, RGB baseline etc. separately, so that the text reads smoother}

On the PhraseCut dataset, we compare the performance of previous approaches with our ZNet and LD-ZNet for the text-based image segmentation task (Table \ref{tab:ris_results}). In order to showcase the performance improvement of our proposed networks, we create a baseline named RGBNet with the same architecture as ZNet except we use the original images as the input instead of its latent space $z$. For RGBNet, we use additional learnable convolutional layers to map the original image to match the input resolution of ZNet. From Table \ref{tab:ris_results}, we observe that our ZNet and LD-ZNet significantly outperform RGBNet. Specifically, the performance improvement from using the latent representation $z$ over the original images is clear (i.e. ZNet vs RGBNet baseline). Performance further improves upon incorporating the LDM visual-linguistic representations (LD-ZNet) - by 6\% overall on the $mIoU$ metric compared to RGBNet. We also highlight this qualitatively in Figure~\ref{fig:visual_results}. In the figure, we show the original image and the GT mask along with outputs from the RGBNet baseline followed by ZNet and LD-ZNet, where both ZNet and LD-ZNet help improve results consistently. For example in the top row, RGBNet detects light fixtures for the ``hanging clock" prompt, and although ZNet does not have as strong activations for these incorrect detections, it is LD-ZNet that correctly segments the ``clock". Similarly in the bottom row, while RGBNet completely got the ``castle" wrong, ZNet correctly has activations on the right buildings, but with lower confidence. However, LD-ZNet improves it further. 

We outperform in all the metrics when compared to previous works, other than MDETR \cite{kamath2021mdetr} and GLIPv2 \cite{zhang2022glipv2}. Notably, these works are pre-trained on detection and phrase grounding for predicting bounding boxes on huge corpus of text-image pairs across various publicly available datasets with bounding box annotations and are later fine-tuned on the Phrasecut dataset for the segmentation task. However, our work is orthogonally focused towards exploring and utilizing LDMs and its internal features for improving the text-based segmentation performance. Note that object detection datasets  have a good overlap with the visual content in PhraseCut, however, they are not representative of the diversity in images available on the internet. For example, while they could learn common concepts like sky, ocean, chair, table and their synonyms, methods like MDETR would not understand concepts like Mikey Mouse, Pikachu etc., which we will show in Section~\ref{discussion}.

\subsection{Generalization to AI Generated Images}

% \begin{figure}
%     \centering
%     \includegraphics[width=0.9\columnwidth]{Images/AI_generated_stats}
%     \caption{Generalization of the proposed LD-ZNet on our AI-generated dataset when compared with other state-of-the-art text-based segmentation methods - CLIPSeg (PC+) and MDETR.}
%     \vspace{-1em}
%     \label{fig:ai_generated_chart}
% \end{figure} 

\begin{table}[t]
    \centering
    \begin{tabular}{|c||c|c|}
    \hline
         Method & mIoU & AP\\
         \hline
         MDETR \cite{kamath2021mdetr} & 53.4 & 63.8 \\
         CLIPSeg (PC+) \cite{luddecke2022image} & 56.4 & 79.0 \\
         SEEM \cite{zou2023segment} & 57.4 & 70.0\\
         \hline
         RGBNet & 63.4 & 84.1\\
         \rowcolor{lightgray} ZNet (Ours) & 68.4 & 85.0\\
         \rowcolor{lightgray} LD-ZNet (Ours) & \textbf{74.1} & \textbf{89.6}\\
    \end{tabular}
    \caption{Generalization of the proposed LD-ZNet on our AIGI dataset when compared with other state-of-the-art text-based segmentation methods.}
    \label{tab:ai_generated_chart}
\end{table}

\begin{figure} [!t]
    \centering
    \begin{subfigure}[t]{0.19\columnwidth}
        \centering
        \includegraphics[width=\linewidth]{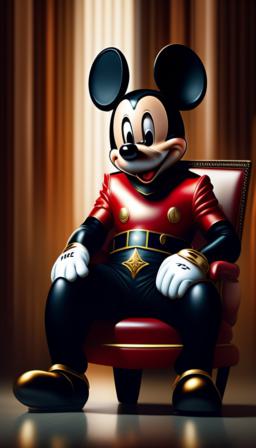}
        \includegraphics[width=\linewidth]{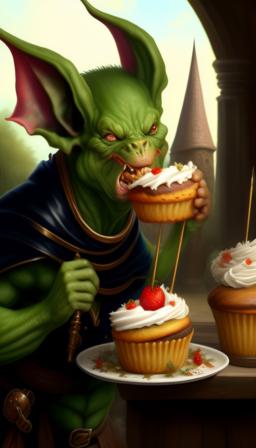}
        \includegraphics[width=\linewidth]{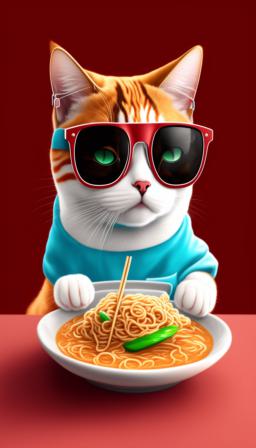}
        \includegraphics[width=\linewidth]{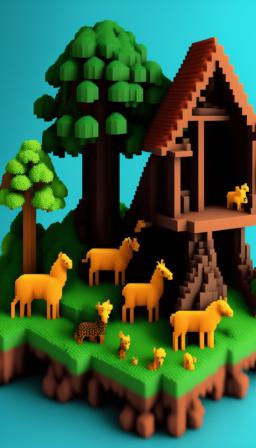}
        \caption*{Input}
    \end{subfigure}%
    \hspace{0.05cm}%
    \begin{subfigure}[t]{0.19\columnwidth}
        \centering
        \includegraphics[width=\linewidth]{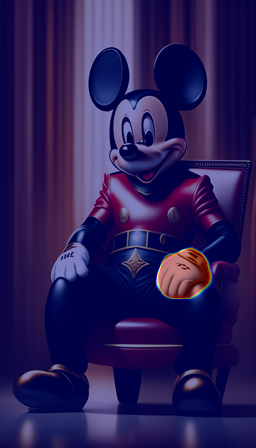}
        \includegraphics[width=\linewidth]{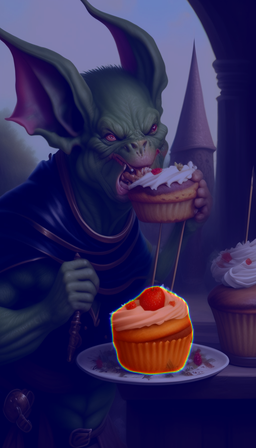}
        \includegraphics[width=\linewidth]{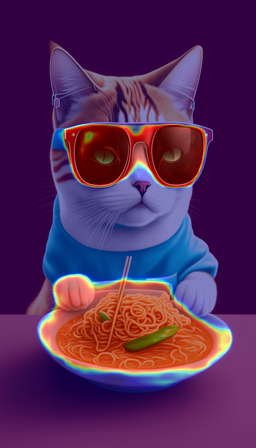}
        \includegraphics[width=\linewidth]{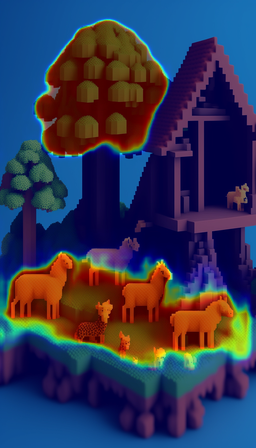}
        \caption*{MDETR \cite{kamath2021mdetr}}
    \end{subfigure}%
    \hspace{0.05cm}%
    \begin{subfigure}[t]{0.19\columnwidth}
        \centering
        \includegraphics[width=\linewidth]{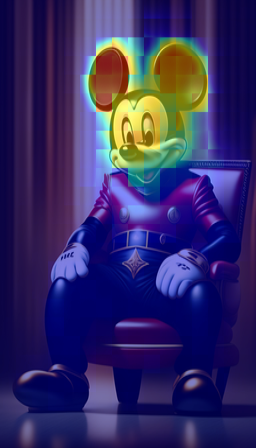}
        \includegraphics[width=\linewidth]{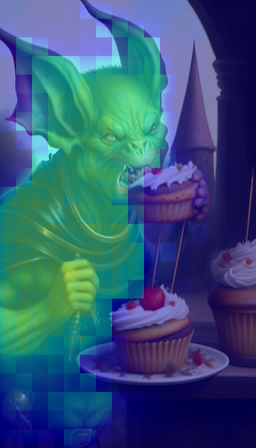}
        \includegraphics[width=\linewidth]{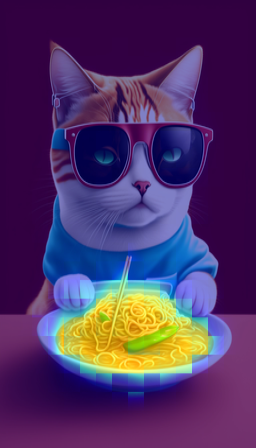}
        \includegraphics[width=\linewidth]{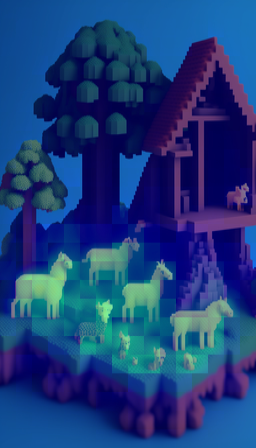}
        \caption*{CLIPSeg \cite{luddecke2022image}}
    \end{subfigure}%
    \hspace{0.05cm}%
    \begin{subfigure}[t]{0.19\columnwidth}
        \centering
        \includegraphics[width=\linewidth]{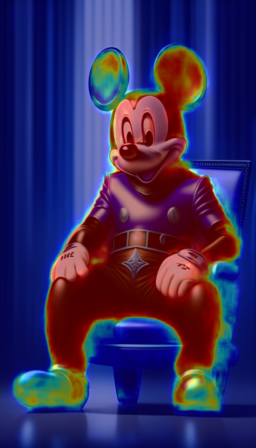}
        \includegraphics[width=\linewidth]{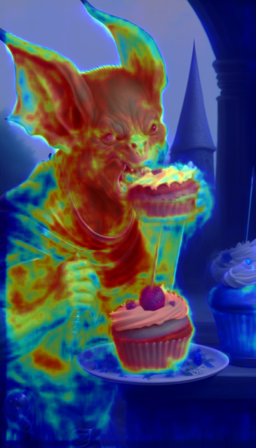}
        \includegraphics[width=\linewidth]{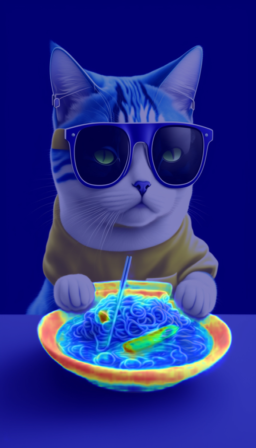}
        \includegraphics[width=\linewidth]{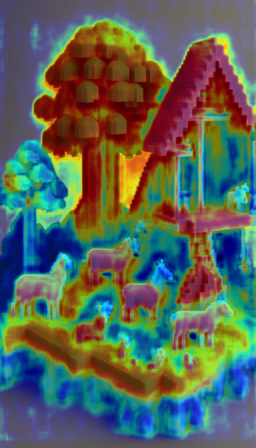}
        \caption*{SEEM \cite{zou2023segment}}
    \end{subfigure}%
    \hspace{0.05cm}%
    \begin{subfigure}[t]{0.19\columnwidth}
        \centering
        \includegraphics[width=\linewidth]{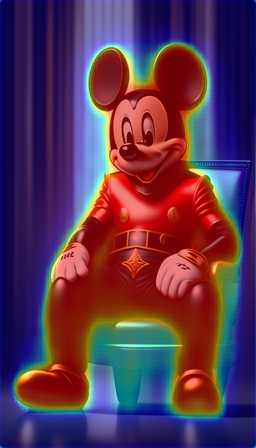}
        \includegraphics[width=\linewidth]{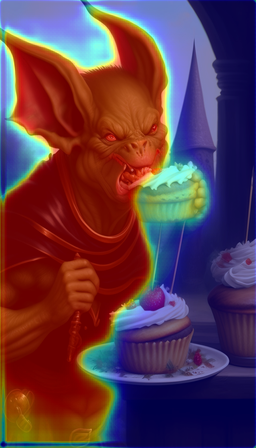}
        \includegraphics[width=\linewidth]{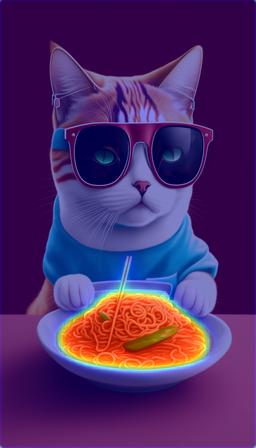}
        \includegraphics[width=\linewidth]{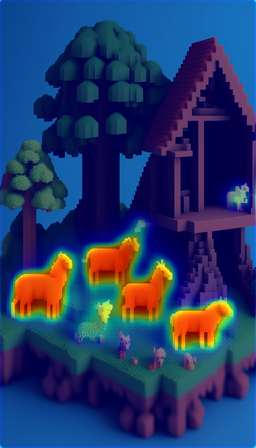}
        \caption*{LD-ZNet}
    \end{subfigure}
    
    \caption{Qualitative comparison on the AI-generated images for text-based segmentation. The text prompts are \emph{``Mickey mouse"}, \emph{``Goblin"}, \emph{``Ramen"} and \emph{``animals"} respectively.}
    \vspace{-1em}
    \label{fig:ai_generated_qualitative}
\end{figure}

\begin{figure}[!t]
    \captionsetup[subfigure]{labelformat=empty,font=small,labelfont={bf,sf}}
    \centering
    \begin{subfigure}{0.24\columnwidth}
        \centering
        \begin{subfigure}{\columnwidth}
            \centering
            \caption{``Hoodie"}
            \includegraphics[width=\columnwidth]{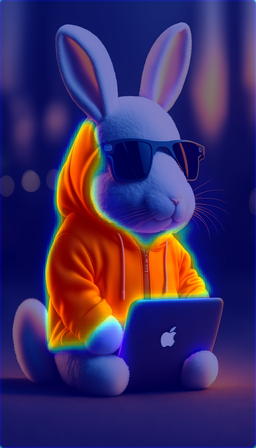}
        \end{subfigure}
        \begin{subfigure}{\columnwidth}
            \centering
            \caption{``Spiderman"}
            \includegraphics[width=\columnwidth]{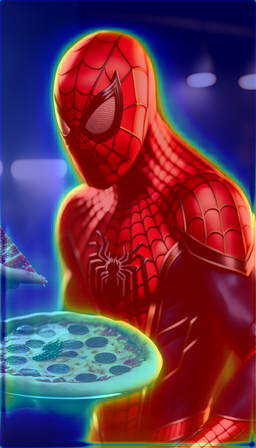}
        \end{subfigure}
    \end{subfigure}
    \hspace{0.05cm}%
    \begin{subfigure}{0.24\columnwidth}
        \centering
        \begin{subfigure}{\columnwidth}
            \centering
            \caption{``Owl"}
            \includegraphics[width=\columnwidth]{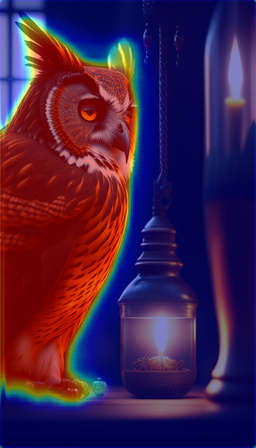}
        \end{subfigure}
        \begin{subfigure}{\columnwidth}
            \centering
            \caption{``Trump"}
            \includegraphics[width=\columnwidth]{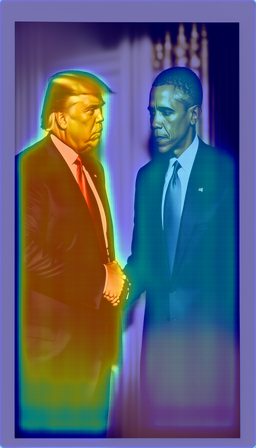}
        \end{subfigure}
    \end{subfigure}%
    \hspace{0.05cm}%
    \begin{subfigure}{0.24\columnwidth}
        \centering
        \begin{subfigure}{\columnwidth}
            \centering
            \caption{``Pikachu"}
            \includegraphics[width=\columnwidth]{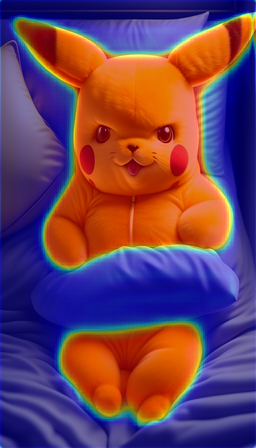}
        \end{subfigure}
        \begin{subfigure}{\columnwidth}
            \centering
            \caption{``Joker"}
            \includegraphics[width=\columnwidth]{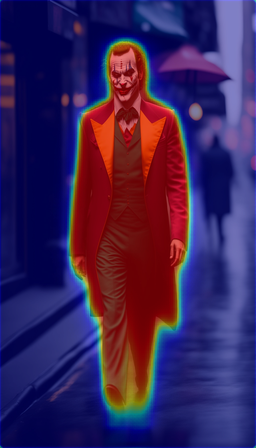}
        \end{subfigure}
    \end{subfigure}%
    \hspace{0.05cm}%
    \begin{subfigure}{0.24\columnwidth}
        \centering
        \begin{subfigure}{\columnwidth}
            \centering
            \caption{``Godzilla"}
            \includegraphics[width=\columnwidth]{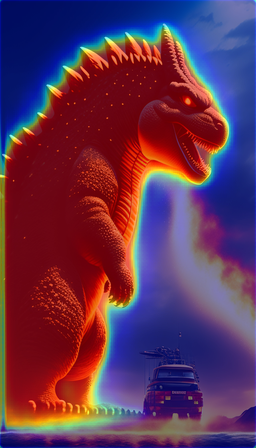}
        \end{subfigure}
        \begin{subfigure}{\columnwidth}
            \centering
            \caption{``Eiffel"}
            \includegraphics[width=\columnwidth]{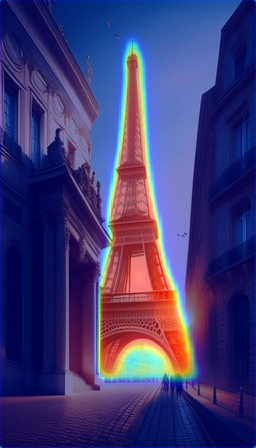}
        \end{subfigure}
    \end{subfigure}
    \caption{More qualitative results of LD-ZNet from AIGI dataset.}
    \label{fig:ldznet_ai_generated}
    \vspace{-1em}
\end{figure}

% We also study the generalization ability of our proposed method to AI-generated images
With the growing popularity of AI generated images, text-based image segmentation is extensively being used by content creators in their daily workflows. Many public libraries \footnote{\href{https://github.com/brycedrennan/imaginAIry}{https://github.com/brycedrennan/imaginAIry}, \href{https://github.com/AUTOMATIC1111/stable-diffusion-webui}{https://github.com/AUTOMATIC1111/stable-diffusion-webui}} widely employ methods such as CLIPSeg \cite{luddecke2022image} for performing segmentation in AI-generated images. So we study the generalization ability of our proposed segmentation approach on AI-generated images. To this extent, we first prepare a dataset of 100 AI-generated images from lexica.art and manually annotate them using 214 text-prompts. We name this dataset AIGI and release it \camready{on our project website \footnote{\href{https://koutilya-pnvr.github.io/LD-ZNet/}{https://koutilya-pnvr.github.io/LD-ZNet/}}} for future research. Next, we evaluate our approaches ZNet and LD-ZNet along with our RGBNet baseline and other text-based segmentation methods - CLIPSeg (PC+) \cite{luddecke2022image}, MDETR \cite{kamath2021mdetr} \camready{and SEEM \cite{zou2023segment}}. Glipv2 \camready{and the SAM model \cite{kirillov2023segment} with textual input were} not publicly available for us to evaluate at the time of this submission. All these methods are trained on the Phrasecut dataset \camready{except for SEEM} and we measure the IoU metric as shown in Table \ref{tab:ai_generated_chart}. It can be seen that RGBNet outperforms CLIPSeg, MDETR and SEEM because its built on the UNet architecture initialized from the LDM weights that contains semantic information for good generalization. Our methods ZNet and LD-ZNet further improve the generalization to these AI-generated images by more than 20\% compared to MDETR. This is largely due to the robust $z$-space of the LDM that resulted from a VQGAN pre-training on a variety of domains like art, cartoons, illustrations \etc. Furthermore, the latent diffusion features that contain useful semantic information for the synthesis task, also help in segmenting the AI-generated images. We show the qualitative comparison of these methods in Figure \ref{fig:ai_generated_qualitative} for four AI-generated images from our dataset. While CLIPSeg can estimate most distinctive regions such as face of the \emph{Mickey mouse} or rough locations of \emph{Goblin}, \emph{Ramen} and \emph{animals}, MDETR \camready{and SEEM} incorrectly segment them because these concepts are unknown to them and because of the domain gap between \camready{their training data} and AIGI images respectively. In both such cases, our proposed LD-ZNet estimates accurate segmentation. More qualitative results for LD-ZNet on images from the AIGI dataset are shown in Figure \ref{fig:ldznet_ai_generated}.

\begin{figure*}[!t]
    \captionsetup[subfigure]{labelformat=empty,font=small,labelfont={bf,sf}}
    \centering
    \begin{subfigure}{\columnwidth}
        \centering
        \begin{subfigure}{0.32\textwidth}
            \centering
            \begin{subfigure}{\columnwidth}
                \centering
                \caption{}
                \includegraphics[width=\columnwidth]{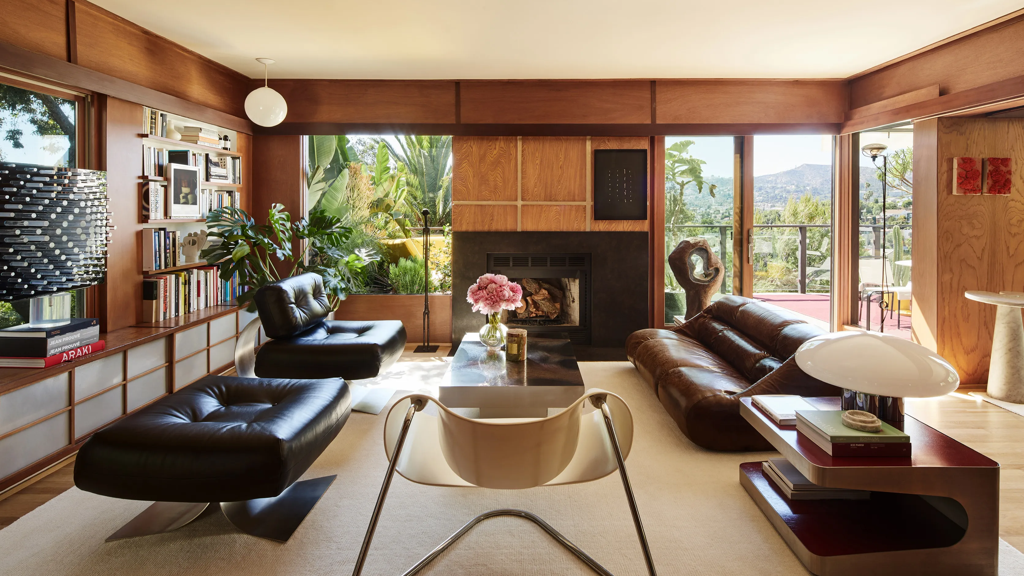}
            \end{subfigure}
            \begin{subfigure}{\columnwidth}
                \centering
                \caption{``Books"}
                \includegraphics[width=\columnwidth]{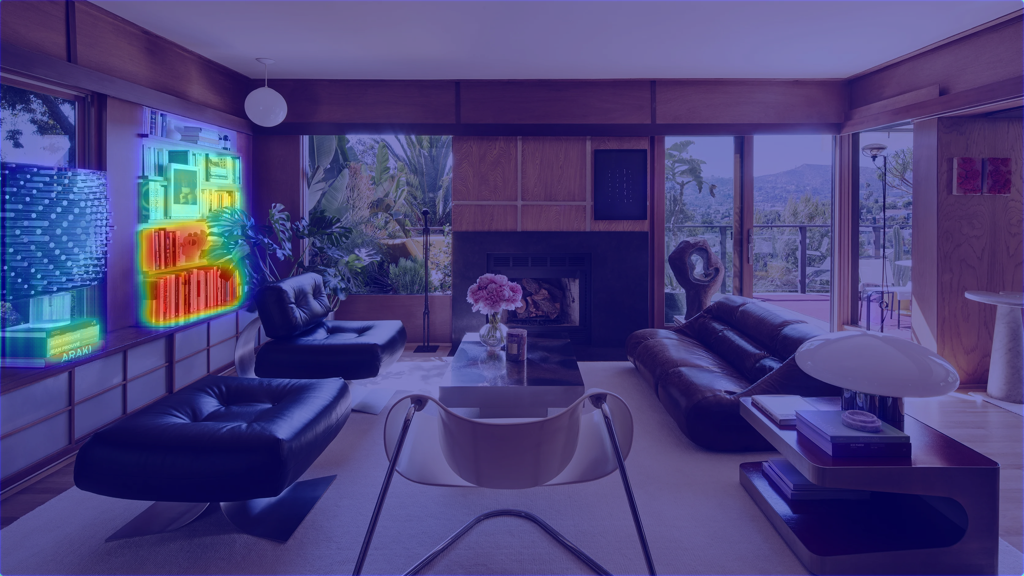}
            \end{subfigure}
        \end{subfigure}%
        \hspace{0.05cm}%
        \begin{subfigure}{0.32\textwidth}
            \centering
            \begin{subfigure}{\columnwidth}
                \centering
                \caption{``Flowers"}
                \includegraphics[width=\columnwidth]{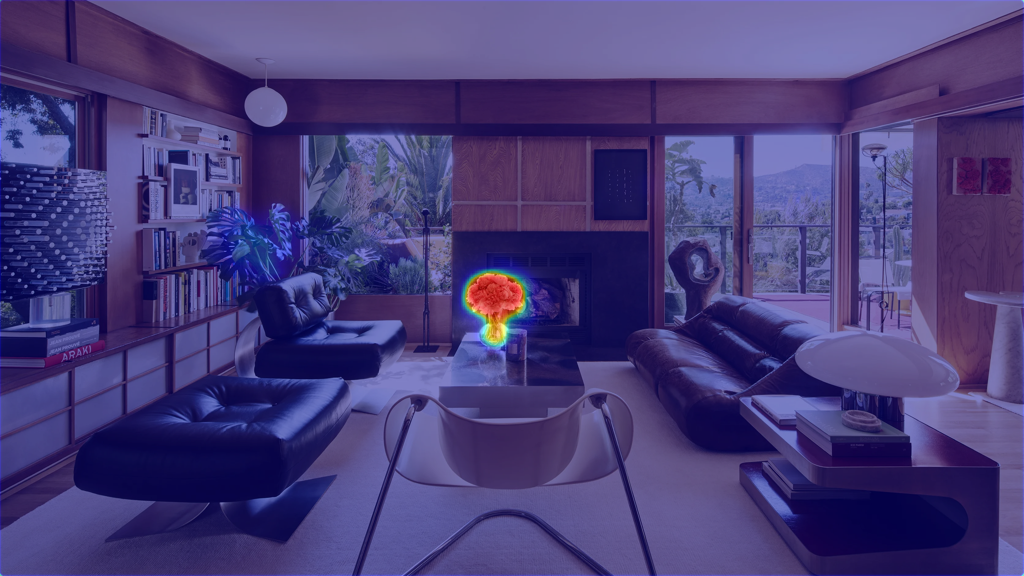}
            \end{subfigure}
            \begin{subfigure}{\columnwidth}
                \centering
                \caption{``Sofa"}
                \includegraphics[width=\columnwidth]{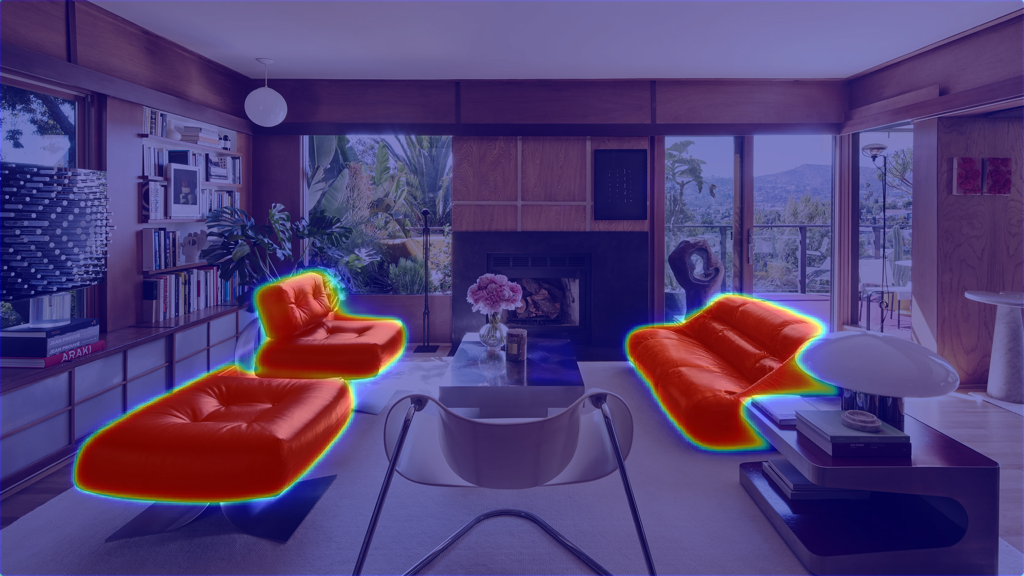}
            \end{subfigure}
        \end{subfigure}%
        \hspace{0.05cm}%
        \begin{subfigure}{0.32\textwidth}
            \centering
            \begin{subfigure}{\columnwidth}
                \centering
                \caption{``Table"}
                \includegraphics[width=\columnwidth]{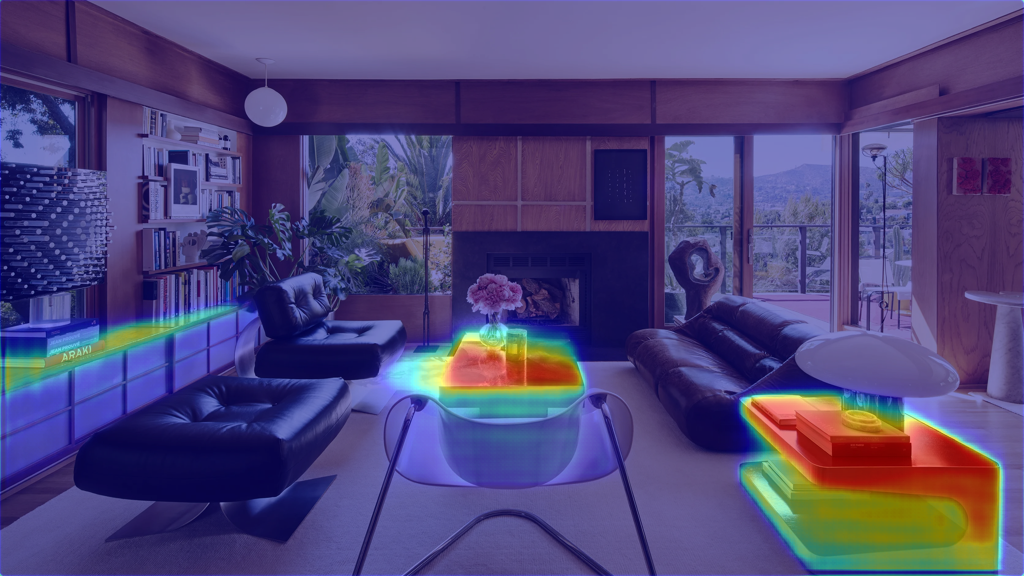}
            \end{subfigure}
            \begin{subfigure}{\columnwidth}
                \centering
                \caption{``Trees"}
                \includegraphics[width=\columnwidth]{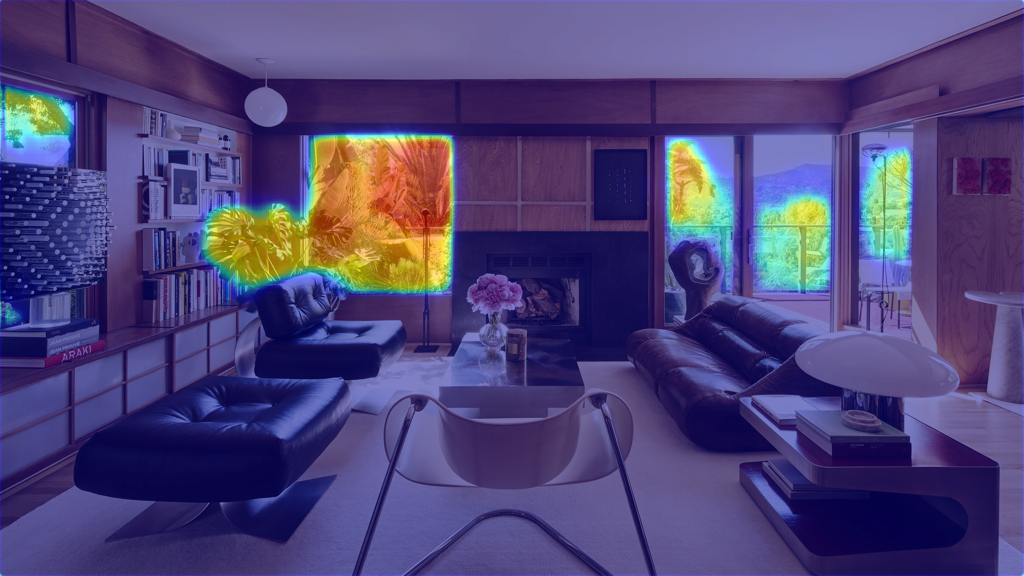}
            \end{subfigure}
        \end{subfigure}
    \end{subfigure}%
    \rulesep
    % \noindent\rule{\textwidth}{0.4pt}
    \begin{subfigure}{\columnwidth}
        \centering
        \begin{subfigure}{0.32\textwidth}
            \centering
            \begin{subfigure}{\columnwidth}
                \centering
                \caption{}
                \includegraphics[width=\columnwidth]{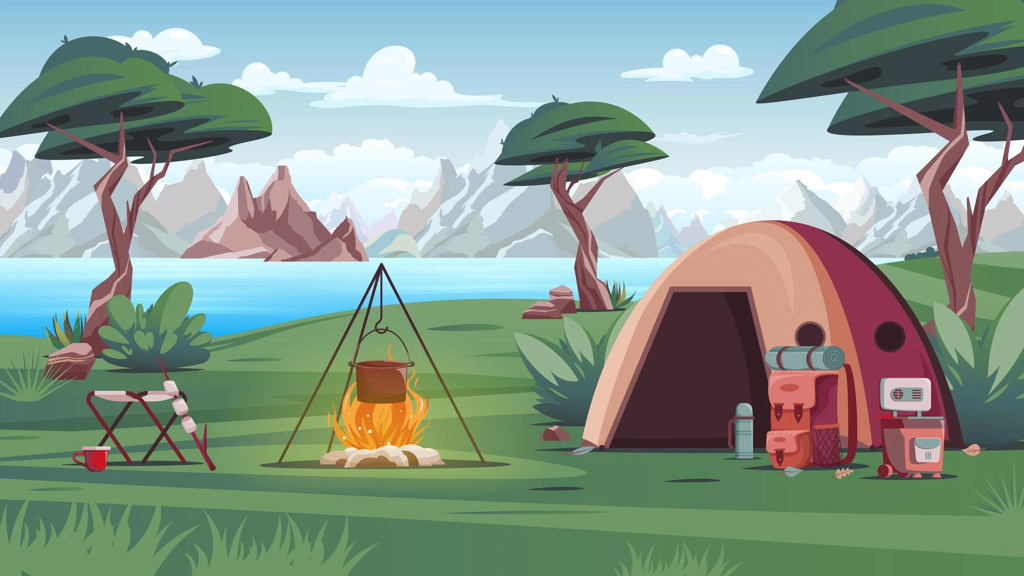}
                % Supp_Images/visual/zseg_SD_features/More_analysis/camping_an_image_of_a_Chair._image.png}
            \end{subfigure}
            \begin{subfigure}{\columnwidth}
                \centering
                \caption{``Chair"}
                \includegraphics[width=\columnwidth]{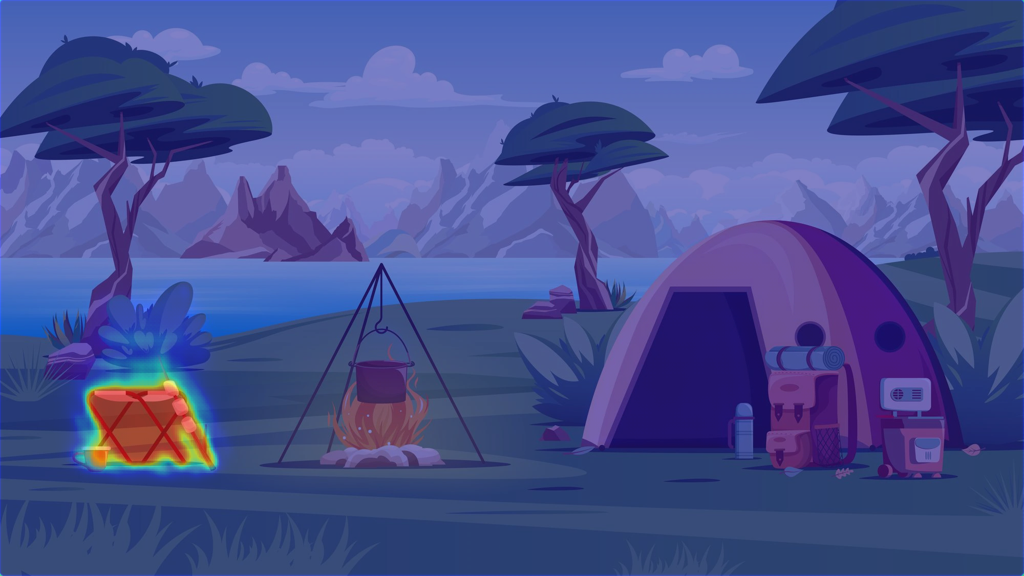}
            \end{subfigure}
        \end{subfigure}%
        \hspace{0.05cm}%
        \begin{subfigure}{0.32\textwidth}
            \centering
            \begin{subfigure}{\columnwidth}
                \centering
                \caption{``Clouds"}
                \includegraphics[width=\columnwidth]{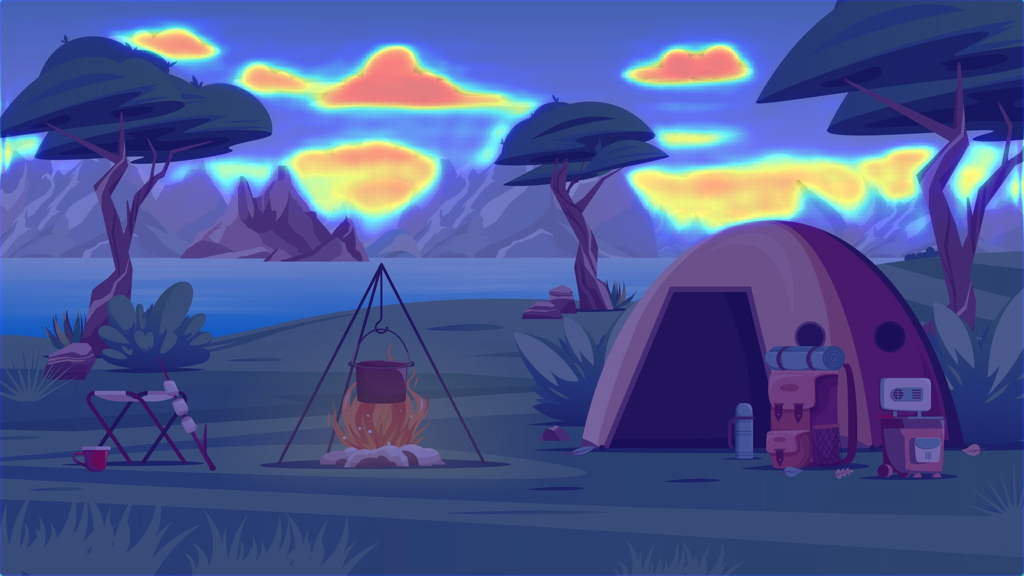}
            \end{subfigure}
            \begin{subfigure}{\columnwidth}
                \centering
                \caption{``Grass"}
                \includegraphics[width=\columnwidth]{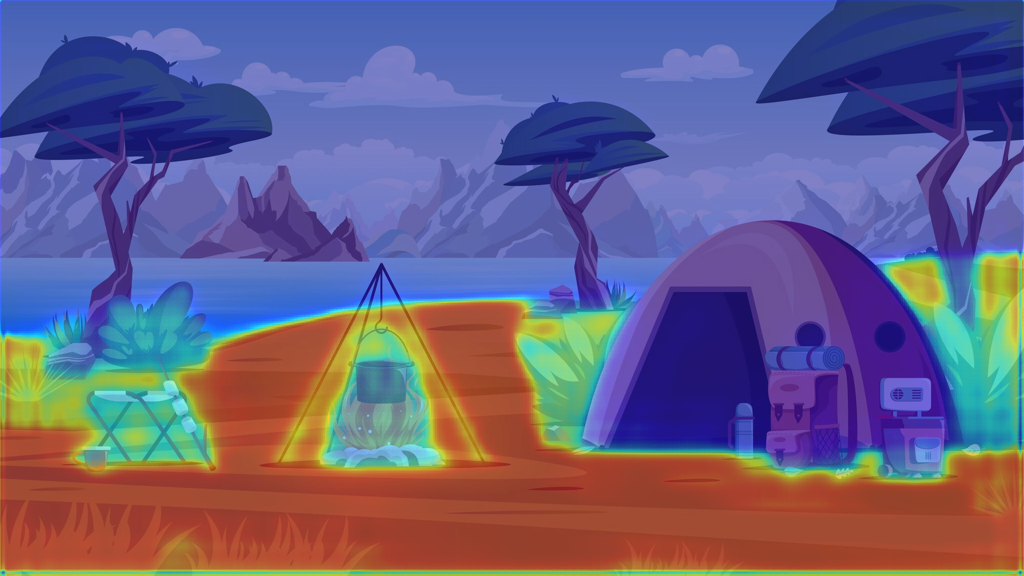}
            \end{subfigure}
        \end{subfigure}%
        \hspace{0.05cm}%
        \begin{subfigure}{0.32\textwidth}
            \centering
            \begin{subfigure}{\columnwidth}
                \centering
                \caption{``Mountains"}
                \includegraphics[width=\columnwidth]{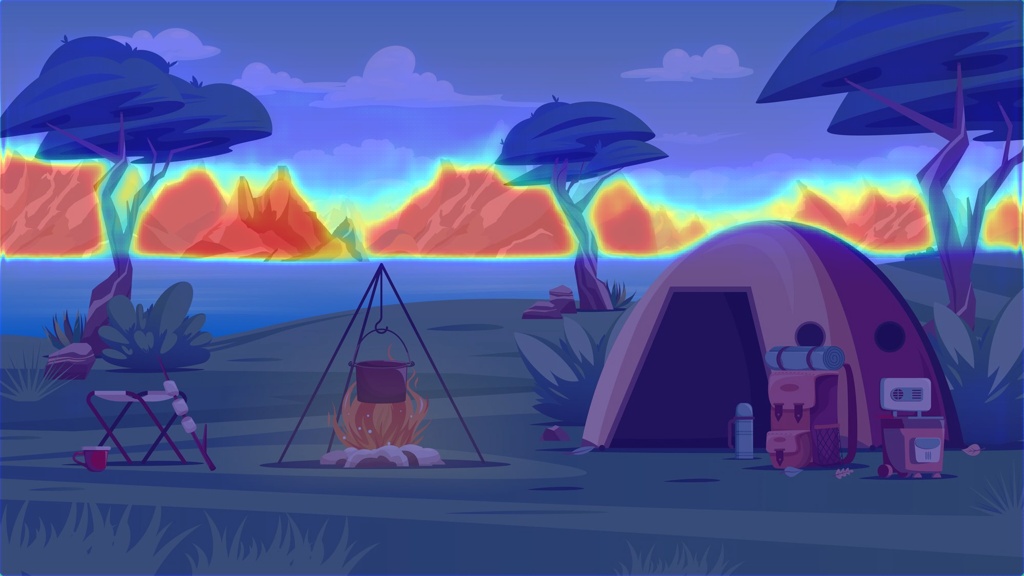}
            \end{subfigure}
            \begin{subfigure}{\columnwidth}
                \centering
                \caption{``River"}
                \includegraphics[width=\columnwidth]{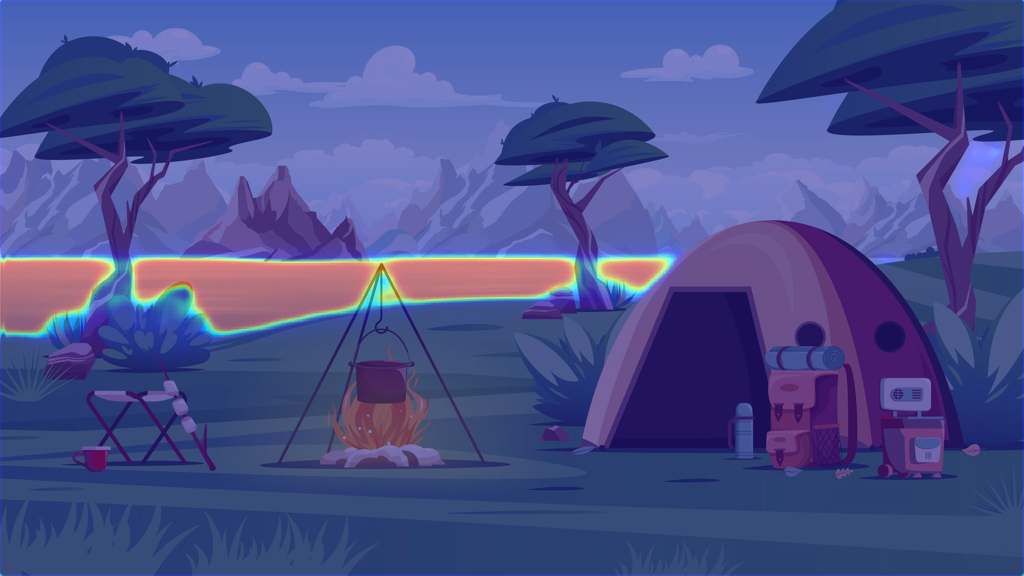}
            \end{subfigure}
        \end{subfigure}
    \end{subfigure}
    \caption{LD-ZNet text-based image segmentation results for a real image and a cartoon on diverse set of things and stuff classes. High quality segmentation across multiple classes suggests that LD-ZNet has a good understanding of the overall scene.}
    % Images used from google and \href{https://www.freepik.com}{freepik}.}
    \label{fig:scene_understanding}
\end{figure*}

\begin{table}[t]
    \centering
    \begin{adjustbox}{width=\columnwidth}
    \begin{tabular}{|c||c|c|c|c|c|c|}
    \hline
    \multirow{2}{*}{Method} & \multicolumn{2}{|c|}{RefCOCO} & \multicolumn{2}{|c|}{RefCOCO+} & \multicolumn{2}{|c|}{G-Ref}\\
    \cline{2-7}
    & IoU & AP & IoU & AP & IoU & AP\\
    \hline
            CLIPSeg (PC+) \cite{luddecke2022image}& 30.1 & 14.1 & 30.3 & 15.5 & 33.8 & 23.7 \\
            \hline\hline
            RGBNet & 36.3 & 15.7 & 37.1 & 16.7 & 41.9 & 27.8 \\
            \rowcolor{lightgray}ZNet (Ours) & 40.1 & 16.8 & 40.9 & 17.8 & 47.1 & 29.2 \\
            \rowcolor{lightgray}LD-ZNet (Ours) & \textbf{41.0} & \textbf{17.2} & \textbf{42.5} & \textbf{18.6} & \textbf{47.8} & \textbf{30.8} \\
    \hline
    \end{tabular}
    \end{adjustbox}
    \caption{Generalization of our proposed approaches to different types of expressions from other datasets. Z-Net and LD-ZNet outperform both the RGBNet baseline and CLIPSeg on the generalization across all datasets.}
    \vspace{-0.5em}
    \label{tab:ris_generalization}
\end{table}

\begin{table}
    \centering
    \begin{adjustbox}{width=0.95\columnwidth}
    \begin{tabular}{|c||c|c|c|c|}
    \hline
    Diffusion features via & mIoU & $IoU_{FG}$ & AP \\
    \hline
        LD-ZNet with concatenation & 50.2 & 59.0 & 78.1 \\
        LD-ZNet with cross-attention & \textbf{52.7} & \textbf{60.0} & \textbf{78.9} \\
    \hline
    \end{tabular}
    \end{adjustbox}
    \caption{Incorporating LDM features into ZNet via cross-attention (LD-ZNet) leverages the visual-linguistic information present in them, compared to concatenation, leading to better performance on the text-based image segmentation task.}
    \label{tab:ris_concat_vs_crossattn}
\end{table}

\subsection{Generalization to Referring Expressions}
Reference expression segmentation task is aimed for robot-localization kind of applications, where segmenting at instance-level is performed through distinctive referring expression. Many works such as \cite{yang2022lavt, wang2022cris} also train the text encoder to learn the complex positional references in the text. However, we are focused on generic text-based segmentation that has support for stuff categories as well as for multiple instances. We study the generalization ability of the proposed approach - using LDM features, to this complex task. Specifically, we use the models trained on the PhraseCut dataset and evaluate them on the RefCOCO \cite{kazemzadeh2014referitgame}, RefCOCO+ \cite{kazemzadeh2014referitgame} and G-Ref \cite{nagaraja16refexp} datasets whose complex referring expressions are for single-instance localization and segmentation. We also evaluated the generalization of CLIPSeg (PC+) \cite{luddecke2022image} model that was trained on an extended version of the PhraseCut dataset (PC+), to further demonstrate the generalization capability of our methods. Table \ref{tab:ris_generalization} summarizes the performance for our models along with the RGBNet baseline. We observe a similar trend in performance improvements across RGBNet $< Z$Net $<$ LD-ZNet. These experiments demonstrate that the LDM features enhance the generalization power of the LD-ZNet model even on complex referring expressions.

\camready{
\subsection{Inference Time} 
During  inference, our proposed LD-ZNet relies on the LDM to extract the internal features for just a single time step (as opposed to around 50 reverse diffusion time steps for the text-to-image synthesis task). We then use these LDM features for further cross-attention into LD-ZNet via the attention pool layer to extract the final mask. Therefore, using the diffusion model increases the overall run time by only a small amount. For the stable-diffusion model, inference takes 2.57s for 50 timesteps to synthesize an image (roughly 51ms per timestep), whereas the average inference times for RGBNet, ZNet and LD-ZNet are only 62ms, 55ms and 101ms, respectively, per image on the AIGI dataset with an RTX A6000 gpu. SEEM \cite{zou2023segment} takes 293ms for the same task. Since we use an architecture similar to UNet (from the second stage of the LDM), as our segmentation network, the proposed LD-ZNet has 925M trainable parameters. 
}

\begin{figure}[!ht]
    \captionsetup[subfigure]{labelformat=empty}
    \centering
    \begin{subfigure}{\linewidth}
        \centering
        \begin{subfigure}{0.32\columnwidth}
            \centering
            \caption{}
            \includegraphics[width=\columnwidth]{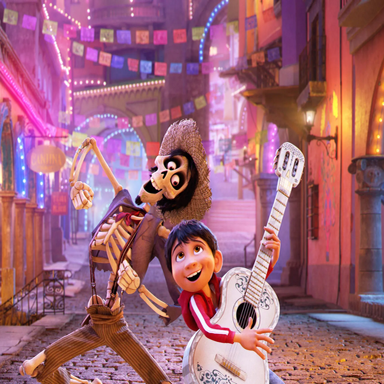}
            \includegraphics[width=\columnwidth]{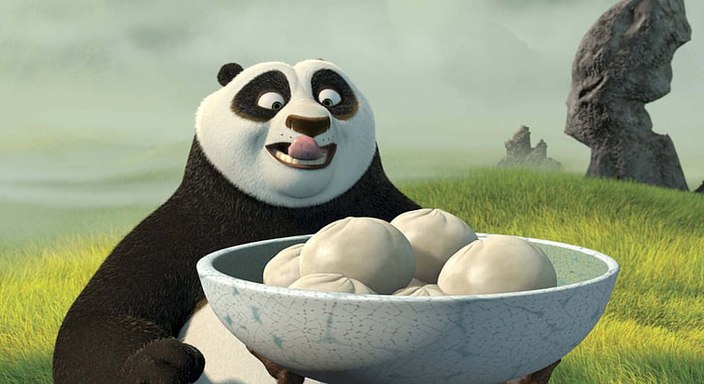}
        \end{subfigure}%
        \hspace{0.05cm}%
        \begin{subfigure}{0.32\columnwidth}
            \centering
            \caption{RGBNet}
            \includegraphics[width=\columnwidth]{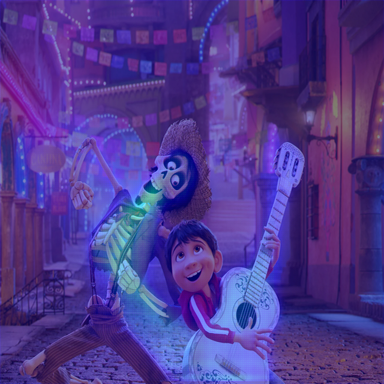}
            \includegraphics[width=\columnwidth]{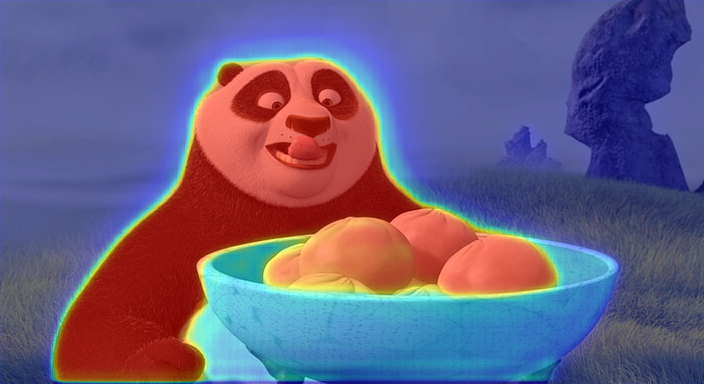}
        \end{subfigure}%
        \hspace{0.05cm}%
        \begin{subfigure}{0.32\columnwidth}
            \centering
            \caption{LD-ZNet}
            \includegraphics[width=\columnwidth]{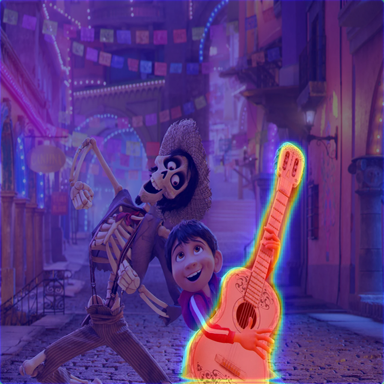}
            \includegraphics[width=\columnwidth]{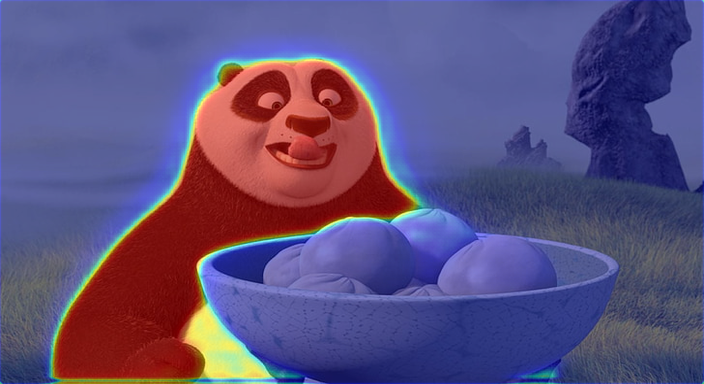}
        \end{subfigure}
        \vspace{-0.5em}
    \end{subfigure}
    
    \noindent\rule{\linewidth}{0.4pt}

    \begin{subfigure}{\linewidth}
        \vspace{-0.75em}
        \centering
        \begin{subfigure}{0.32\columnwidth}
            \centering
            \includegraphics[width=\columnwidth]{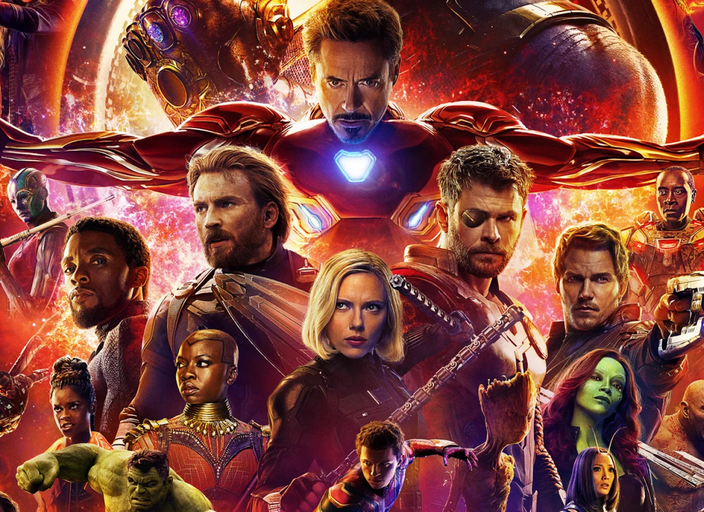}
            \includegraphics[width=\columnwidth]{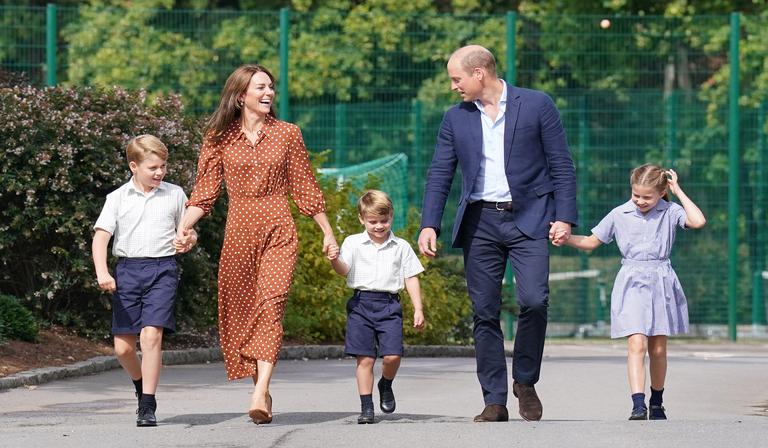}
        \end{subfigure}%
        \hspace{0.05cm}%
        \begin{subfigure}{0.32\columnwidth}
            \centering
            \caption{}
            \includegraphics[width=\columnwidth]{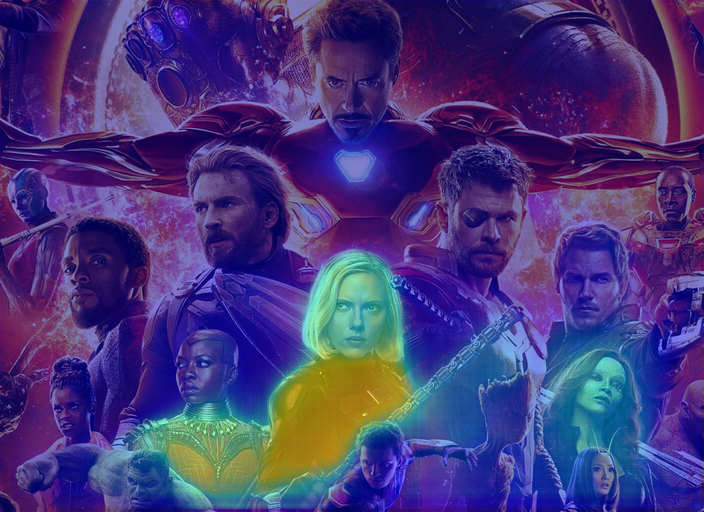}
            \includegraphics[width=\columnwidth]{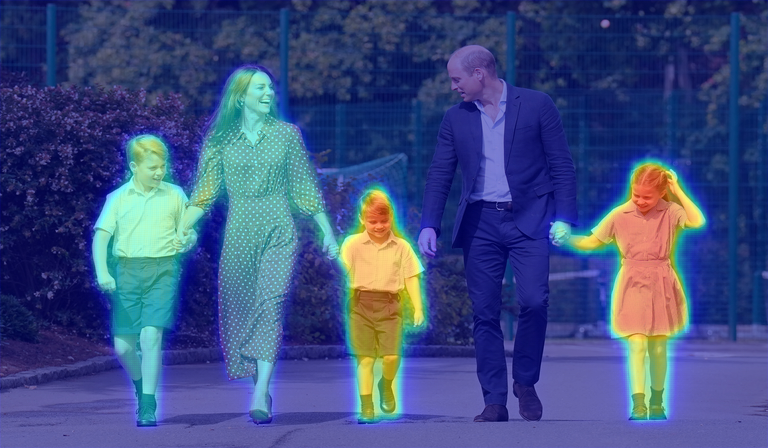}
        \end{subfigure}%
        \hspace{0.05cm}%
        \begin{subfigure}{0.32\columnwidth}
            \centering
            \caption{}
            \includegraphics[width=\columnwidth]{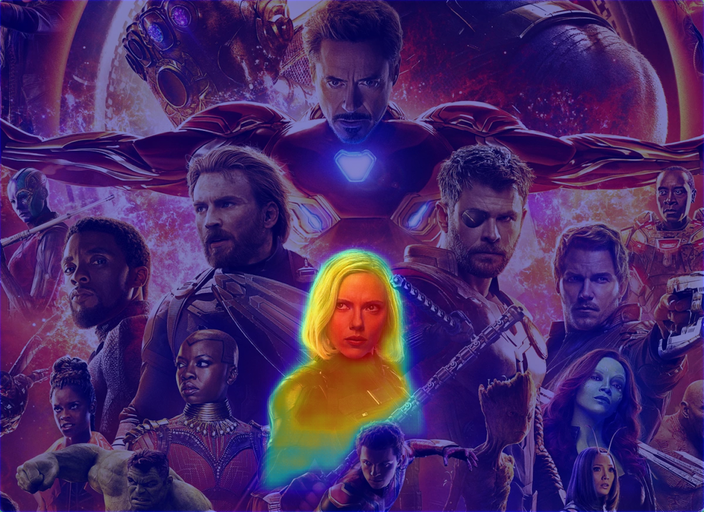}
            \includegraphics[width=\columnwidth]{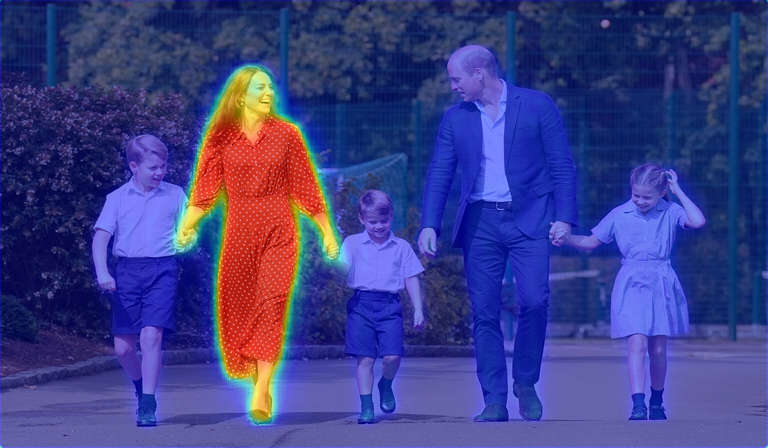}
        \end{subfigure}
        \vspace{-0.5em}
    \end{subfigure}
    
    \noindent\rule{\linewidth}{0.4pt}
    
    \begin{subfigure}{\linewidth}
        \vspace{-0.75em}
        \centering
        \begin{subfigure}{0.32\columnwidth}
            \centering
            \includegraphics[width=\columnwidth]{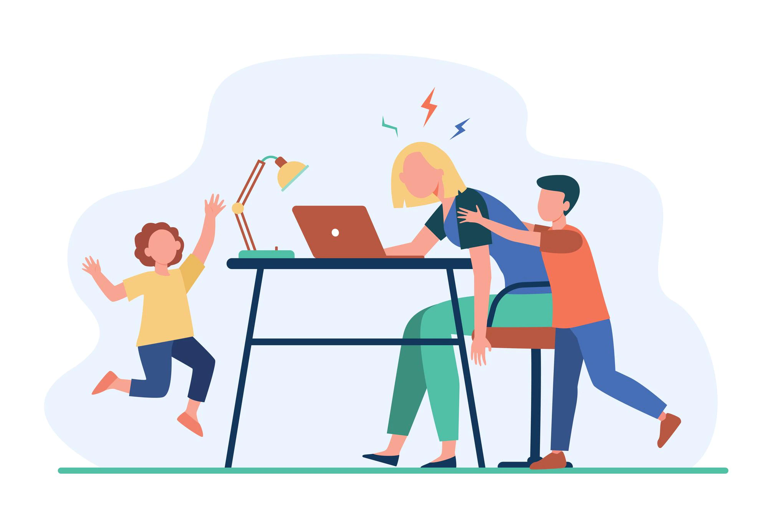}
            \includegraphics[width=\columnwidth]{Supp_Images/visual/RGBNet/camping_a_photograph_of_a_Trees._image.png}
        \end{subfigure}%
        \hspace{0.05cm}%
        \begin{subfigure}{0.32\columnwidth}
            \centering
            \caption{}
            \includegraphics[width=\columnwidth]{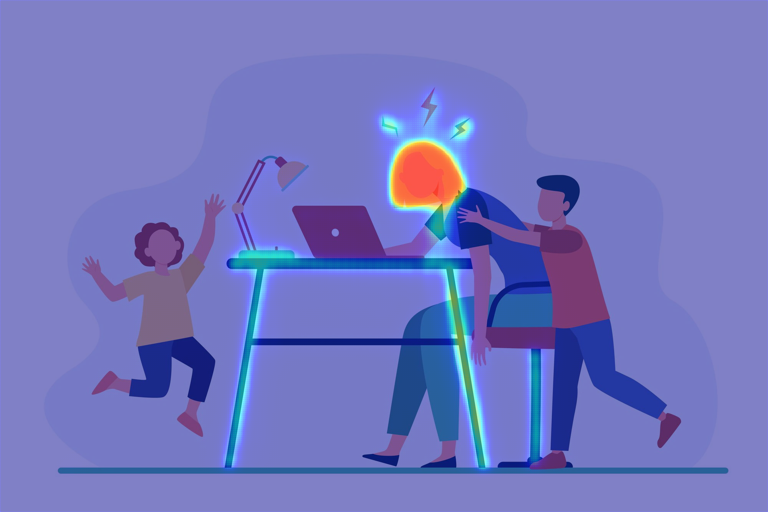}
            \includegraphics[width=\columnwidth]{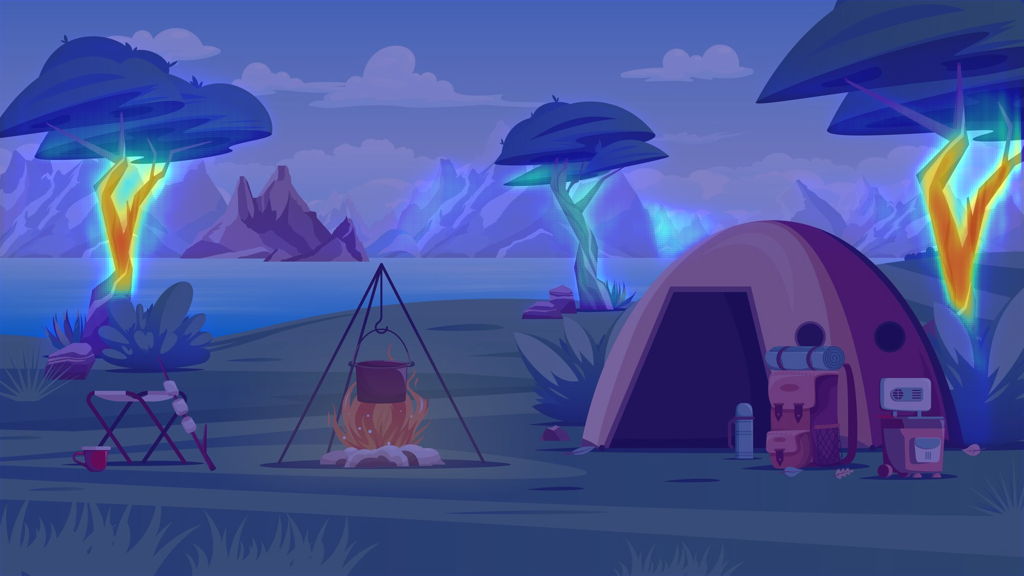}
        \end{subfigure}%
        \hspace{0.05cm}%
        \begin{subfigure}{0.32\columnwidth}
            \centering
            \caption{}
            \includegraphics[width=\columnwidth]{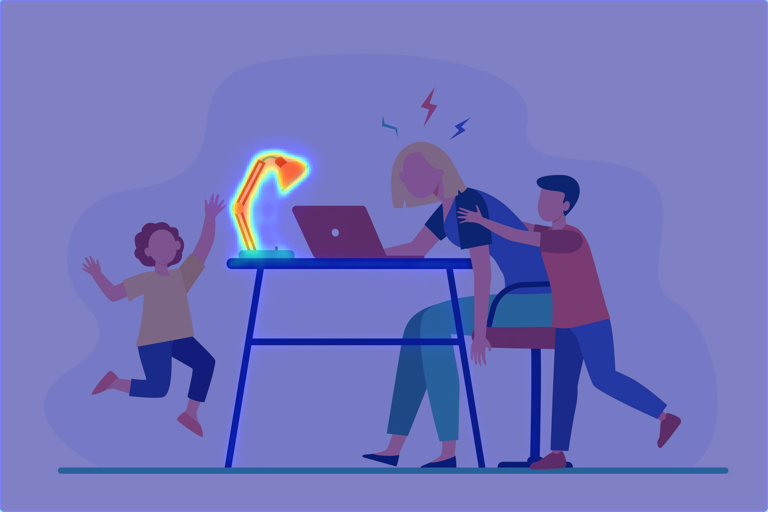}
            \includegraphics[width=\columnwidth]{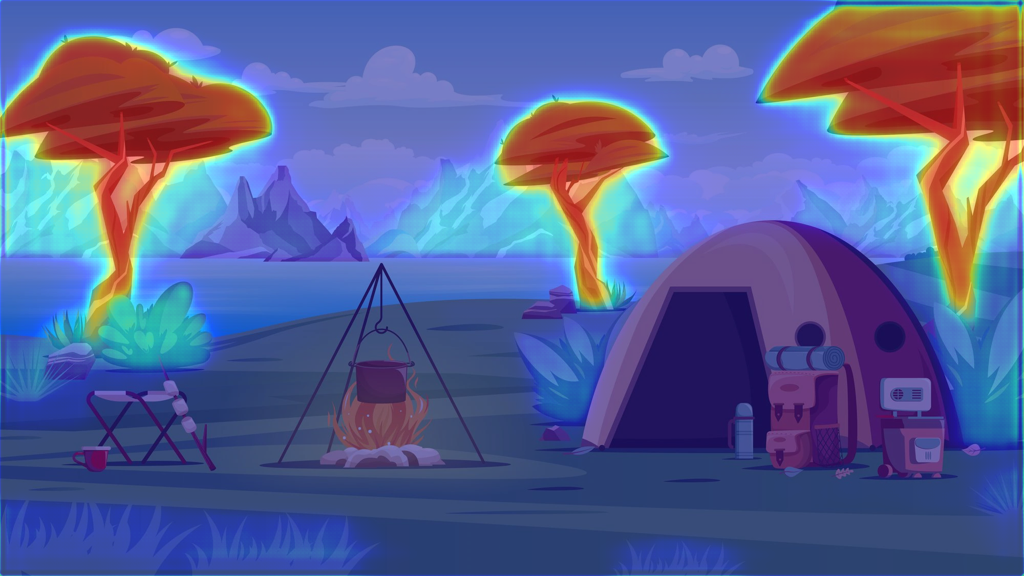}
        \end{subfigure}
    \end{subfigure}
    
    \caption{More qualitative examples where RGBNet fails to localize \emph{``Guitar", ``Panda"} from animation images (top row), famous celebrities \emph{``Scarlett Johansson", ``Kate Middleton"} (second row) and objects such as \emph{``Lamp", ``Trees"} from illustrations (bottom row). LD-ZNet benefits from using $z$ combined with the internal LDM features to correctly segment these text prompts.}
    \vspace{-1em}
    \label{fig:visual_results3}
\end{figure}

%------------------------------------------------------------------------
% Ablation studies
\subsection{Cross-attention vs Concat for LDM features}
\label{ablations}

In LD-ZNet, we inject LDM features into the ZNet model using cross-attention (Figure \ref{fig:spatial-attention}). In order to understand the importance of the cross-attention layer, we also train and evaluate another model where the LDM features are concatenated with the features of the ZNet right before the spatial-attention layer. The results are summarized in Table \ref{tab:ris_concat_vs_crossattn} and it shows that concatenating the LDM features yields inferior results compared to the proposed method. This is because of the \emph{attention pool} layer which serves as a learnable layer and also encodes positional information into the LDM features for setting up the cross-attention. Moreover, the cross-attention layer learns how feature pixels from the ZNet attend to feature pixels from the LDM, thereby leveraging context and correlations from the entire image. With concatenation however, we only fuse the corresponding features of LDM and ZNet which is sub-optimal. 

% \begin{table}
%     \centering
%     \begin{adjustbox}{width=\columnwidth}
%     \begin{tabular}{|c||c|c|c|c|}
%     \hline
%     Diffusion features via & mIoU & $IoU_{FG}$ & AP \\
%     \hline
%         Concatenation & 50.22 & 58.04 & 78.19 \\
%         Cross-attention & \textbf{52.7} & \textbf{59.12} & \textbf{78.93} \\
%     \hline
%     \end{tabular}
%     \end{adjustbox}
%     \caption{Cross-attention of the diffusion features into the ZNet results in better usage of visual-linguistic information compared to concatenation}
%     \label{tab:ris_concat_vs_crossattn}
% \end{table}

%------------------------------------------------------------------------

% Discussion
\section{Discussion}
\label{discussion}

In this section we present more qualitative results to demonstrate several interesting aspects of our proposed technique when applied towards downstream segmentation tasks. %Figure~\ref{fig:visual_results} shows the original image and the GT mask along with outputs from the RGBNet baseline followed by ZNet and LD-ZNet, where both ZNet and LD-ZNet help improve results consistently. For example in the top row, RGBNet detects light fixtures for the ``hanging clock" prompt, and although ZNet does not have as strong activations for these incorrect detections, it is LD-ZNet that correctly segments the ``clock". Similarly in the bottom row, while RGBNet completely got the ``castle" wrong, ZNet correctly has activations on the right buildings, but with lower confidence. However, LD-ZNet improves it further. %In Figure~\ref{fig:category} we compare the RGBNet baseline with LD-ZNet for different categories in the PhraseCut test dataset and we see a consistent improvement in almost all categories except ``sky" and ``street". This suggests that LDM contain more information for foreground ``objects" vs background ``stuff". 
In \cref{fig:ai_generated_qualitative,fig:ldznet_ai_generated,fig:scene_understanding,fig:visual_results3}, we visualize results of text-based image segmentation on a diverse set of images, which include AI generated images, illustrations and generic photographs.
%Specifically in Figure~\ref{fig:ldznet_ai_generated}, we show segmentation results on animated images and illustrations and that LD-ZNet, by virtue of using LDM features is able to perform better than RGBNet which fails due to the domain gap.  %\review{talk about figures 12 and 13}. 
%
 In Figure~\ref{fig:scene_understanding}, we show that when LD-ZNet is applied on the same image with various text prompts, it is able to correctly segment the object and stuff classes being referred to in both examples. This capability is crucial for open-world segmentation and overall understanding of the scene. The results also highlights that the algorithm works remarkably well on other domains like cartoons/illustrations. It is noteworthy that LD-ZNet  can perform accurate segmentation for text prompts which include cartoons (Pikachu, Godzilla), celebrities (Donald Trump, Spiderman), famous landmarks (Eiffel Tower), as seen in Figure~\ref{fig:ldznet_ai_generated}. \camready{Finally, Figure~\ref{fig:visual_results3} shows the advantages of leveraging semantic information present in the latent diffusion features. Compared to our baseline RGBNet, the proposed LD-ZNet generates better segmentation maps across animations, celebrity images and illustrations.}
%
% Finally Figure~\ref{fig:visual_results_supp} shows that the LDM features help LD-ZNet to correctly recognize color, relative height, action pose etc.

\section{Conclusion}

We presented a novel approach for text-based image segmentation using large scale latent diffusion models. By training the segmentation models on the latent z-space, we were able to improve the generalization of segmentation models to new domains, like AI generated images. We also showed that this z-space is a better representation for text-to-image tasks in natural images. By utilizing the internal features of the LDM at appropriate time-steps, we were able to tap into the semantic information hidden inside the image synthesis pipeline using a cross-attention mechanism, which further improved the segmentation performance both on natural and AI generated images. This was experimentally validated on several publicly available datasets and on a new dataset of AI generated images, which we will make publicly available. 

\label{conclusion}

\section{Acknowledgments}
Koutilya PNVR and David Jacobs were supported in part by the National Science Foundation under grant number IIS-1910132 and IIS-2213335.
% \section{Final copy}

% You must include your signed IEEE copyright release form when you submit
% your finished paper. We MUST have this form before your paper can be
% published in the proceedings.

{\small
\bibliographystyle{ieee_fullname}
\bibliography{ms}

\begin{thebibliography}{10}\itemsep=-1pt

\bibitem{baranchuk2021label}
Dmitry Baranchuk, Ivan Rubachev, Andrey Voynov, Valentin Khrulkov, and Artem
  Babenko.
\newblock Label-efficient semantic segmentation with diffusion models.
\newblock {\em arXiv preprint arXiv:2112.03126}, 2021.

\bibitem{brooks2022instructpix2pix}
Tim Brooks, Aleksander Holynski, and Alexei~A Efros.
\newblock Instructpix2pix: Learning to follow image editing instructions.
\newblock {\em arXiv preprint arXiv:2211.09800}, 2022.

\bibitem{chen2018language}
Jianbo Chen, Yelong Shen, Jianfeng Gao, Jingjing Liu, and Xiaodong Liu.
\newblock Language-based image editing with recurrent attentive models.
\newblock In {\em Proceedings of the IEEE Conference on Computer Vision and
  Pattern Recognition}, pages 8721--8729, 2018.

\bibitem{pmlr-v119-chen20j}
Ting Chen, Simon Kornblith, Mohammad Norouzi, and Geoffrey Hinton.
\newblock A simple framework for contrastive learning of visual
  representations.
\newblock In Hal~Daumé III and Aarti Singh, editors, {\em Proceedings of the
  37th International Conference on Machine Learning}, volume 119 of {\em
  Proceedings of Machine Learning Research}, pages 1597--1607. PMLR, 13--18 Jul
  2020.

\bibitem{couairon2022diffedit}
Guillaume Couairon, Jakob Verbeek, Holger Schwenk, and Matthieu Cord.
\newblock Diffedit: Diffusion-based semantic image editing with mask guidance.
\newblock {\em arXiv preprint arXiv:2210.11427}, 2022.

\bibitem{de2001robust}
Fernando De~la Torre and Michael~J Black.
\newblock Robust principal component analysis for computer vision.
\newblock In {\em Proceedings Eighth IEEE International Conference on Computer
  Vision. ICCV 2001}, volume~1, pages 362--369. IEEE, 2001.

\bibitem{deng2009imagenet}
Jia Deng, Wei Dong, Richard Socher, Li-Jia Li, Kai Li, and Li Fei-Fei.
\newblock Imagenet: A large-scale hierarchical image database.
\newblock In {\em 2009 IEEE conference on computer vision and pattern
  recognition}, pages 248--255. Ieee, 2009.

\bibitem{NEURIPS2021_49ad23d1}
Prafulla Dhariwal and Alexander Nichol.
\newblock Diffusion models beat gans on image synthesis.
\newblock In M. Ranzato, A. Beygelzimer, Y. Dauphin, P.S. Liang, and J.~Wortman
  Vaughan, editors, {\em Advances in Neural Information Processing Systems},
  volume~34, pages 8780--8794. Curran Associates, Inc., 2021.

\bibitem{NEURIPS2021_a4d92e2c}
Ming Ding, Zhuoyi Yang, Wenyi Hong, Wendi Zheng, Chang Zhou, Da Yin, Junyang
  Lin, Xu Zou, Zhou Shao, Hongxia Yang, and Jie Tang.
\newblock Cogview: Mastering text-to-image generation via transformers.
\newblock In M. Ranzato, A. Beygelzimer, Y. Dauphin, P.S. Liang, and J.~Wortman
  Vaughan, editors, {\em Advances in Neural Information Processing Systems},
  volume~34, pages 19822--19835. Curran Associates, Inc., 2021.

\bibitem{dosovitskiy2016generating}
Alexey Dosovitskiy and Thomas Brox.
\newblock Generating images with perceptual similarity metrics based on deep
  networks.
\newblock {\em Advances in neural information processing systems}, 29, 2016.

\bibitem{esser2021taming}
Patrick Esser, Robin Rombach, and Bjorn Ommer.
\newblock Taming transformers for high-resolution image synthesis.
\newblock In {\em Proceedings of the IEEE/CVF conference on computer vision and
  pattern recognition}, pages 12873--12883, 2021.

\bibitem{fei2006one}
Li Fei-Fei, Robert Fergus, and Pietro Perona.
\newblock One-shot learning of object categories.
\newblock {\em IEEE transactions on pattern analysis and machine intelligence},
  28(4):594--611, 2006.

\bibitem{gafni2022make}
Oran Gafni, Adam Polyak, Oron Ashual, Shelly Sheynin, Devi Parikh, and Yaniv
  Taigman.
\newblock Make-a-scene: Scene-based text-to-image generation with human priors.
\newblock {\em arXiv preprint arXiv:2203.13131}, 2022.

\bibitem{gu2022vector}
Shuyang Gu, Dong Chen, Jianmin Bao, Fang Wen, Bo Zhang, Dongdong Chen, Lu Yuan,
  and Baining Guo.
\newblock Vector quantized diffusion model for text-to-image synthesis.
\newblock In {\em Proceedings of the IEEE/CVF Conference on Computer Vision and
  Pattern Recognition}, pages 10696--10706, 2022.

\bibitem{hertz2022prompt}
Amir Hertz, Ron Mokady, Jay Tenenbaum, Kfir Aberman, Yael Pritch, and Daniel
  Cohen-Or.
\newblock Prompt-to-prompt image editing with cross attention control.
\newblock {\em arXiv preprint arXiv:2208.01626}, 2022.

\bibitem{hu2016segmentation}
Ronghang Hu, Marcus Rohrbach, and Trevor Darrell.
\newblock Segmentation from natural language expressions.
\newblock In {\em European Conference on Computer Vision}, pages 108--124.
  Springer, 2016.

\bibitem{isola2017image}
Phillip Isola, Jun-Yan Zhu, Tinghui Zhou, and Alexei~A Efros.
\newblock Image-to-image translation with conditional adversarial networks.
\newblock In {\em Proceedings of the IEEE conference on computer vision and
  pattern recognition}, pages 1125--1134, 2017.

\bibitem{kamath2021mdetr}
Aishwarya Kamath, Mannat Singh, Yann LeCun, Gabriel Synnaeve, Ishan Misra, and
  Nicolas Carion.
\newblock Mdetr-modulated detection for end-to-end multi-modal understanding.
\newblock In {\em Proceedings of the IEEE/CVF International Conference on
  Computer Vision}, pages 1780--1790, 2021.

\bibitem{karras2019style}
Tero Karras, Samuli Laine, and Timo Aila.
\newblock A style-based generator architecture for generative adversarial
  networks.
\newblock In {\em Proceedings of the IEEE/CVF conference on computer vision and
  pattern recognition}, pages 4401--4410, 2019.

\bibitem{kazemzadeh2014referitgame}
Sahar Kazemzadeh, Vicente Ordonez, Mark Matten, and Tamara Berg.
\newblock Referitgame: Referring to objects in photographs of natural scenes.
\newblock In {\em Proceedings of the 2014 conference on empirical methods in
  natural language processing (EMNLP)}, pages 787--798, 2014.

\bibitem{ke2004pca}
Yan Ke and Rahul Sukthankar.
\newblock Pca-sift: A more distinctive representation for local image
  descriptors.
\newblock In {\em Proceedings of the 2004 IEEE Computer Society Conference on
  Computer Vision and Pattern Recognition, 2004. CVPR 2004.}, volume~2, pages
  II--II. IEEE, 2004.

\bibitem{kirillov2023segment}
Alexander Kirillov, Eric Mintun, Nikhila Ravi, Hanzi Mao, Chloe Rolland, Laura
  Gustafson, Tete Xiao, Spencer Whitehead, Alexander~C. Berg, Wan-Yen Lo, Piotr
  Dollár, and Ross Girshick.
\newblock Segment anything, 2023.

\bibitem{semanticGAN}
Daiqing Li, Junlin Yang, Karsten Kreis, Antonio Torralba, and Sanja Fidler.
\newblock Semantic segmentation with generative models: Semi-supervised
  learning and strong out-of-domain generalization.
\newblock In {\em Conference on Computer Vision and Pattern Recognition
  (CVPR)}, 2021.

\bibitem{li2018referring}
Ruiyu Li, Kaican Li, Yi-Chun Kuo, Michelle Shu, Xiaojuan Qi, Xiaoyong Shen, and
  Jiaya Jia.
\newblock Referring image segmentation via recurrent refinement networks.
\newblock In {\em Proceedings of the IEEE Conference on Computer Vision and
  Pattern Recognition}, pages 5745--5753, 2018.

\bibitem{liu2017recurrent}
Chenxi Liu, Zhe Lin, Xiaohui Shen, Jimei Yang, Xin Lu, and Alan Yuille.
\newblock Recurrent multimodal interaction for referring image segmentation.
\newblock In {\em Proceedings of the IEEE International Conference on Computer
  Vision}, pages 1271--1280, 2017.

\bibitem{luddecke2022image}
Timo L{\"u}ddecke and Alexander Ecker.
\newblock Image segmentation using text and image prompts.
\newblock In {\em Proceedings of the IEEE/CVF Conference on Computer Vision and
  Pattern Recognition}, pages 7086--7096, 2022.

\bibitem{Lugmayr_2022_CVPR}
Andreas Lugmayr, Martin Danelljan, Andres Romero, Fisher Yu, Radu Timofte, and
  Luc Van~Gool.
\newblock Repaint: Inpainting using denoising diffusion probabilistic models.
\newblock In {\em Proceedings of the IEEE/CVF Conference on Computer Vision and
  Pattern Recognition (CVPR)}, pages 11461--11471, June 2022.

\bibitem{margffoy2018dynamic}
Edgar Margffoy-Tuay, Juan~C P{\'e}rez, Emilio Botero, and Pablo Arbel{\'a}ez.
\newblock Dynamic multimodal instance segmentation guided by natural language
  queries.
\newblock In {\em Proceedings of the European Conference on Computer Vision
  (ECCV)}, pages 630--645, 2018.

\bibitem{melas2021finding}
Luke Melas-Kyriazi, Christian Rupprecht, Iro Laina, and Andrea Vedaldi.
\newblock Finding an unsupervised image segmenter in each of your deep
  generative models.
\newblock {\em arXiv preprint arXiv:2105.08127}, 2021.

\bibitem{nagaraja16refexp}
Varun~K. Nagaraja, Vlad~I. Morariu, and Larry~S. Davis.
\newblock Modeling context between objects for referring expression
  understanding.
\newblock In {\em ECCV}, 2016.

\bibitem{nichol2021glide}
Alex Nichol, Prafulla Dhariwal, Aditya Ramesh, Pranav Shyam, Pamela Mishkin,
  Bob McGrew, Ilya Sutskever, and Mark Chen.
\newblock Glide: Towards photorealistic image generation and editing with
  text-guided diffusion models.
\newblock {\em arXiv preprint arXiv:2112.10741}, 2021.

\bibitem{nichol2021improved}
Alexander~Quinn Nichol and Prafulla Dhariwal.
\newblock Improved denoising diffusion probabilistic models.
\newblock In {\em International Conference on Machine Learning}, pages
  8162--8171. PMLR, 2021.

\bibitem{pakhomov2021segmentation}
Daniil Pakhomov, Sanchit Hira, Narayani Wagle, Kemar~E. Green, and Nassir
  Navab.
\newblock Segmentation in style: Unsupervised semantic image segmentation with
  stylegan and clip, 2021.

\bibitem{radford2021learning}
Alec Radford, Jong~Wook Kim, Chris Hallacy, Aditya Ramesh, Gabriel Goh,
  Sandhini Agarwal, Girish Sastry, Amanda Askell, Pamela Mishkin, Jack Clark,
  et~al.
\newblock Learning transferable visual models from natural language
  supervision.
\newblock In {\em International Conference on Machine Learning}, pages
  8748--8763. PMLR, 2021.

\bibitem{ramesh2022hierarchical}
Aditya Ramesh, Prafulla Dhariwal, Alex Nichol, Casey Chu, and Mark Chen.
\newblock Hierarchical text-conditional image generation with clip latents.
\newblock {\em arXiv preprint arXiv:2204.06125}, 2022.

\bibitem{ramesh2021zero}
Aditya Ramesh, Mikhail Pavlov, Gabriel Goh, Scott Gray, Chelsea Voss, Alec
  Radford, Mark Chen, and Ilya Sutskever.
\newblock In {\em International Conference on Machine Learning}, pages
  8821--8831. PMLR, 2021.

\bibitem{razavi2019generating}
Ali Razavi, Aaron Van~den Oord, and Oriol Vinyals.
\newblock Generating diverse high-fidelity images with vq-vae-2.
\newblock {\em Advances in neural information processing systems}, 32, 2019.

\bibitem{rombach2022high}
Robin Rombach, Andreas Blattmann, Dominik Lorenz, Patrick Esser, and Bj{\"o}rn
  Ommer.
\newblock High-resolution image synthesis with latent diffusion models.
\newblock In {\em Proceedings of the IEEE/CVF Conference on Computer Vision and
  Pattern Recognition}, pages 10684--10695, 2022.

\bibitem{UNet}
Olaf Ronneberger, Philipp Fischer, and Thomas Brox.
\newblock U-net: Convolutional networks for biomedical image segmentation,
  2015.

\bibitem{saharia2022photorealistic}
Chitwan Saharia, William Chan, Saurabh Saxena, Lala Li, Jay Whang, Emily
  Denton, Seyed Kamyar~Seyed Ghasemipour, Burcu~Karagol Ayan, S~Sara Mahdavi,
  Rapha~Gontijo Lopes, et~al.
\newblock Photorealistic text-to-image diffusion models with deep language
  understanding.
\newblock {\em arXiv preprint arXiv:2205.11487}, 2022.

\bibitem{schuhmann2022laionb}
Christoph Schuhmann, Romain Beaumont, Richard Vencu, Cade~W Gordon, Ross
  Wightman, Mehdi Cherti, Theo Coombes, Aarush Katta, Clayton Mullis, Mitchell
  Wortsman, Patrick Schramowski, Srivatsa~R Kundurthy, Katherine Crowson,
  Ludwig Schmidt, Robert Kaczmarczyk, and Jenia Jitsev.
\newblock {LAION}-5b: An open large-scale dataset for training next generation
  image-text models.
\newblock In {\em Thirty-sixth Conference on Neural Information Processing
  Systems Datasets and Benchmarks Track}, 2022.

\bibitem{selvaraju2017grad}
Ramprasaath~R Selvaraju, Michael Cogswell, Abhishek Das, Ramakrishna Vedantam,
  Devi Parikh, and Dhruv Batra.
\newblock Grad-cam: Visual explanations from deep networks via gradient-based
  localization.
\newblock In {\em Proceedings of the IEEE international conference on computer
  vision}, pages 618--626, 2017.

\bibitem{shi2018key}
Hengcan Shi, Hongliang Li, Fanman Meng, and Qingbo Wu.
\newblock Key-word-aware network for referring expression image segmentation.
\newblock In {\em Proceedings of the European Conference on Computer Vision
  (ECCV)}, pages 38--54, 2018.

\bibitem{tang2022improved}
Zhicong Tang, Shuyang Gu, Jianmin Bao, Dong Chen, and Fang Wen.
\newblock Improved vector quantized diffusion models.
\newblock {\em arXiv preprint arXiv:2205.16007}, 2022.

\bibitem{tao2022df}
Ming Tao, Hao Tang, Fei Wu, Xiao-Yuan Jing, Bing-Kun Bao, and Changsheng Xu.
\newblock Df-gan: A simple and effective baseline for text-to-image synthesis.
\newblock In {\em Proceedings of the IEEE/CVF Conference on Computer Vision and
  Pattern Recognition}, pages 16515--16525, 2022.

\bibitem{tritrong2021repurposing}
Nontawat Tritrong, Pitchaporn Rewatbowornwong, and Supasorn Suwajanakorn.
\newblock Repurposing gans for one-shot semantic part segmentation.
\newblock In {\em Proceedings of the IEEE/CVF conference on computer vision and
  pattern recognition}, pages 4475--4485, 2021.

\bibitem{turk1991eigenfaces}
Matthew Turk and Alex Pentland.
\newblock Eigenfaces for recognition.
\newblock {\em Journal of cognitive neuroscience}, 3(1):71--86, 1991.

\bibitem{van2017neural}
Aaron Van Den~Oord, Oriol Vinyals, et~al.
\newblock Neural discrete representation learning.
\newblock {\em Advances in neural information processing systems}, 30, 2017.

\bibitem{voynov2021object}
Andrey Voynov, Stanislav Morozov, and Artem Babenko.
\newblock Object segmentation without labels with large-scale generative
  models.
\newblock In {\em International Conference on Machine Learning}, pages
  10596--10606. PMLR, 2021.

\bibitem{wang2022cris}
Zhaoqing Wang, Yu Lu, Qiang Li, Xunqiang Tao, Yandong Guo, Mingming Gong, and
  Tongliang Liu.
\newblock Cris: Clip-driven referring image segmentation.
\newblock In {\em Proceedings of the IEEE/CVF Conference on Computer Vision and
  Pattern Recognition}, pages 11686--11695, 2022.

\bibitem{wu2020phrasecut}
Chenyun Wu, Zhe Lin, Scott Cohen, Trung Bui, and Subhransu Maji.
\newblock Phrasecut: Language-based image segmentation in the wild.
\newblock In {\em Proceedings of the IEEE/CVF Conference on Computer Vision and
  Pattern Recognition}, pages 10216--10225, 2020.

\bibitem{xie2022smartbrush}
Shaoan Xie, Zhifei Zhang, Zhe Lin, Tobias Hinz, and Kun Zhang.
\newblock Smartbrush: Text and shape guided object inpainting with diffusion
  model.
\newblock {\em arXiv preprint arXiv:2212.05034}, 2022.

\bibitem{Xu_2018_CVPR}
Tao Xu, Pengchuan Zhang, Qiuyuan Huang, Han Zhang, Zhe Gan, Xiaolei Huang, and
  Xiaodong He.
\newblock Attngan: Fine-grained text to image generation with attentional
  generative adversarial networks.
\newblock In {\em Proceedings of the IEEE Conference on Computer Vision and
  Pattern Recognition (CVPR)}, June 2018.

\bibitem{yang2022lavt}
Zhao Yang, Jiaqi Wang, Yansong Tang, Kai Chen, Hengshuang Zhao, and Philip~HS
  Torr.
\newblock Lavt: Language-aware vision transformer for referring image
  segmentation.
\newblock In {\em Proceedings of the IEEE/CVF Conference on Computer Vision and
  Pattern Recognition}, pages 18155--18165, 2022.

\bibitem{ye2021improving}
Hui Ye, Xiulong Yang, Martin Takac, Rajshekhar Sunderraman, and Shihao Ji.
\newblock Improving text-to-image synthesis using contrastive learning.
\newblock {\em The 32nd British Machine Vision Conference (BMVC)}, 2021.

\bibitem{ye2019cross}
Linwei Ye, Mrigank Rochan, Zhi Liu, and Yang Wang.
\newblock Cross-modal self-attention network for referring image segmentation.
\newblock In {\em Proceedings of the IEEE/CVF conference on computer vision and
  pattern recognition}, pages 10502--10511, 2019.

\bibitem{yu15lsun}
Fisher Yu, Yinda Zhang, Shuran Song, Ari Seff, and Jianxiong Xiao.
\newblock Lsun: Construction of a large-scale image dataset using deep learning
  with humans in the loop.
\newblock {\em arXiv preprint arXiv:1506.03365}, 2015.

\bibitem{yu2021vector}
Jiahui Yu, Xin Li, Jing~Yu Koh, Han Zhang, Ruoming Pang, James Qin, Alexander
  Ku, Yuanzhong Xu, Jason Baldridge, and Yonghui Wu.
\newblock Vector-quantized image modeling with improved vqgan.
\newblock {\em arXiv preprint arXiv:2110.04627}, 2021.

\bibitem{yu2018mattnet}
Licheng Yu, Zhe Lin, Xiaohui Shen, Jimei Yang, Xin Lu, Mohit Bansal, and
  Tamara~L Berg.
\newblock Mattnet: Modular attention network for referring expression
  comprehension.
\newblock In {\em Proceedings of the IEEE Conference on Computer Vision and
  Pattern Recognition}, pages 1307--1315, 2018.

\bibitem{zeiler2014visualizing}
Matthew~D Zeiler and Rob Fergus.
\newblock Visualizing and understanding convolutional networks.
\newblock In {\em Computer Vision--ECCV 2014: 13th European Conference, Zurich,
  Switzerland, September 6-12, 2014, Proceedings, Part I 13}, pages 818--833.
  Springer, 2014.

\bibitem{zhang2021cross}
Han Zhang, Jing~Yu Koh, Jason Baldridge, Honglak Lee, and Yinfei Yang.
\newblock Cross-modal contrastive learning for text-to-image generation.
\newblock In {\em Proceedings of the IEEE/CVF conference on computer vision and
  pattern recognition}, pages 833--842, 2021.

\bibitem{zhang2022glipv2}
Haotian Zhang, Pengchuan Zhang, Xiaowei Hu, Yen-Chun Chen, Liunian~Harold Li,
  Xiyang Dai, Lijuan Wang, Lu Yuan, Jenq-Neng Hwang, and Jianfeng Gao.
\newblock Glipv2: Unifying localization and vision-language understanding.
\newblock In {\em Advances in Neural Information Processing Systems}, 2022.

\bibitem{zhang2018unreasonable}
Richard Zhang, Phillip Isola, Alexei~A Efros, Eli Shechtman, and Oliver Wang.
\newblock The unreasonable effectiveness of deep features as a perceptual
  metric.
\newblock In {\em Proceedings of the IEEE conference on computer vision and
  pattern recognition}, pages 586--595, 2018.

\bibitem{zhang2021datasetgan}
Yuxuan Zhang, Huan Ling, Jun Gao, Kangxue Yin, Jean-Francois Lafleche, Adela
  Barriuso, Antonio Torralba, and Sanja Fidler.
\newblock Datasetgan: Efficient labeled data factory with minimal human effort.
\newblock In {\em Proceedings of the IEEE/CVF Conference on Computer Vision and
  Pattern Recognition}, pages 10145--10155, 2021.

\bibitem{zhou2022towards}
Yufan Zhou, Ruiyi Zhang, Changyou Chen, Chunyuan Li, Chris Tensmeyer, Tong Yu,
  Jiuxiang Gu, Jinhui Xu, and Tong Sun.
\newblock Towards language-free training for text-to-image generation.
\newblock In {\em Proceedings of the IEEE/CVF Conference on Computer Vision and
  Pattern Recognition}, pages 17907--17917, 2022.

\bibitem{Zhu_2019_CVPR}
Minfeng Zhu, Pingbo Pan, Wei Chen, and Yi Yang.
\newblock Dm-gan: Dynamic memory generative adversarial networks for
  text-to-image synthesis.
\newblock In {\em Proceedings of the IEEE/CVF Conference on Computer Vision and
  Pattern Recognition (CVPR)}, June 2019.

\bibitem{zou2023segment}
Xueyan Zou, Jianwei Yang, Hao Zhang, Feng Li, Linjie Li, Jianfeng Gao, and
  Yong~Jae Lee.
\newblock Segment everything everywhere all at once, 2023.

\end{thebibliography}
}

% %%%%%%%%%% Merge with supplemental materials %%%%%%%%%%
% Supplementary
% \twocolumngrid
\clearpage
% \twocolumngrid
% \begin{center}
% \large\textbf{Supplementary Material}
% \end{center}
\twocolumn[{%
 \centering
 \LARGE \textbf{Supplementary Material}\\[1em]
 % \large Author: Anton van der Vegt\\[1em]
}]
%%%%%%%%%% Merge with supplemental materials %%%%%%%%%%
%%%%%%%%%% Prefix a "S" to all equations, figures, tables and reset the counter %%%%%%%%%%

\setcounter{equation}{0}
\setcounter{figure}{0}
\setcounter{table}{0}
\setcounter{page}{1}
\setcounter{section}{0}
\makeatletter
\renewcommand{\theequation}{S\arabic{equation}}
\renewcommand{\thefigure}{S\arabic{figure}}
\renewcommand{\thesection}{S\arabic{section}}

\section{Text-Based Image Segmentation}
In this supplementary work, we illustrate some more qualitative text-based image segmentation results using the proposed LD-ZNet model on a diverse set of images. Specifically, we focus on segmenting and localizing 1) objects described by their attributes 2) objects in AI-generated images from the AIGI dataset and 3) multiple different things and stuff in a scene. We also perform a visual comparison with the RGBNet baseline on a diverse set of images, including examples from the PhraseCut test dataset.

\subsection{Attributes}
\begin{figure}[!t]
    \centering
    \captionsetup[subfigure]{labelformat=empty,font=small,labelfont={bf,sf}}
    \begin{subfigure}{\columnwidth}
        \begin{subfigure}{0.49\columnwidth}
            \centering
            \caption{``Blue car"}
            \includegraphics[width=\columnwidth]{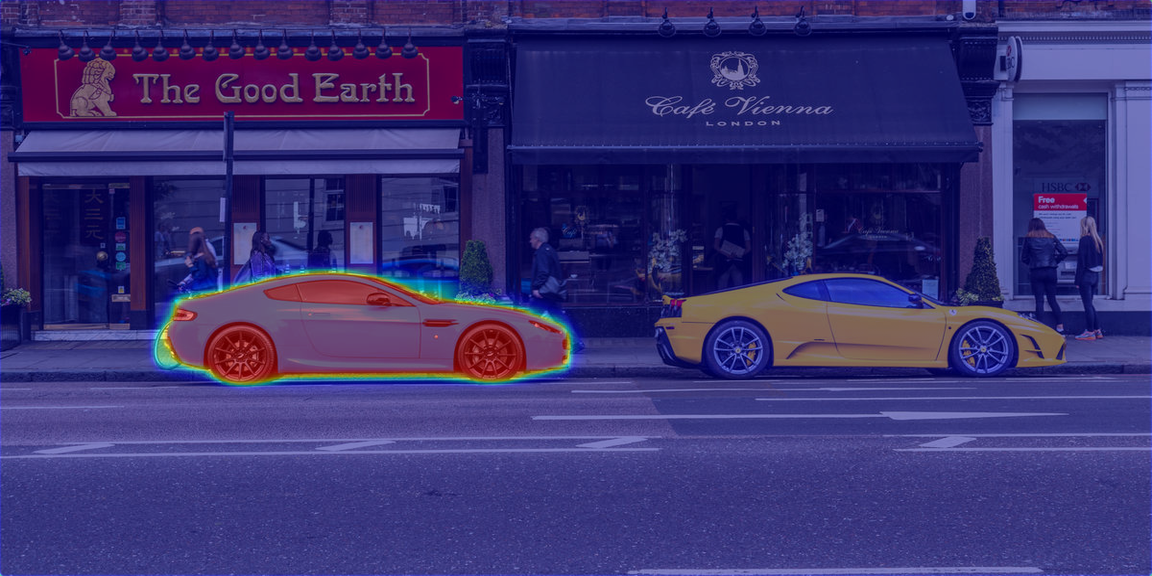}
        \end{subfigure}%
        \hspace{0.05cm}%
        \begin{subfigure}{0.49\columnwidth}
            \centering
            \caption{``Yellow car"}
            \includegraphics[width=\columnwidth]{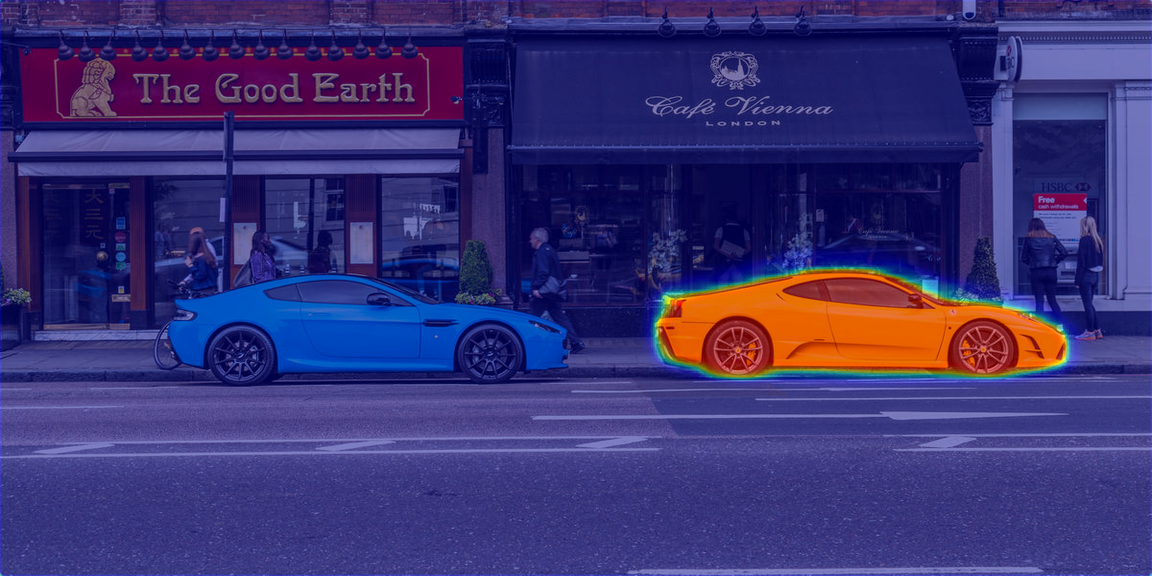}
        \end{subfigure}%
    \end{subfigure}
    % \noindent\rule{\columnwidth}{0.4pt}
    \begin{subfigure}{\columnwidth}
        \begin{subfigure}{0.49\columnwidth}
            \centering
            \caption{``Tallest person"}
            \includegraphics[width=\columnwidth]{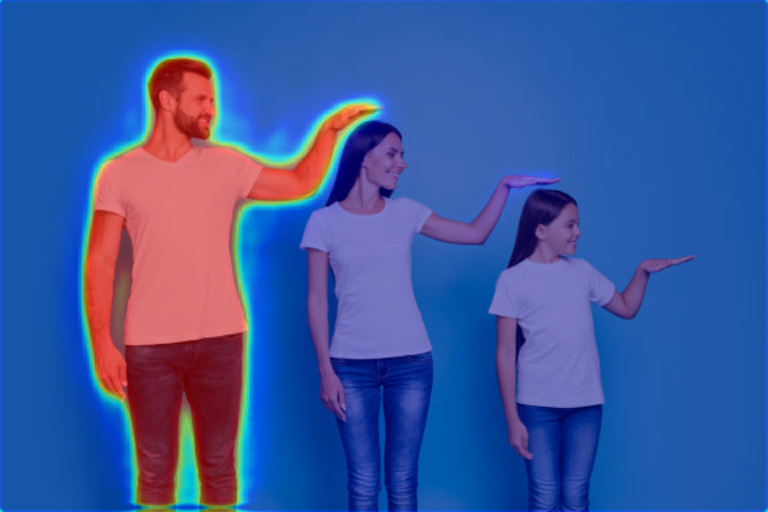}
        \end{subfigure}%
        \hspace{0.05cm}%
        \begin{subfigure}{0.49\columnwidth}
            \centering
            \caption{``Player swinging the bat"}
            \includegraphics[width=\columnwidth]{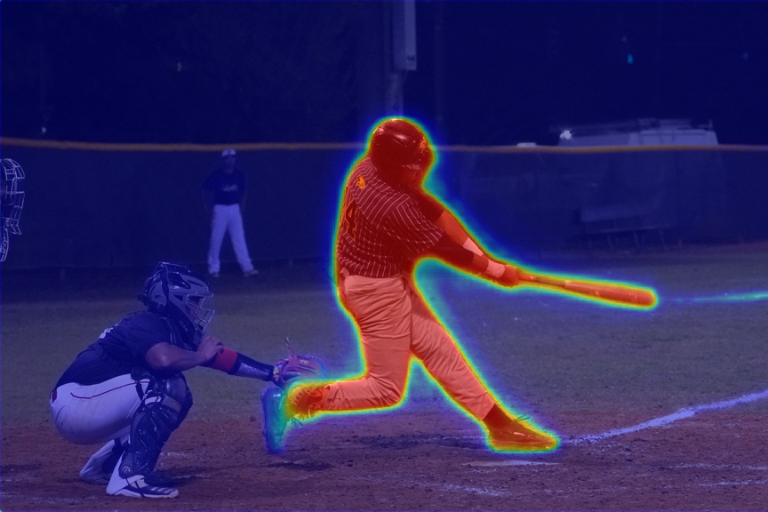}
        \end{subfigure}%
    \end{subfigure}
    \caption{Qualitative results showing LD-ZNet correctly segmenting attribute based queries. Text prompts for objects based on color attribute (top row), relative property and action attributes (bottom row) are well segmented by LD-ZNet.}
    \vspace{-1em}
    \label{fig:visual_results_supp}
\end{figure}

Figure \ref{fig:visual_results_supp} depicts attribute-based segmentation. Specifically, objects described by attributes based on color or relative properties (such as height) or actions are well segmented by LD-ZNet.

\subsection{AI-Generated Images}

\begin{figure*} [!t]
    \centering
    \begin{subfigure}[t]{0.19\textwidth}
        \centering
        \includegraphics[width=\linewidth]{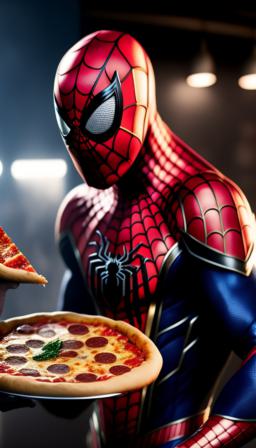}
        \includegraphics[width=\linewidth]{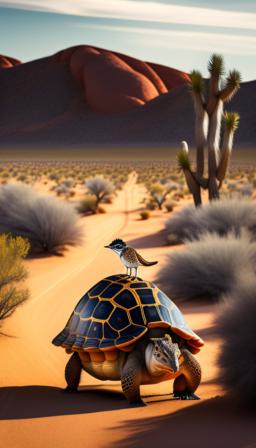}
        \includegraphics[width=\linewidth]{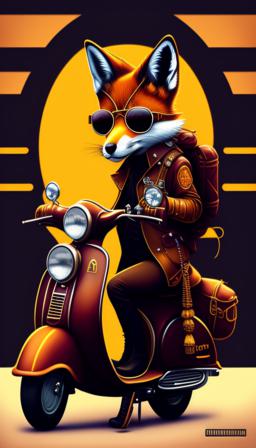}
        \includegraphics[width=\linewidth]{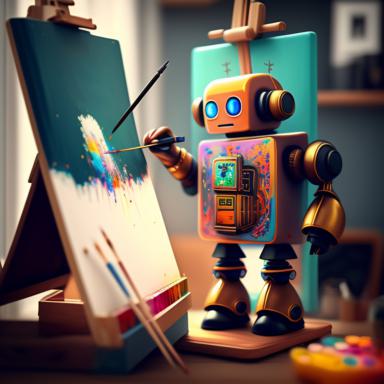}
        \caption{Input}
    \end{subfigure}%
    \hspace{0.05cm}%
    \begin{subfigure}[t]{0.19\textwidth}
        \centering
        \includegraphics[width=\linewidth]{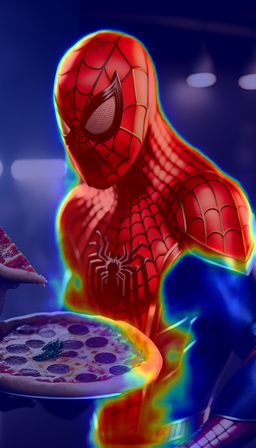}
        \includegraphics[width=\linewidth]{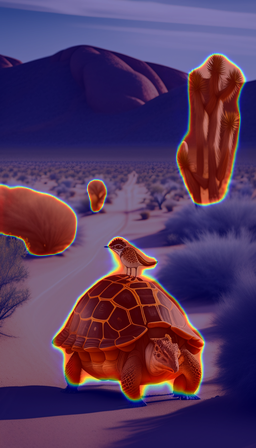}
        \includegraphics[width=\linewidth]{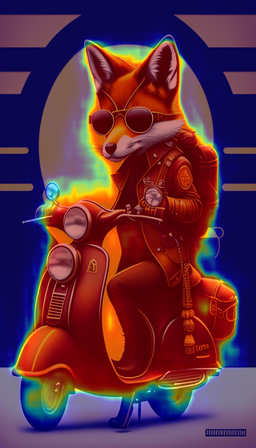}
        \includegraphics[width=\linewidth]{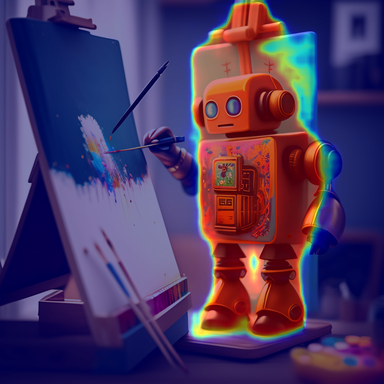}
        \caption{MDETR}
    \end{subfigure}%
    \hspace{0.05cm}%
    \begin{subfigure}[t]{0.19\textwidth}
        \centering
        \includegraphics[width=\linewidth]{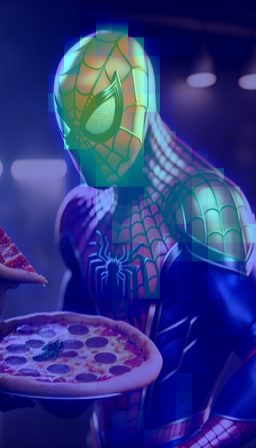}
        \includegraphics[width=\linewidth]{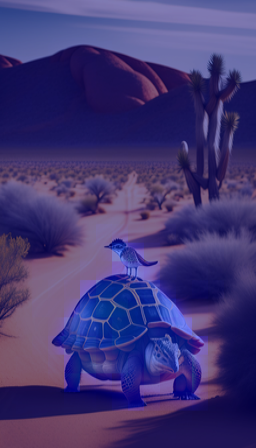}
        \includegraphics[width=\linewidth]{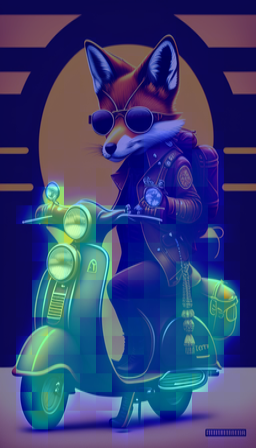}
        \includegraphics[width=\linewidth]{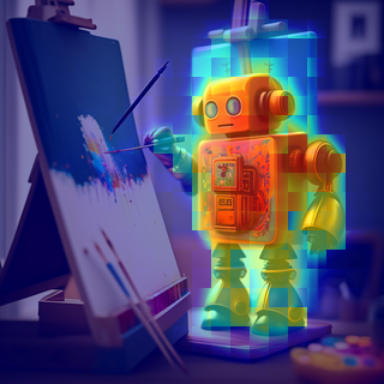}
        \caption{CLIPSeg}
    \end{subfigure}%
    \hspace{0.05cm}%
    \begin{subfigure}[t]{0.19\textwidth}
        \centering
        \includegraphics[width=\linewidth]{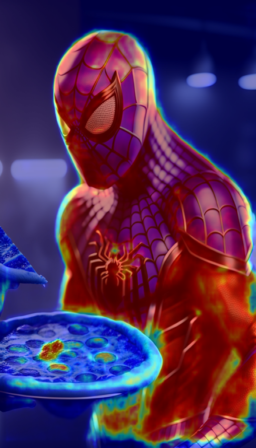}
        \includegraphics[width=\linewidth]{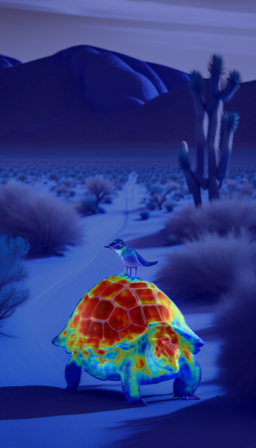}
        \includegraphics[width=\linewidth]{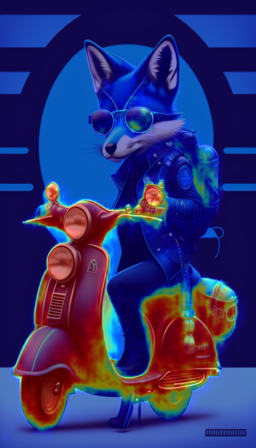}
        \includegraphics[width=\linewidth]{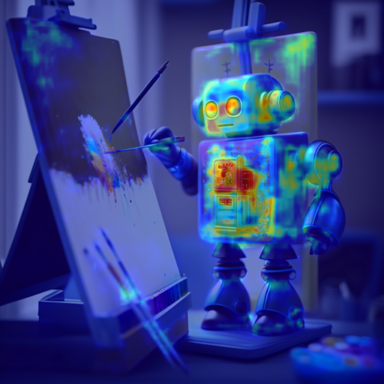}
        \caption{SEEM}
    \end{subfigure}%
    \hspace{0.05cm}%
    \begin{subfigure}[t]{0.19\textwidth}
        \centering
        \includegraphics[width=\linewidth]{Images/AI_gen_final/LD-ZNet/spider-man_as_a_robot_serving_pizza_spider-man.png}
        \includegraphics[width=\linewidth]{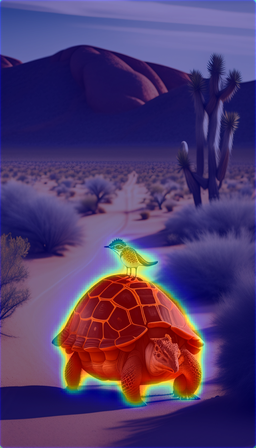}
        \includegraphics[width=\linewidth]{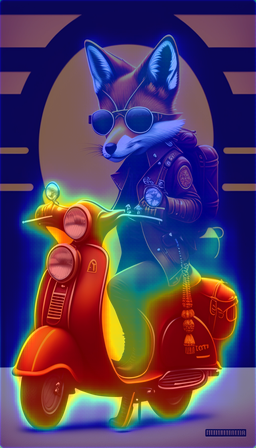}
        \includegraphics[width=\linewidth]{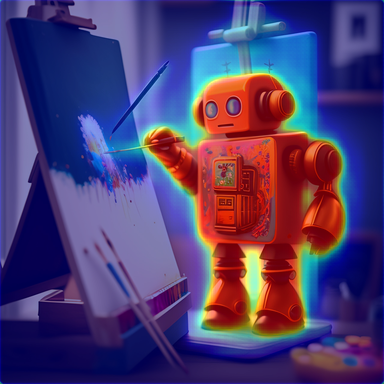}
        \caption{LD-ZNet}
    \end{subfigure}
    
    \caption{Qualitative comparison on the AI-generated images from AIGI dataset for text-based segmentation. The text prompts are \emph{``Spiderman"}, \emph{``tortoise"}, \emph{``vespa"} and \emph{``robot"} respectively.}
    \vspace{-1em}
    \label{fig:ai_generated_qualitative_supplementary}
\end{figure*}

% On our AIGI dataset, we also evaluate the newly released CLIPSeg model with fine-grained predictions and found it to be 3.5\% better (mIoU=59.9) than the originally released CLIPSeg version. Furthermore, 
We show more qualitative comparisons on our AIGI dataset in Figure \ref{fig:ai_generated_qualitative_supplementary}. We observe similar trend. While MDETR fails to segment the text prompts \emph{``Spiderman"}, \emph{``tortoise"}, \emph{``vespa"} and \emph{``robot"} due to novel concepts and domain gap, CLIPSeg estimates a rough segmentation on the most discriminative regions with lower confidence. However, LD-ZNet accurately segments in all the cases.

\subsection{Qualitative Comparisons on Diverse Domains}

\begin{figure*}[!t]
    \captionsetup[subfigure]{labelformat=empty}
    \centering
    \begin{subfigure}{0.85\linewidth}
        \centering
        \begin{subfigure}{0.32\columnwidth}
            \centering
            \caption{}
            \includegraphics[width=\columnwidth]{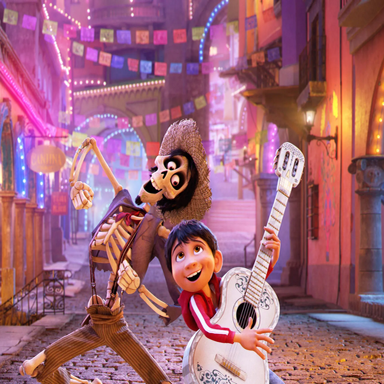}
            \includegraphics[width=\columnwidth]{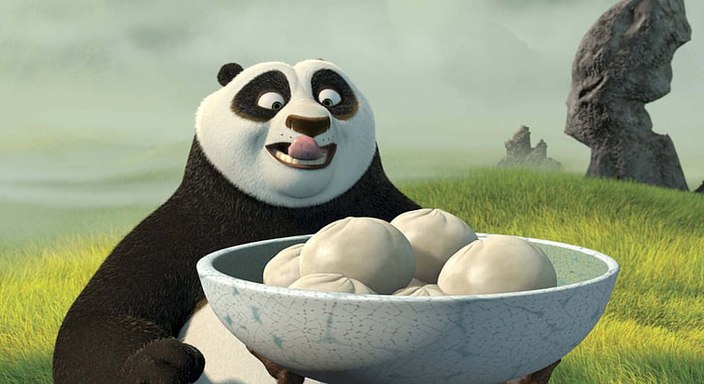}
        \end{subfigure}%
        \hspace{0.05cm}%
        \begin{subfigure}{0.32\columnwidth}
            \centering
            \caption{RGBNet}
            \includegraphics[width=\columnwidth]{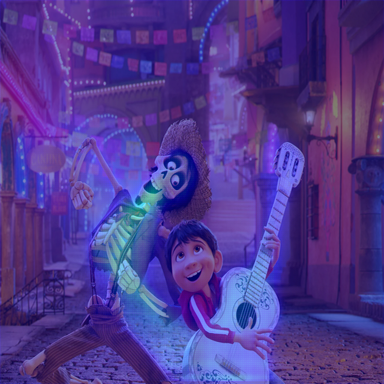}
            \includegraphics[width=\columnwidth]{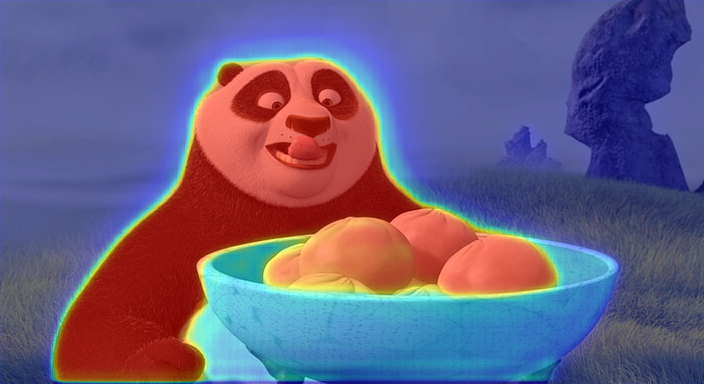}
        \end{subfigure}%
        \hspace{0.05cm}%
        \begin{subfigure}{0.32\columnwidth}
            \centering
            \caption{LD-ZNet}
            \includegraphics[width=\columnwidth]{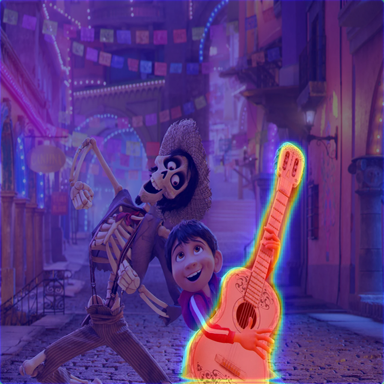}
            \includegraphics[width=\columnwidth]{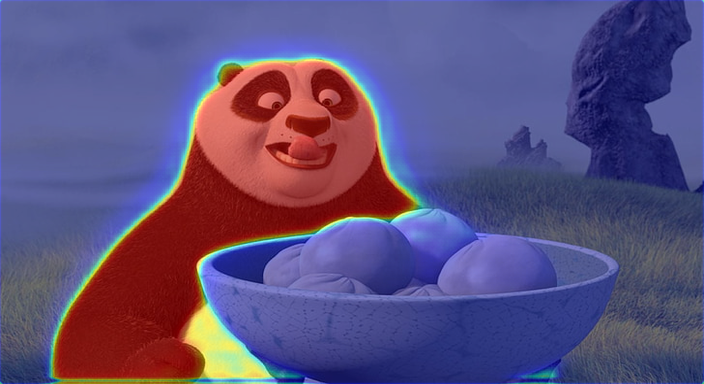}
        \end{subfigure}
        \vspace{-0.5em}
    \end{subfigure}
    
    \noindent\rule{0.9\linewidth}{0.4pt}

    \begin{subfigure}{0.85\linewidth}
        \vspace{-0.75em}
        \centering
        \begin{subfigure}{0.32\columnwidth}
            \centering
            \includegraphics[width=\columnwidth]{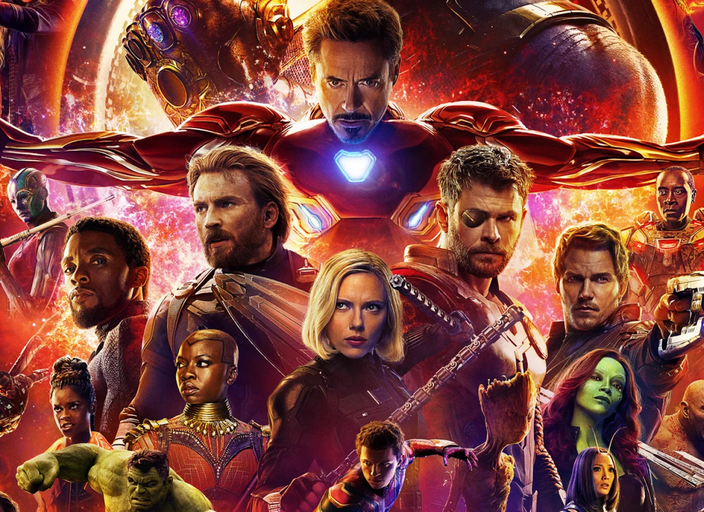}
            \includegraphics[width=\columnwidth]{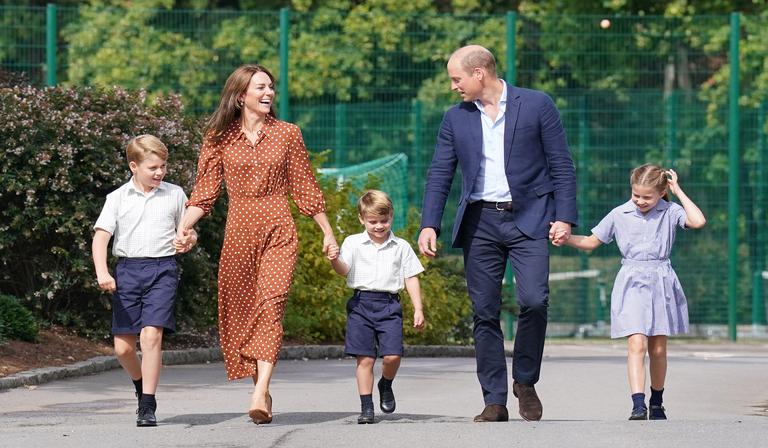}
        \end{subfigure}%
        \hspace{0.05cm}%
        \begin{subfigure}{0.32\columnwidth}
            \centering
            \caption{}
            \includegraphics[width=\columnwidth]{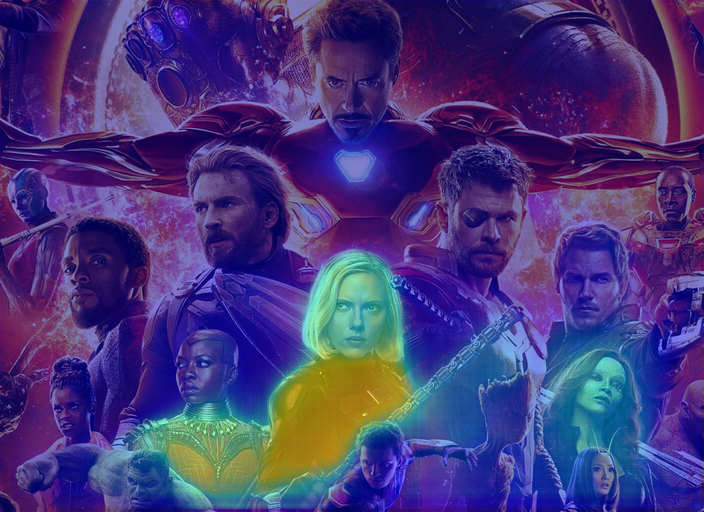}
            \includegraphics[width=\columnwidth]{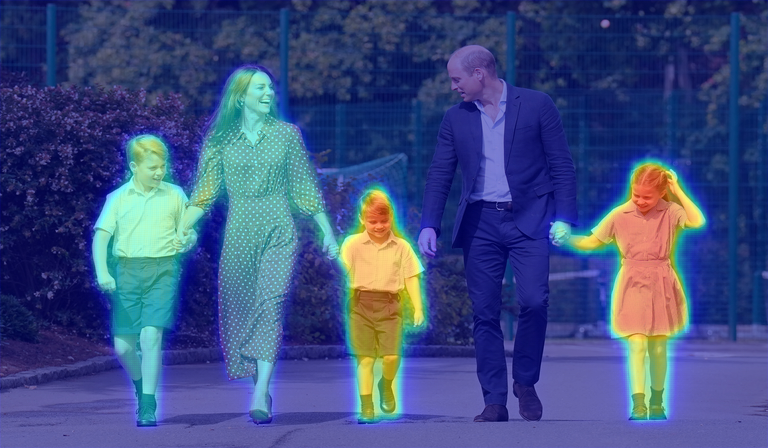}
        \end{subfigure}%
        \hspace{0.05cm}%
        \begin{subfigure}{0.32\columnwidth}
            \centering
            \caption{}
            \includegraphics[width=\columnwidth]{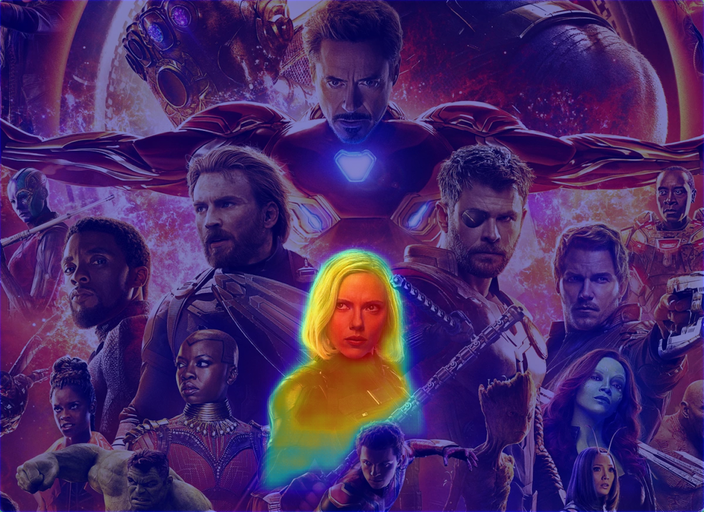}
            \includegraphics[width=\columnwidth]{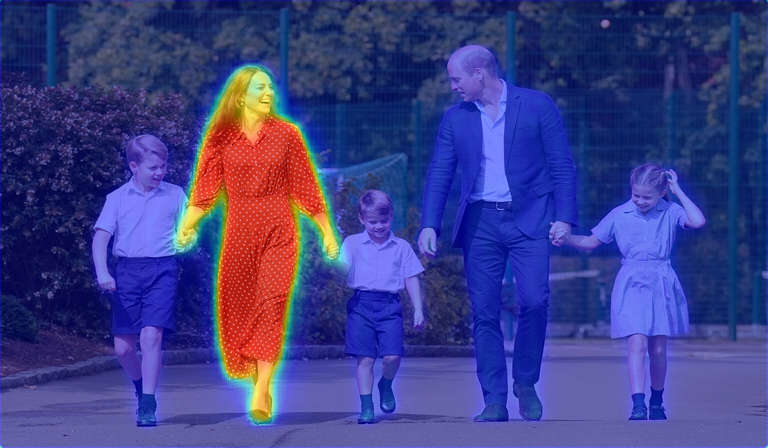}
        \end{subfigure}
        \vspace{-0.5em}
    \end{subfigure}
    
    \noindent\rule{0.9\linewidth}{0.4pt}
    
    \begin{subfigure}{0.85\linewidth}
        \vspace{-0.75em}
        \centering
        \begin{subfigure}{0.32\columnwidth}
            \centering
            \includegraphics[width=\columnwidth]{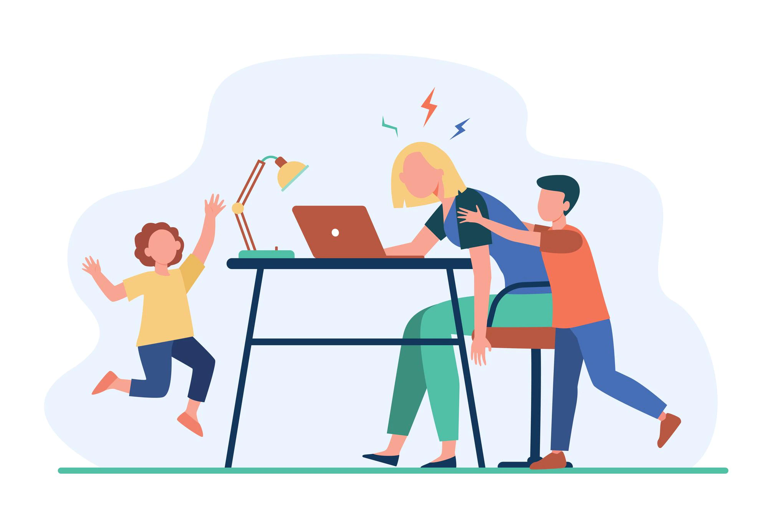}
            \includegraphics[width=\columnwidth]{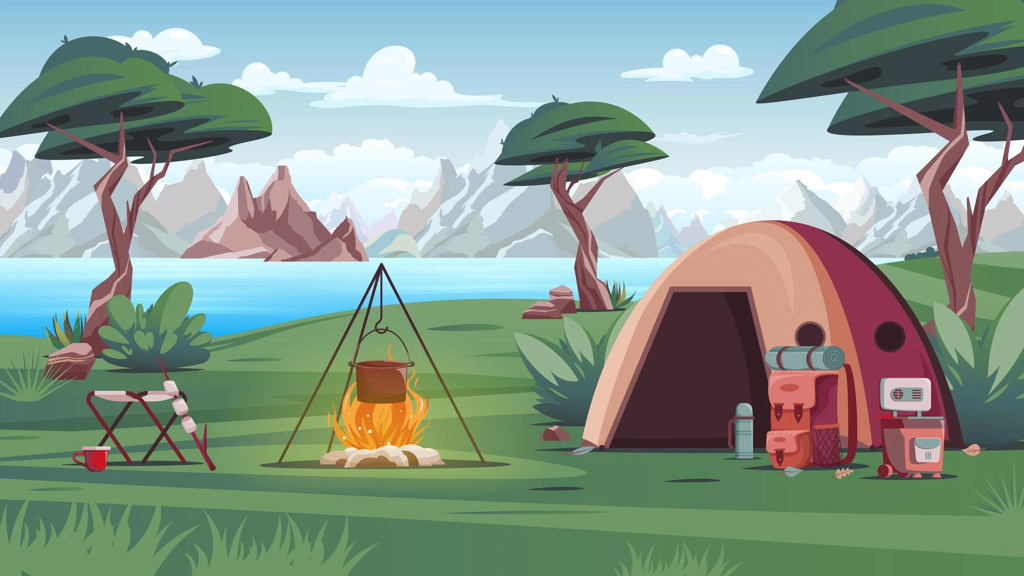}
        \end{subfigure}%
        \hspace{0.05cm}%
        \begin{subfigure}{0.32\columnwidth}
            \centering
            \caption{}
            \includegraphics[width=\columnwidth]{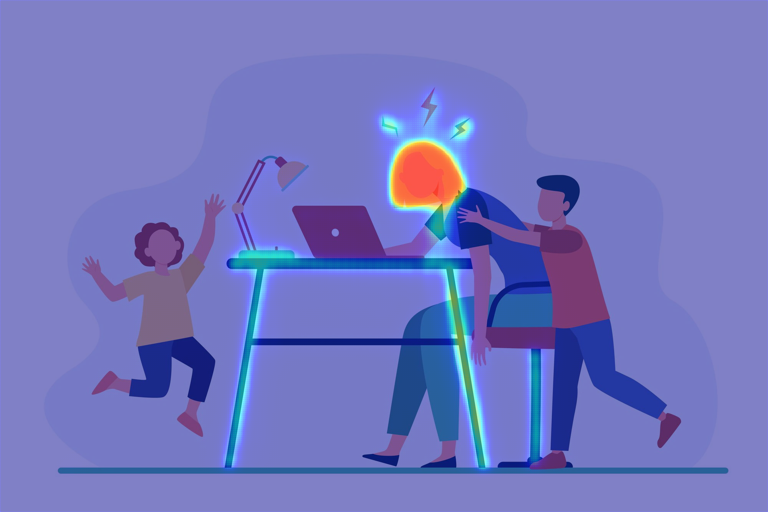}
            \includegraphics[width=\columnwidth]{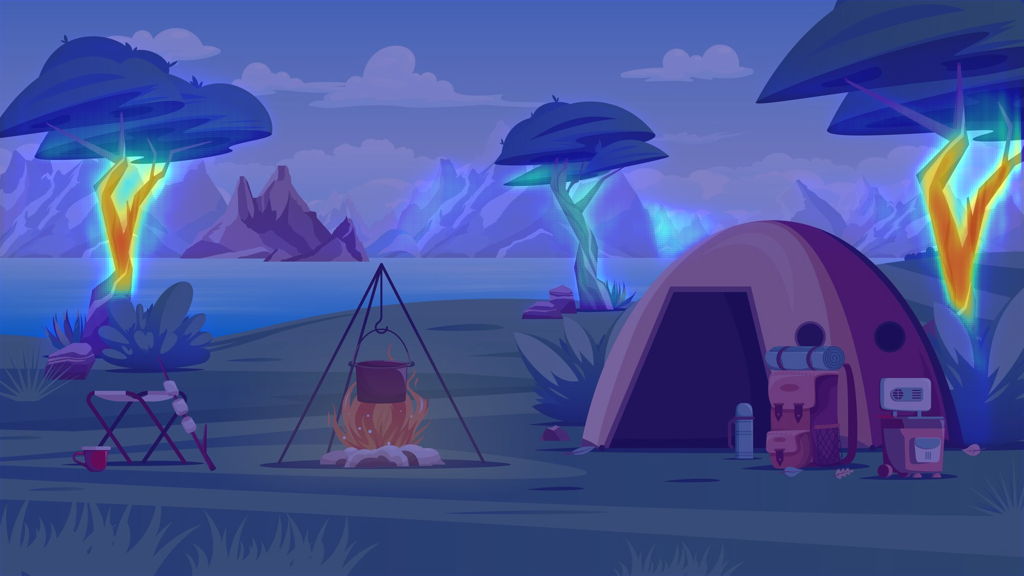}
        \end{subfigure}%
        \hspace{0.05cm}%
        \begin{subfigure}{0.32\columnwidth}
            \centering
            \caption{}
            \includegraphics[width=\columnwidth]{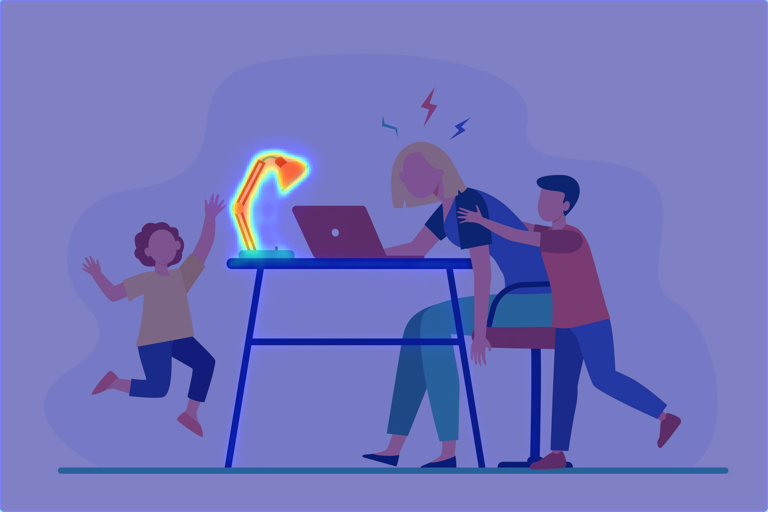}
            \includegraphics[width=\columnwidth]{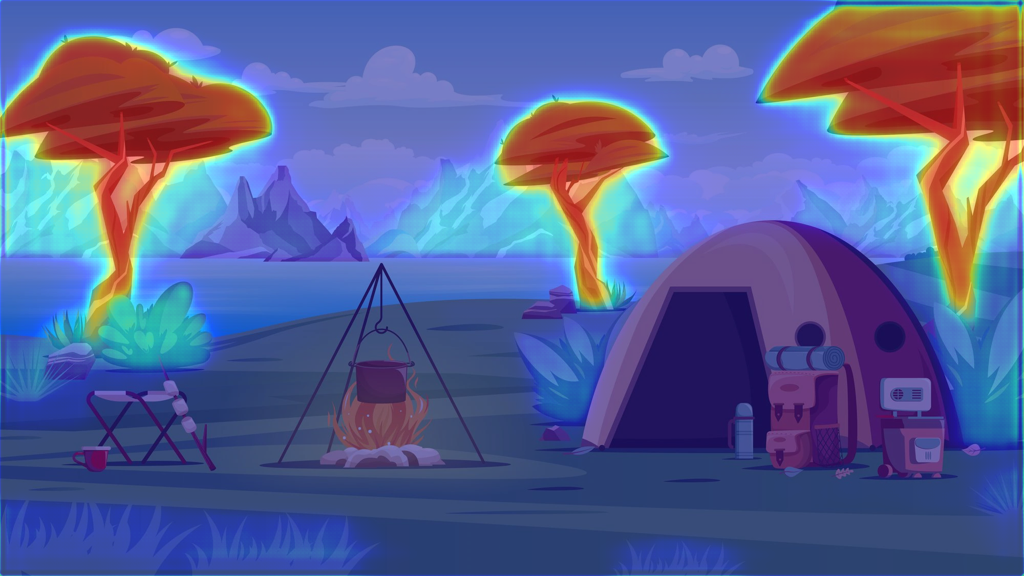}
        \end{subfigure}
    \end{subfigure}
    
    \caption{More qualitative examples where RGBNet fails to localize \emph{``Guitar", ``Panda"} from animation images (top row), famous celebrities \emph{``Scarlett Johansson", ``Kate Middleton"} (second row) and objects such as \emph{``Lamp", ``Trees"} from illustrations (bottom row). LD-ZNet benefits from using $z$ combined with the internal LDM features to correctly segment these text prompts.}
    \vspace{-1em}
    \label{fig:visual_results3_supp}
\end{figure*}

Figure \ref{fig:visual_results3_supp} demonstrates some specific cases where RGBNet fails to segment or poorly segments the object being referred to, where as LD-ZNet segments the objects better.

\subsection{Scene understanding}
\begin{figure*}[p]
    \captionsetup[subfigure]{labelformat=empty,font=small,labelfont={bf,sf}}
    \centering
    \begin{subfigure}{\textwidth}
        \centering
        \begin{subfigure}{0.32\textwidth}
            \centering
            \begin{subfigure}{\columnwidth}
                \centering
                \caption{}
                \includegraphics[width=\columnwidth]{Supp_Images/visual/zseg_SD_features/More_analysis/indoor_an_image_of_a_Books._image.png}
            \end{subfigure}
            \begin{subfigure}{\columnwidth}
                \centering
                \caption{``Books"}
                \includegraphics[width=\columnwidth]{Supp_Images/visual/zseg_SD_features/More_analysis/indoor_an_image_of_a_Books._mask.png}
            \end{subfigure}
        \end{subfigure}%
        \hspace{0.05cm}%
        \begin{subfigure}{0.32\textwidth}
            \centering
            \begin{subfigure}{\columnwidth}
                \centering
                \caption{``Flowers"}
                \includegraphics[width=\columnwidth]{Supp_Images/visual/zseg_SD_features/More_analysis/indoor_an_image_of_a_Flowers._mask.png}
            \end{subfigure}
            \begin{subfigure}{\columnwidth}
                \centering
                \caption{``Sofa"}
                \includegraphics[width=\columnwidth]{Supp_Images/visual/zseg_SD_features/More_analysis/indoor_an_image_of_a_sofa._mask.png}
            \end{subfigure}
        \end{subfigure}%
        \hspace{0.05cm}%
        \begin{subfigure}{0.32\textwidth}
            \centering
            \begin{subfigure}{\columnwidth}
                \centering
                \caption{``Table"}
                \includegraphics[width=\columnwidth]{Supp_Images/visual/zseg_SD_features/More_analysis/indoor_an_image_of_a_Table._mask.png}
            \end{subfigure}
            \begin{subfigure}{\columnwidth}
                \centering
                \caption{``Trees"}
                \includegraphics[width=\columnwidth]{Supp_Images/visual/zseg_SD_features/More_analysis/indoor_an_image_of_a_Trees._mask.png}
            \end{subfigure}
        \end{subfigure}
    \end{subfigure}
    \noindent\rule{\textwidth}{0.4pt}
    \begin{subfigure}{\textwidth}
        \centering
        \begin{subfigure}{0.32\textwidth}
            \centering
            \begin{subfigure}{\columnwidth}
                \centering
                \caption{}
                \includegraphics[width=\columnwidth]{Supp_Images/visual/RGBNet/camping_a_photograph_of_a_Trees._image.png}
            \end{subfigure}
            \begin{subfigure}{\columnwidth}
                \centering
                \caption{``Chair"}
                \includegraphics[width=\columnwidth]{Supp_Images/visual/zseg_SD_features/More_analysis/camping_an_image_of_a_Chair._mask.png}
            \end{subfigure}
        \end{subfigure}%
        \hspace{0.05cm}%
        \begin{subfigure}{0.32\textwidth}
            \centering
            \begin{subfigure}{\columnwidth}
                \centering
                \caption{``Clouds"}
                \includegraphics[width=\columnwidth]{Supp_Images/visual/zseg_SD_features/More_analysis/camping_an_image_of_a_Clouds._mask.png}
            \end{subfigure}
            \begin{subfigure}{\columnwidth}
                \centering
                \caption{``Grass"}
                \includegraphics[width=\columnwidth]{Supp_Images/visual/zseg_SD_features/More_analysis/camping_an_image_of_a_Grass._mask.png}
            \end{subfigure}
        \end{subfigure}%
        \hspace{0.05cm}%
        \begin{subfigure}{0.32\textwidth}
            \centering
            \begin{subfigure}{\columnwidth}
                \centering
                \caption{``Mountains"}
                \includegraphics[width=\columnwidth]{Supp_Images/visual/zseg_SD_features/More_analysis/camping_an_image_of_a_Mountains._mask.png}
            \end{subfigure}
            \begin{subfigure}{\columnwidth}
                \centering
                \caption{``River"}
                \includegraphics[width=\columnwidth]{Supp_Images/visual/zseg_SD_features/More_analysis/camping_an_image_of_a_River._mask.png}
            \end{subfigure}
        \end{subfigure}
    \end{subfigure}
    \noindent\rule{\textwidth}{0.4pt}
    \begin{subfigure}{\textwidth}
        \centering
        \begin{subfigure}{0.32\textwidth}
            \centering
            \begin{subfigure}{\columnwidth}
                \centering
                \caption{}
                \includegraphics[width=\columnwidth]{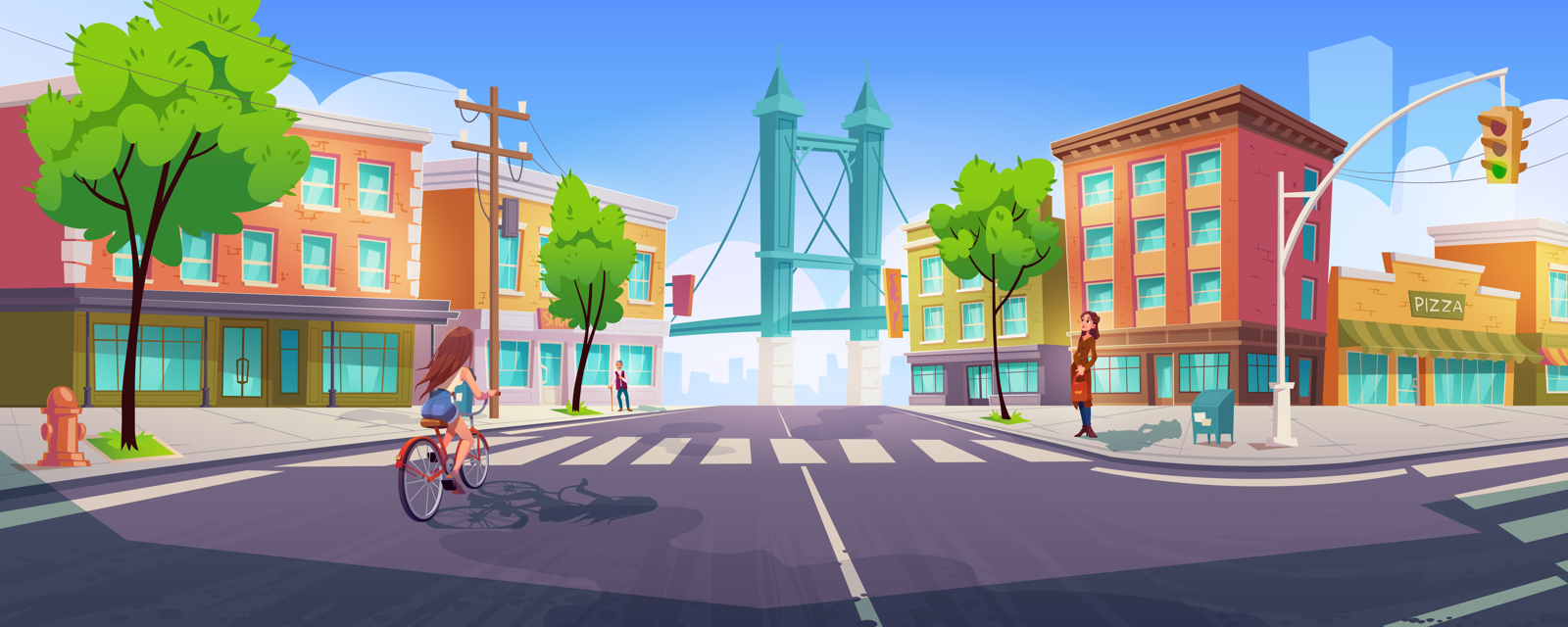}
            \end{subfigure}
            \begin{subfigure}{\columnwidth}
                \centering
                \caption{``Buildings"}
                \includegraphics[width=\columnwidth]{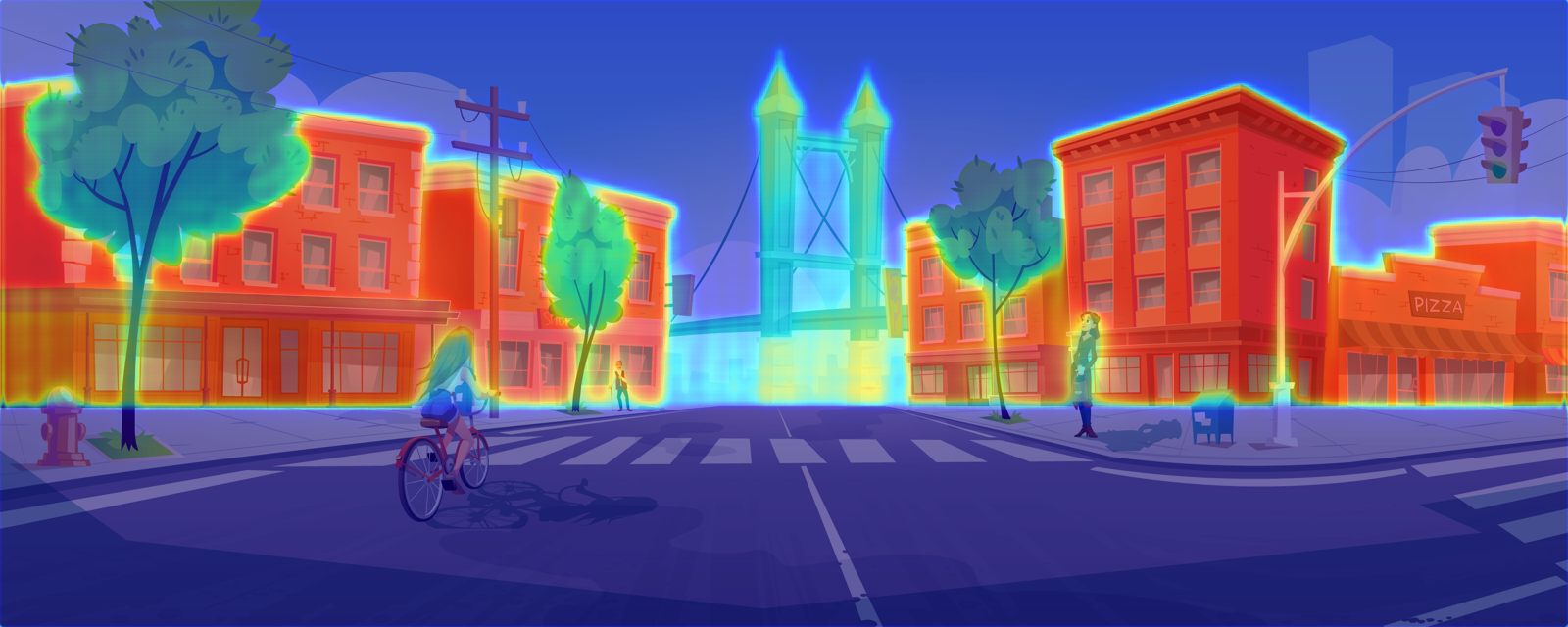}
            \end{subfigure}
        \end{subfigure}%
        \hspace{0.05cm}%
        \begin{subfigure}{0.32\textwidth}
            \centering
            \begin{subfigure}{\columnwidth}
                \centering
                \caption{``Trees"}
                \includegraphics[width=\columnwidth]{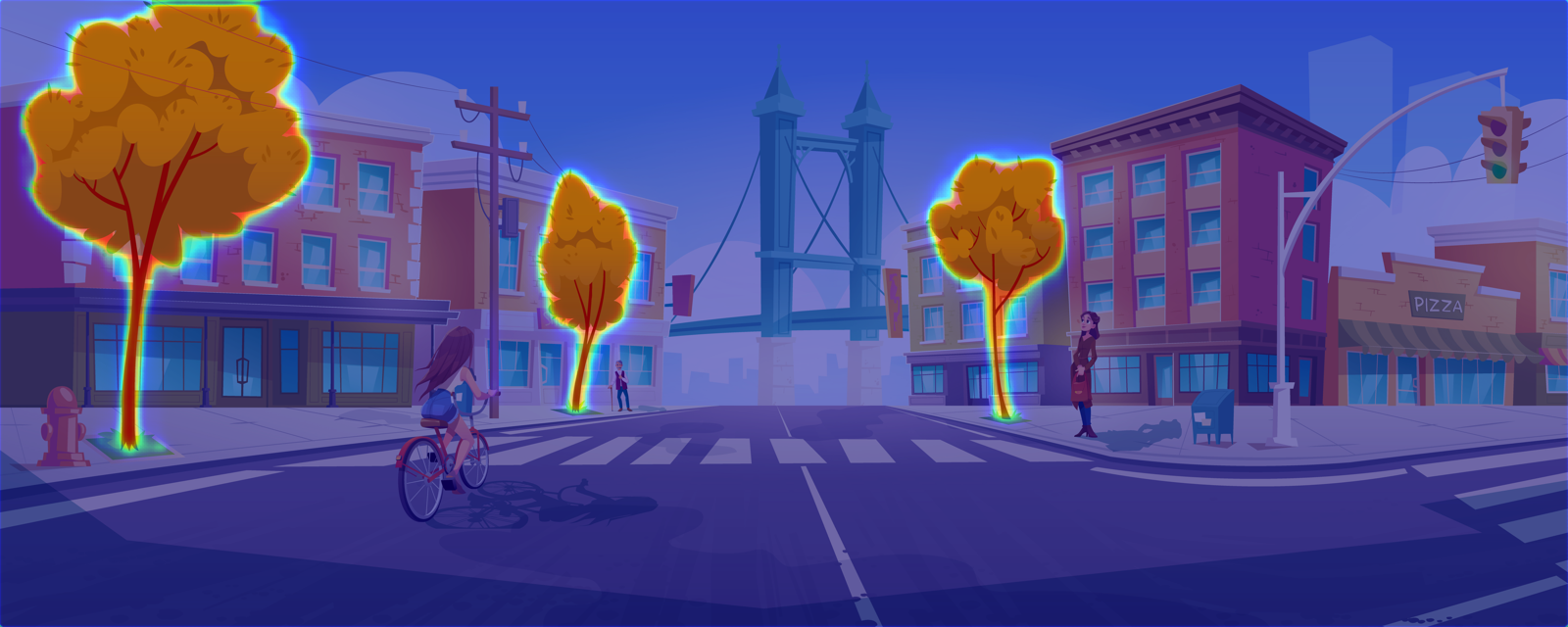}
            \end{subfigure}
            \begin{subfigure}{\columnwidth}
                \centering
                \caption{``Crosswalk"}
                \includegraphics[width=\columnwidth]{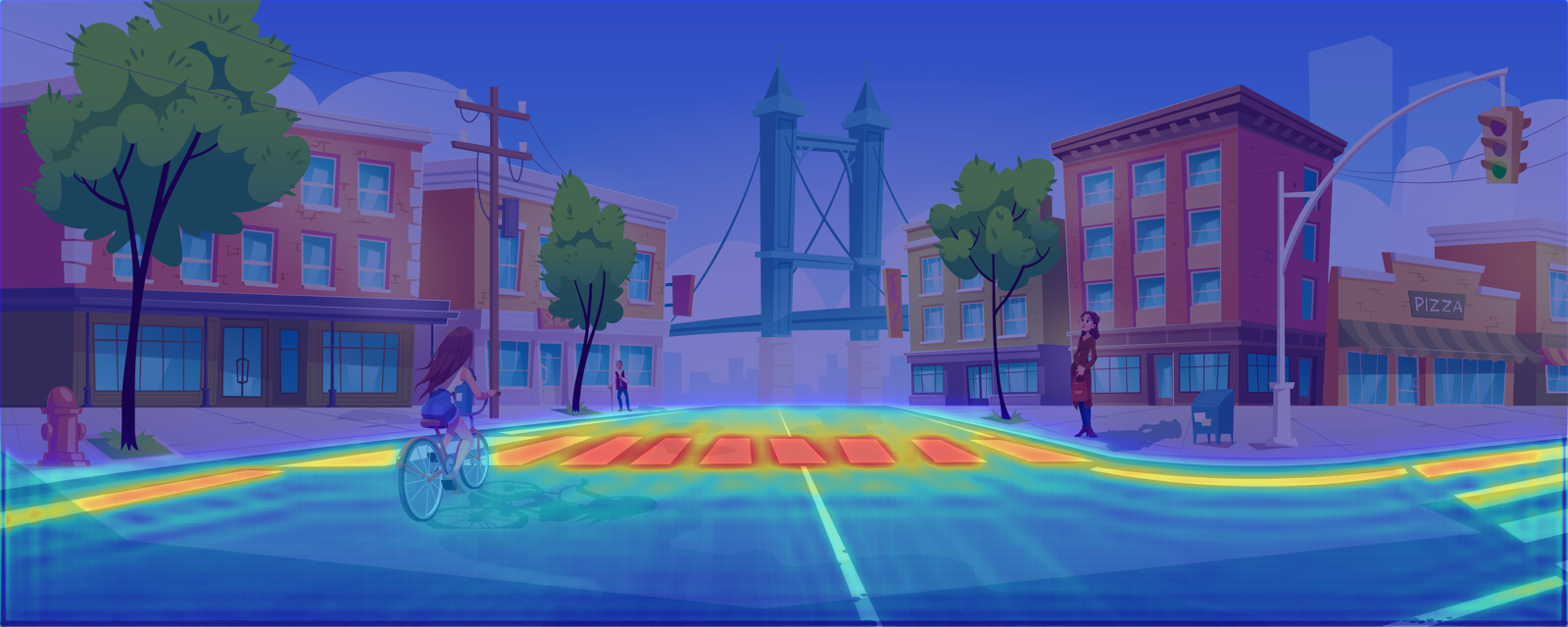}
            \end{subfigure}
        \end{subfigure}%
        \hspace{0.05cm}%
        \begin{subfigure}{0.32\textwidth}
            \centering
            \begin{subfigure}{\columnwidth}
                \centering
                \caption{``Bicycle"}
                \includegraphics[width=\columnwidth]{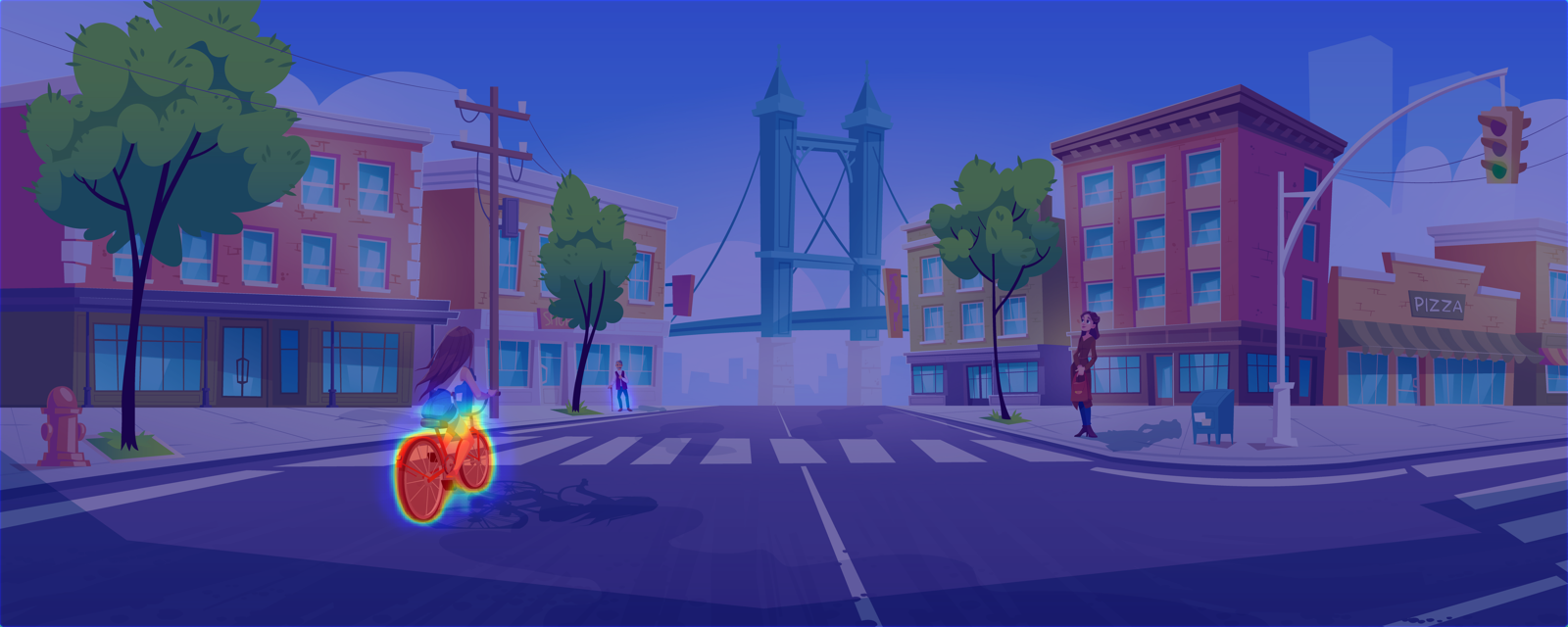}
            \end{subfigure}
            \begin{subfigure}{\columnwidth}
                \centering
                \caption{``Bridge"}
                \includegraphics[width=\columnwidth]{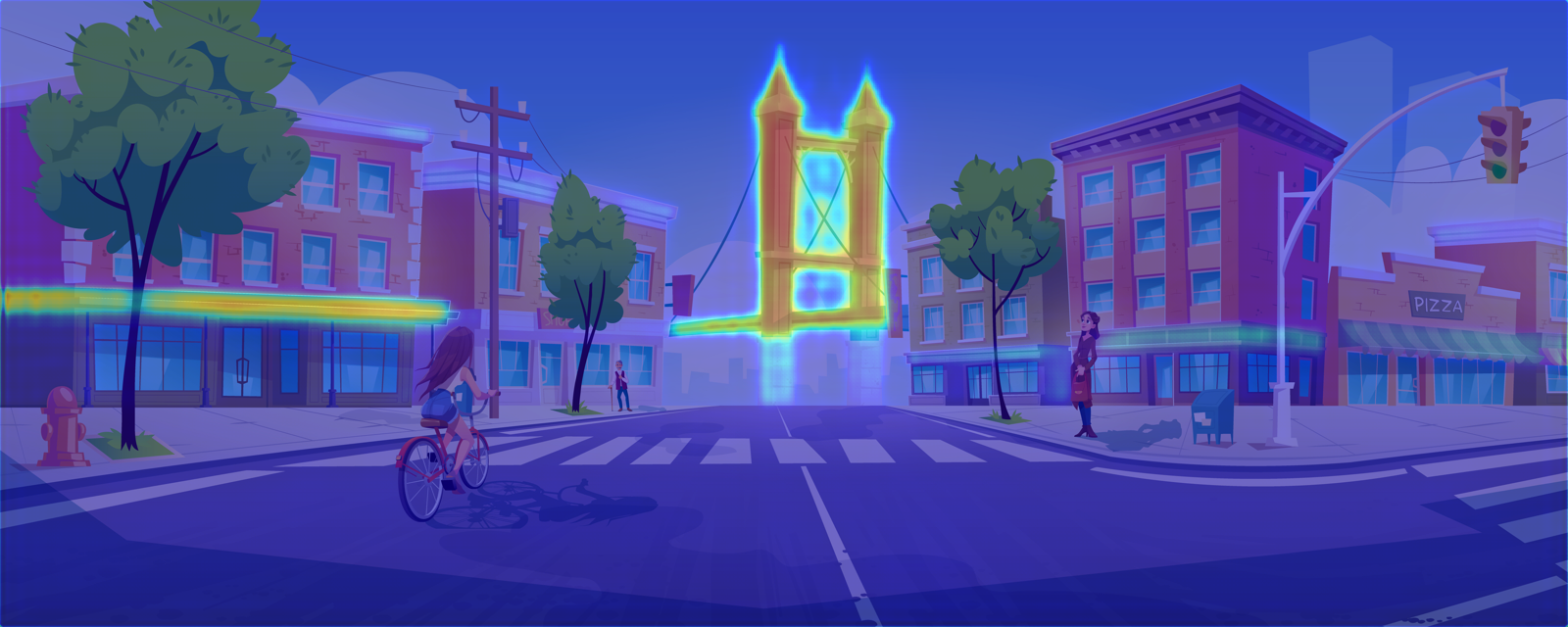}
            \end{subfigure}
        \end{subfigure}
    \end{subfigure}
    \caption{LD-ZNet text-based segmentation results for a diverse set of things and stuff classes across  both real images (top row) and illustrations (middle and bottom rows). High-quality segmentation across multiple object classes suggests that LD-ZNet has a good understanding of the overall scene. Images used from google and freepik.}
    \vspace{-1em}
    \label{fig:scene_understanding}
\end{figure*}

Figure \ref{fig:scene_understanding} shows the segmentation performance of LD-ZNet for several objects and regions in an image. Specifically, we show the segmentation for stuff classes such as ``Clouds", ``Mountains", ``Chair", ``Grass", ``River" \etc and thing classes such as ``Trees", ``Bicycle", ``Sofa", ``Books" \etc. The quality of the segmentation across multiple object classes suggests that LD-ZNet has a good understanding of the overall scene.

\begin{figure*}
    \centering
    \begin{subfigure}[t]{0.19\textwidth}
        \centering
        \includegraphics[width=\linewidth]{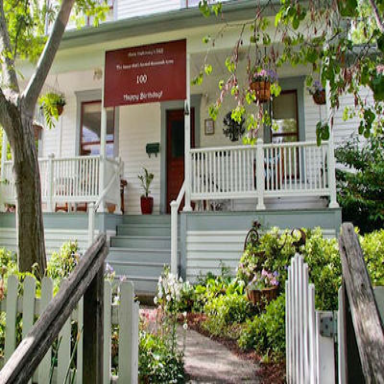}
        \includegraphics[width=\linewidth]{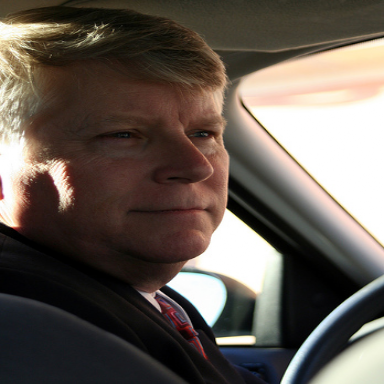}
        \includegraphics[width=\linewidth]{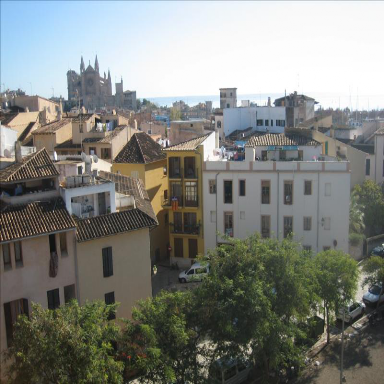}
        \includegraphics[width=\linewidth]{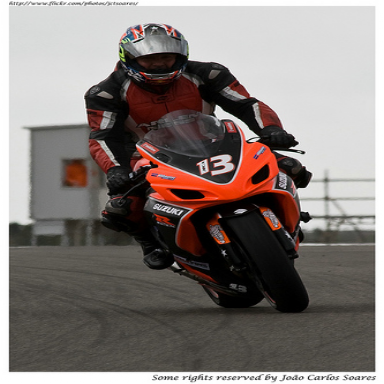}
        \includegraphics[width=\linewidth]{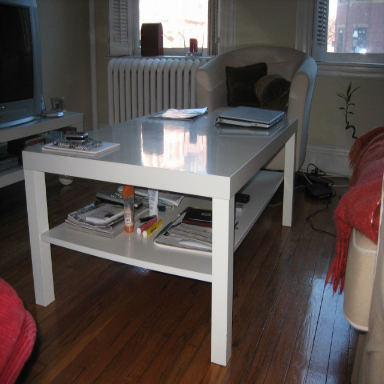}
        \caption{Input image}
        % \label{fig:architecture}
    \end{subfigure}%
    \hspace{0.05cm}%
    \begin{subfigure}[t]{0.19\textwidth}
        \centering
        \includegraphics[width=\linewidth]{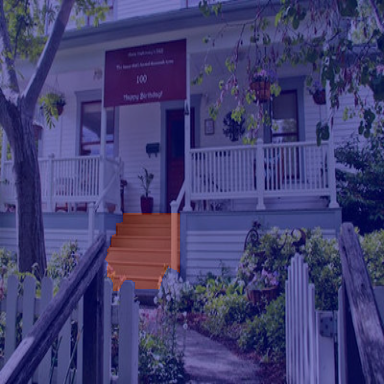}
        \includegraphics[width=\linewidth]{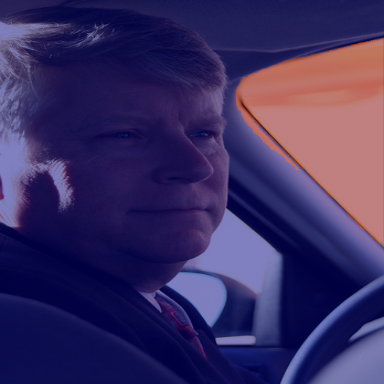}
        \includegraphics[width=\linewidth]{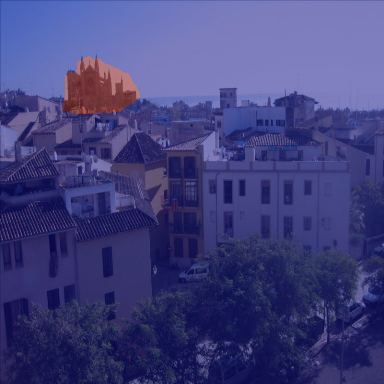}
        \includegraphics[width=\linewidth]{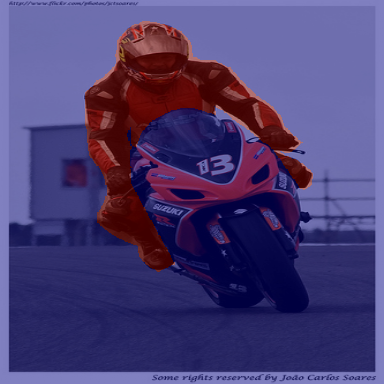}
        \includegraphics[width=\linewidth]{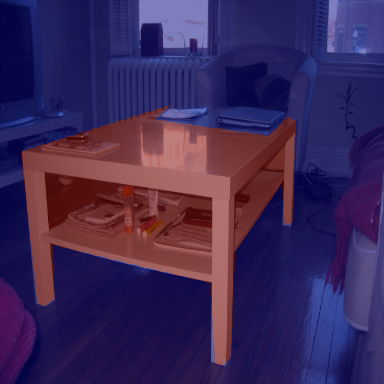}
        \caption{GT mask}
        % \label{fig:architecture}
    \end{subfigure}%
    \hspace{0.05cm}%
    \begin{subfigure}[t]{0.19\textwidth}
        \centering
        \includegraphics[width=\linewidth]{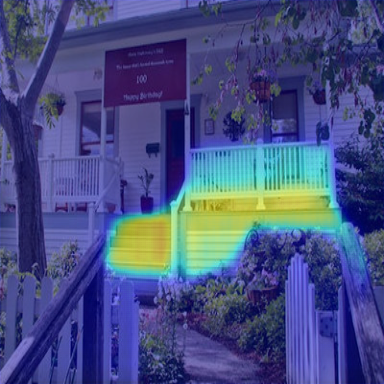}
        \includegraphics[width=\linewidth]{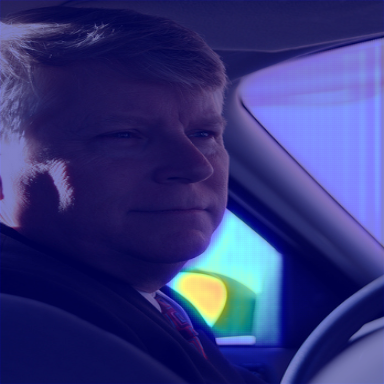}
        \includegraphics[width=\linewidth]{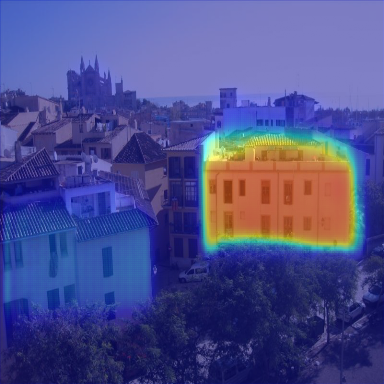}
        \includegraphics[width=\linewidth]{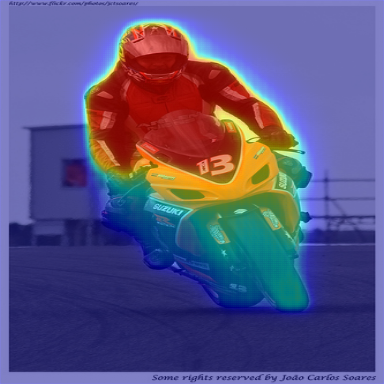}
        \includegraphics[width=\linewidth]{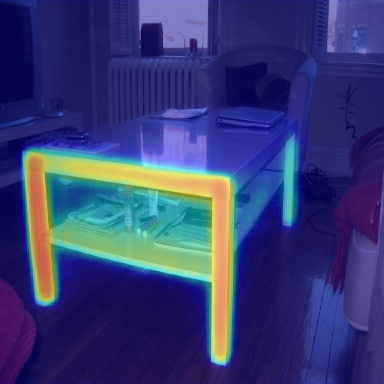}
        \caption{RGBNet}
        % \label{fig:architecture}
    \end{subfigure}%
    \hspace{0.05cm}%
    \begin{subfigure}[t]{0.19\textwidth}
        \centering
        \includegraphics[width=\linewidth]{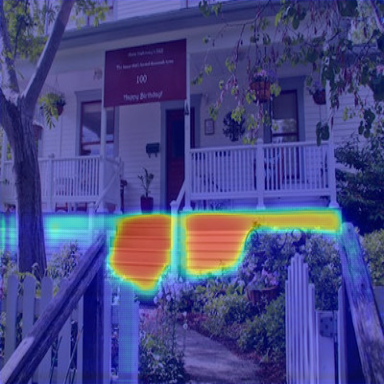}
        \includegraphics[width=\linewidth]{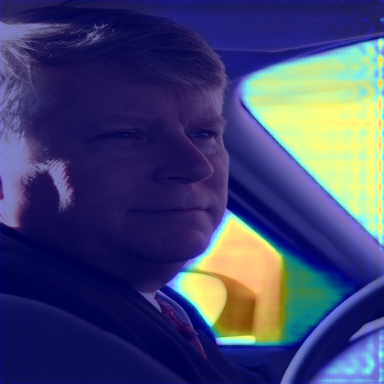}
        \includegraphics[width=\linewidth]{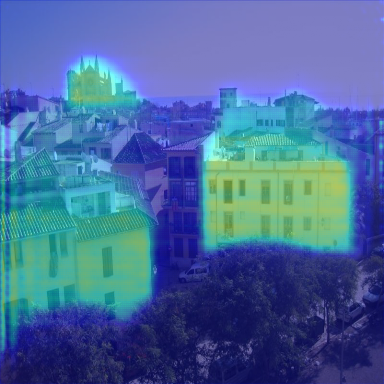}
        \includegraphics[width=\linewidth]{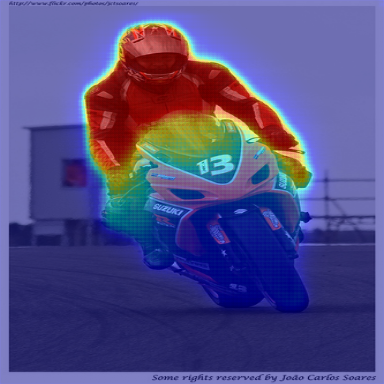}
        \includegraphics[width=\linewidth]{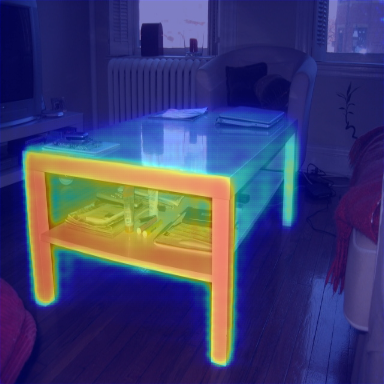}
        \caption{ZNet}
        % \label{fig:architecture}
    \end{subfigure}%
    \hspace{0.05cm}%
    \begin{subfigure}[t]{0.19\textwidth}
        \centering
        \includegraphics[width=\linewidth]{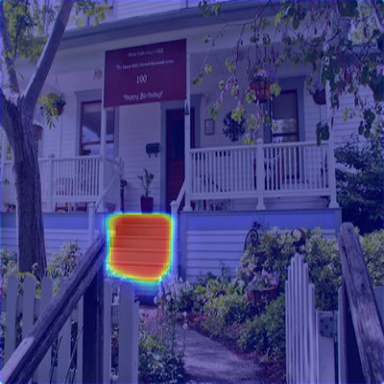}
        \includegraphics[width=\linewidth]{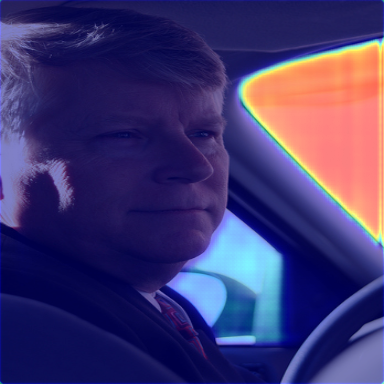}
        \includegraphics[width=\linewidth]{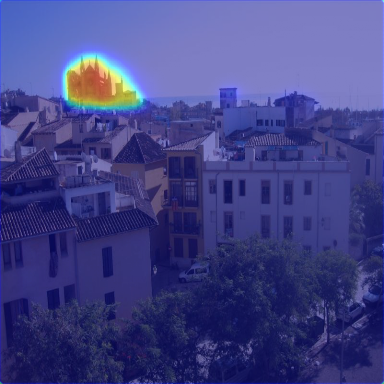}
        \includegraphics[width=\linewidth]{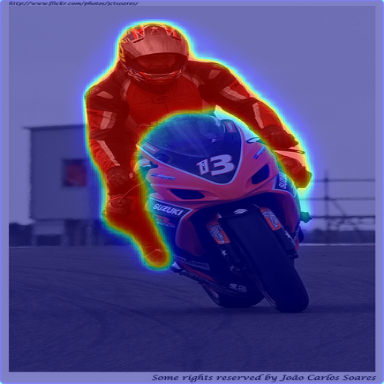}
        \includegraphics[width=\linewidth]{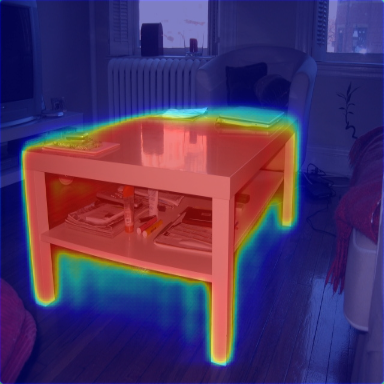}
        \caption{LD-ZNet}
        % \label{fig:architecture}
    \end{subfigure}
    
    \caption{Qualitative comparisons on the PhraseCut test dataset. Each row contains an RGB image along with a reference text as an input, with the goal being to segment out the image regions corresponding to the reference text. The reference texts are \emph{``grey steps"}, \emph{``glass windshield"}, \emph{``the tallest building"}, \emph{``riding person"}, \emph{``white stand"} for rows 1, 2, 3, 4 and 5 respectively. We show improvements using ZNet and LD-ZNet compared to the RGBNet.}
    \vspace*{3in}
    \label{fig:visual_results2_supp}
\end{figure*}

\subsection{Qualitative Comparisons on Phrasecut}
Figure \ref{fig:visual_results2_supp} shows qualitative comparisons of ZNet and LD-ZNet with the RGBNet baseline on the test dataset of PhraseCut. Attributes such as ``grey", ``glass", ``tallest", and ``riding" are well understood and localized in LD-ZNet.

% \section{Limitations}

% %%%%%%%%% REFERENCES
% {\small
% \bibliographystyle{ieee_fullname}
% \bibliography{egbib}
% }

% \end{document}

\end{document}